\documentclass{article}

\usepackage[eandd, final]{neurips_2026}

\usepackage[utf8]{inputenc} %
\usepackage[T1]{fontenc}    %
\usepackage{hyperref}       %
\usepackage{url}            %
\usepackage{booktabs}       %
\usepackage{amsfonts}       %
\usepackage{nicefrac}       %
\usepackage{microtype}      %
\usepackage[dvipsnames]{xcolor}
\usepackage{makecell}
\usepackage{transparent}
\usepackage{longtable}

\PassOptionsToPackage{authoryear,round}{natbib}
\definecolor{fgreen}{HTML}{406825}
\definecolor{flgreen}{HTML}{e7f2e1}
\definecolor{fblue}{HTML}{1f4985}
\definecolor{flblue}{HTML}{ecf2fe}
\definecolor{fred}{HTML}{b44529}
\definecolor{flred}{HTML}{fbeae5}
\definecolor{fyellow}{HTML}{af8725}
\definecolor{flyellow}{HTML}{fef0c9}
\definecolor{fwhite}{HTML}{8e8e8e} %
\definecolor{flwhite}{HTML}{fafafa}
\newcommand{\highlight}[2]{\colorbox{fl#1}{\textcolor{f#1}{#2}}}
\newcommand{\highlightem}[2]{\emph{#2} (\highlight{#1}{#1})}

\usepackage{amsmath}
\usepackage{graphicx}
\usepackage{cleveref}
\usepackage{subcaption}
\usepackage{enumitem}

\usepackage{listings}
\lstdefinestyle{mystyle}{
    basicstyle=\ttfamily\footnotesize, %
    breaklines=true, %
    breakatwhitespace=true, %
    frame=single, %
    backgroundcolor=\color{lightgray!20}, %
}
\usepackage{colortbl}
\usepackage{xcolor}
\usepackage{array}
\usepackage{booktabs}
\definecolor{poscolor}{HTML}{9eb0ff} %
\definecolor{negcolor}{HTML}{ffadad} %
\definecolor{altrow}{HTML}{f5f5f5}   %
\definecolor{headercolor}{HTML}{fcf1cf} %
\definecolor{lightgrey}{RGB}{210,210,210}  %
\newlength{\rowspacing}
\newcommand{\tablefontsize}{\scriptsize}

\renewcommand{\tablefontsize}{\scriptsize}
\newcommand{\tablesmallfontsize}{\tiny}

\newcommand{\ghurl}{\texttt{\href{https://github.com/rdnfn/feedback-forensics}{github.com/rdnfn/feedback-forensics}}}
\newcommand{\appurl}{\texttt{\href{https://app.feedbackforensics.com/}{app.feedbackforensics.com}}}
\newcommand{\dataurl}{\texttt{\href{https://hf.co/datasets/rdnfn/ff-annotations}{hf.co/datasets/rdnfn/ff-annotations}}}
\newcommand{\generationsurl}{\href{https://hf.co/datasets/rdnfn/ff-model-personality}{hf.co/datasets/rdnfn/ff-model-personality}}

\newcommand{\strengthexpl}{Values are \emph{strength} metric with 95\% CI and insignificant results greyed out.}

\title{Feedback Forensics:\\A Toolkit to Measure AI Personality}

\newcommand{\authorcol}[1]{\makebox[\dimexpr 0.5\textwidth-2\tabcolsep\relax]{#1}}
\author{%
  Arduin Findeis\thanks{\texttt{arduin.findeis@cst.cam.ac.uk}; \textsuperscript{\dag}\texttt{timo.kaufmann@ifi.lmu.de}
  } \\
  \authorcol{University of Cambridge} \\
  \And
  Timo Kaufmann\footnotemark[2] \\
  \authorcol{LMU Munich, MCML Munich} \\
  \AND
  Eyke Hüllermeier \\
  \authorcol{LMU Munich, MCML Munich, DFKI} \\
  \And
  Robert Mullins \\
  \authorcol{University of Cambridge} \\
}

\begin{document}

\maketitle

\begin{abstract}

Some traits making a ``good'' AI model are hard to describe upfront. For example, should responses be more \emph{polite} or more \emph{casual}? Such traits are sometimes summarized as model \emph{character} or \emph{personality}.
Without a clear objective, conventional benchmarks based on automatic checks struggle to measure such traits.
Evaluation methods using human feedback such as Chatbot Arena have emerged as a popular alternative. 
These methods infer ``better'' personality and other desirable traits \emph{implicitly} by ranking multiple model responses relative to each other.
Recent issues with model releases highlight limitations of these existing opaque evaluation approaches: goblin-obsessed, sycophantic and generally over-the-top personalities have resulted in model rollbacks and extensive media coverage.
Despite these known issues, limited public tooling exists to \emph{explicitly} evaluate model personality. We introduce \emph{Feedback Forensics}: an open-source toolkit to track AI personality and related behavioural traits, both those \emph{encouraged by human (or AI) feedback}, and those \emph{exhibited across AI models} trained and evaluated on such feedback. Feedback Forensics is \emph{adaptive} and can automatically find new traits appearing in models. We demonstrate the toolkit's usefulness in two steps: (A)~first we analyse the personality traits encouraged in popular human feedback datasets including \emph{MultiPref}, \emph{PRISM} and data from \emph{Chatbot Arena}; and (B)~then use our toolkit to analyse how much popular models exhibit such traits.
We release (1)~our \emph{Feedback Forensics} toolkit alongside (2) a \emph{web app} tracking AI personality in popular models and feedback datasets and (3) the underlying \emph{annotation data}.%
\footnote{%
\parbox[t]{\linewidth}{%
    \texttt{Code:}~\ghurl{}, \texttt{Web app:}~\appurl{},\\%
    \texttt{Data:}~\dataurl{}%
  }%
}
\end{abstract}

\begin{figure}[ht]
\centering
\includegraphics[width=0.90\textwidth]{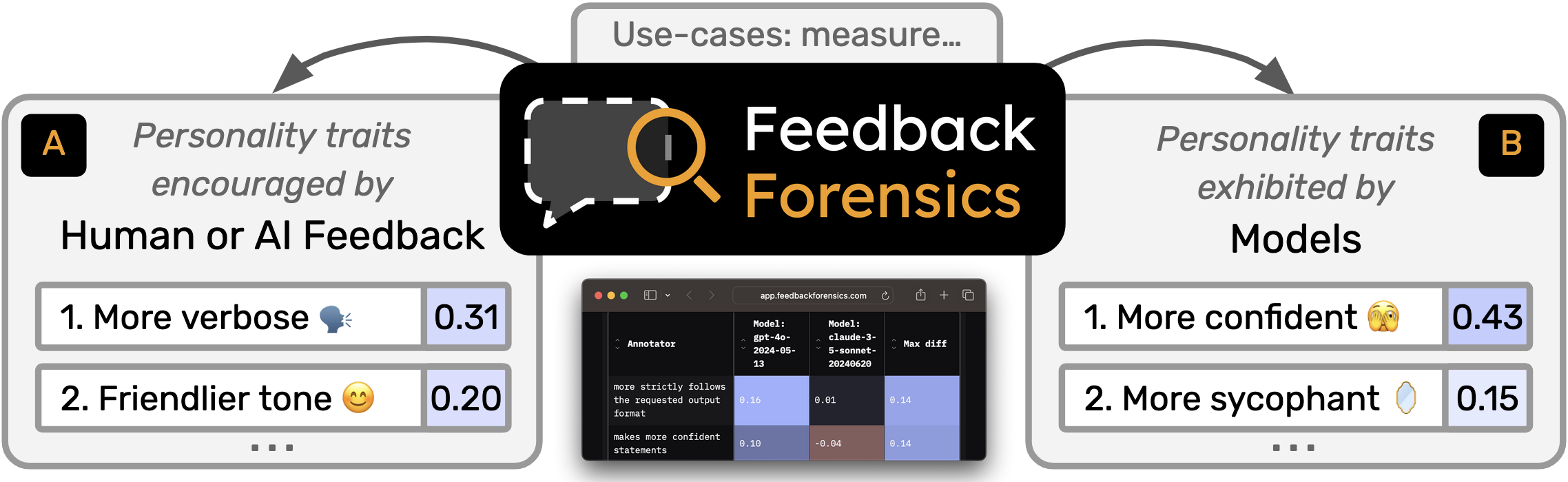}
\caption{\textbf{Overview of our \emph{Feedback Forensics} toolkit.}}
\label{fig:overview}
\end{figure}

\begin{figure}[ht]
\centering
\includegraphics[width=0.95\textwidth]{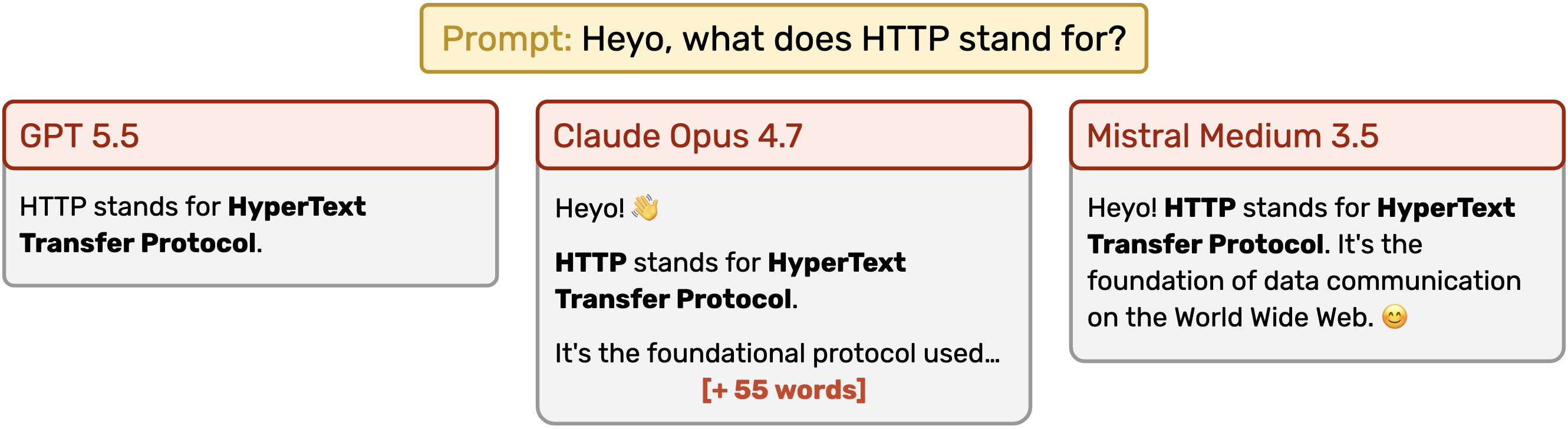}
\caption{\textbf{Example of model personality differences.} All models decipher the HTTP acronym correctly but the \emph{manner} or \emph{personality} of their responses varies. GPT-5.5 keeps the response very \emph{concise}. The Claude and Mistral models create more \emph{engaging} responses, with more \emph{bold} and \emph{emojis}. The Claude model responds more \emph{verbosely} than the others. Standard benchmarks fail to identify these differences in models' personalities. Our Feedback Forensics toolkit can quantify them.}
\label{fig:intro:example_personality}
\end{figure}

\section{Introduction}

Conventional benchmarks for evaluating large language models, such as MMLU \citep{hendrycks2021MeasuringMassiveMultitask}, do not capture many aspects of AI model behaviour. 
Beyond factual correctness and coding capabilities, traits such as \emph{tone} or \emph{style} also matter to users, but are more challenging to evaluate.
As illustrated in \Cref{fig:intro:example_personality}, not just the \emph{content} but also the \emph{manner} of responses is important for the user experience \citep{lambert2025CharacterTrainingUnderstanding}. Such behaviour traits relating to the manner of responses are sometimes collectively referred to as model \emph{character} or \emph{personality}. %
In this work, we take a closer look at \emph{model personality} in this general sense, using the term \emph{personality trait} to refer to any characteristic of a model's responses that (1) distinguishes that model's from other models' responses and (2) is not commonly considered a model capability.\footnote{For example, we consider \emph{writing style} as a personality trait but not \emph{coding capabilities}. See \Cref{sec:related_work} for a discussion of how our definition relates to others in the literature.}

Due to the ambiguous nature of style and manner, \emph{``good''} model personality is difficult to define explicitly. Conventional benchmarks based on multiple choice or other forms of automated validation cannot be applied directly. Evaluation methods based on feedback datasets, such as Chatbot Arena \citep{chiang2024ChatbotArenaOpen}, have emerged as a popular alternative. Such methods are able to capture subtle behaviour improvements, including in terms of personality, without needing to explicitly define what a \emph{``good''} personality is. Instead, \emph{``better''} personality is implicitly defined by ranking multiple model responses relative to each other. Given the implicit setup, our understanding of the concrete \emph{personality changes} encouraged by such feedback datasets and \emph{personality differences} between models is typically limited.

Recent issues with the personality of frontier models further highlight the limits of current evaluation methods. OpenAI models excessively mentioned goblins due to flawed personality training \citep{openai2026WhereGoblinsCame}. Previously, a model in ChatGPT was rolled back over concerns of an overly sycophantic personality, excessively flattering and agreeing with users~\citep{openai2025ExpandingWhatWe}. Concerns were also raised around the verbose and emoji-heavy personality of an experimental version of Llama-4-Maverick on Chatbot Arena \citep{wiggers2025MetasBenchmarksIts}. %
These observations highlight the need for more robust tooling to measure personality traits -- better tooling could make such drifts in personality more visible and help create models with more desirable traits. 

\textbf{Contributions.} 
We introduce \emph{Feedback Forensics}, a Python toolkit to measure personality traits, and release a corresponding web app and annotation data:

\begin{enumerate}[wide, labelwidth=0pt, labelindent=0pt, left=6pt]
    \item \textbf{Open-source \emph{Feedback Forensics} Python toolkit for measuring AI personality traits.} We implement a comprehensive Python toolkit to measure personality and related behavioural traits \emph{exhibited by models} and \emph{encouraged by pairwise feedback data}. Leveraging the \emph{Inverse Constitutional AI} (ICAI) pipeline by \citet{findeis2025inverse}, our toolkit can be used to measure and visualize personality traits.
    \item \textbf{Web platform tracking personality in popular models and feedback datasets.} In addition to the Python toolkit for local usage, we also provide a web platform to inspect personality traits observed in popular models and datasets, available at~\appurl{}.
    \item \textbf{Annotation data from experiments.} Accompanying our experimental results, we release the underlying AI-annotator-generated personality annotations publicly to enable further analysis, available at \dataurl{}. See \Cref{app:ff-annotations} for further details.
\end{enumerate}

\begin{figure}[b!]
\centering
\includegraphics[width=0.99\textwidth]{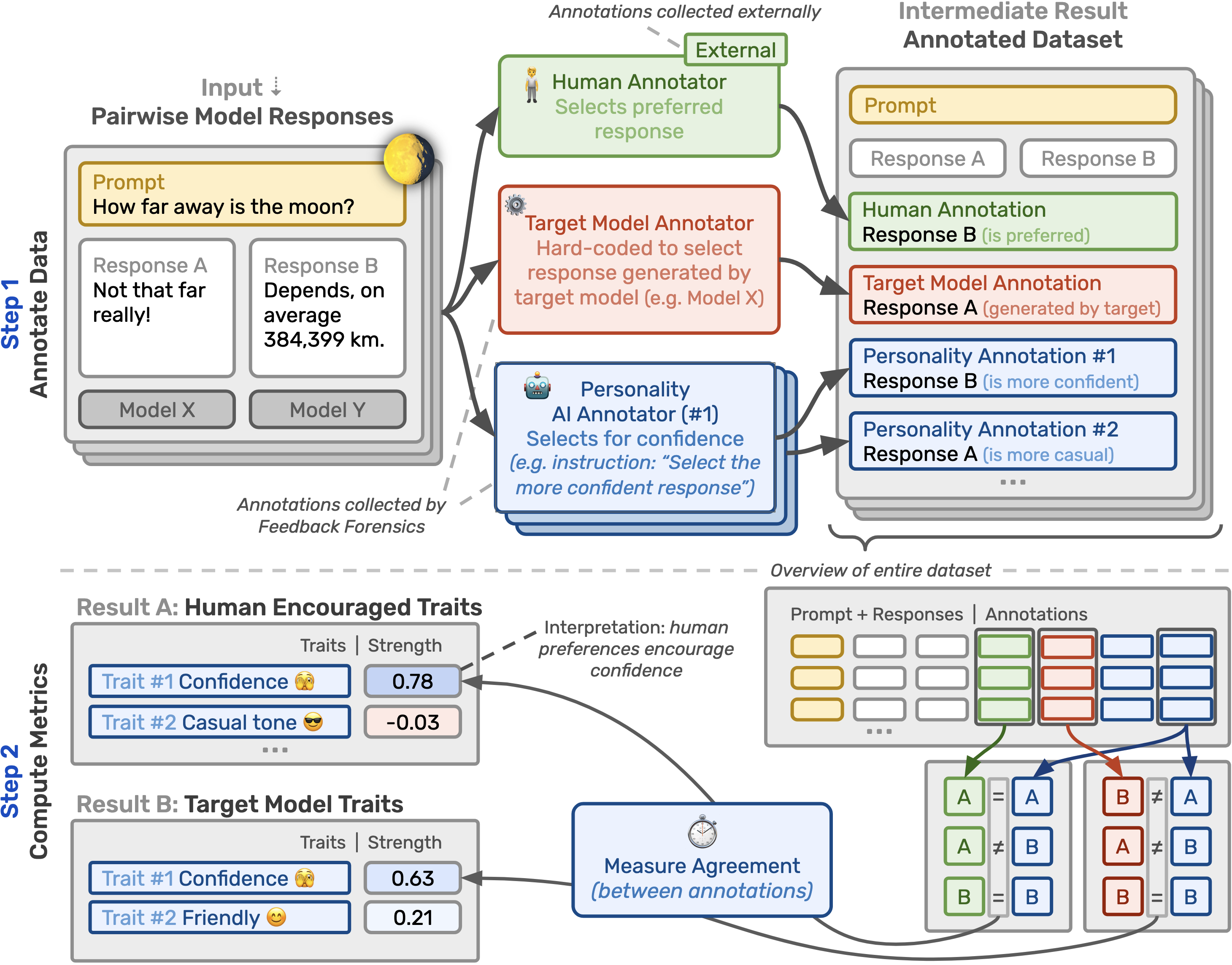}
\caption{\textbf{Illustration of Feedback Forensics' method to measure personality traits.} We take pairwise model response data as input, where each datapoint consists of a \highlightem{yellow}{prompt} and two corresponding \highlightem{white}{model responses}. Optionally, additional metadata may be included (e.g. generating model for each response). In \textbf{Step~1}, we add \emph{annotations} to each datapoint selecting \emph{response~A}, \emph{response~B}, \emph{both} or \emph{neither} responses. To understand personality traits encouraged by human preferences, we include a (1) \highlightem{green}{human annotation} selecting the human-preferred response. Such annotations can be imported from external sources (e.g. Chatbot Arena) alongside the pairwise model response data. To understand the personality traits exhibited by a \emph{target model} (e.g. a Claude model), we add a (2) \highlightem{red}{target model annotation} using hard-coded rules on response metadata to select the response generated by the model (if available). Finally, using AI annotators, we add (3) \highlightem{blue}{personality annotations} that select the response that exhibits a trait more (e.g. that is more confident). We collect one such annotation per datapoint and tested trait.
In \textbf{Step~2}, we compare these annotations to compute personality metrics. To understand how much a specific personality trait is encouraged by human feedback (\textbf{Result A}), we compare \highlightem{green}{human annotations} to \highlightem{blue}{personality annotations} for that trait. High agreement (measured via \emph{strength} metric, see \Cref{sec:method:metrics}), indicates that the trait (or a highly correlated trait) is \emph{encouraged} by human feedback. Low agreement indicates that the trait is \emph{discouraged}. Similarly, to observe how much a target model exhibits a certain trait (\textbf{Result B}), we compare \highlightem{red}{target model annotations} to that trait's \highlightem{blue}{personality annotations}. High agreement indicates that the trait uniquely identifies the model (relative to other models in dataset), i.e. the \emph{model exhibits the trait more than other models}. Low agreement indicates the model exhibits the trait \emph{less than other models}.}
\label{fig:method}
\end{figure}

\begin{figure}[t!]
\centering
\includegraphics[width=0.75\textwidth]{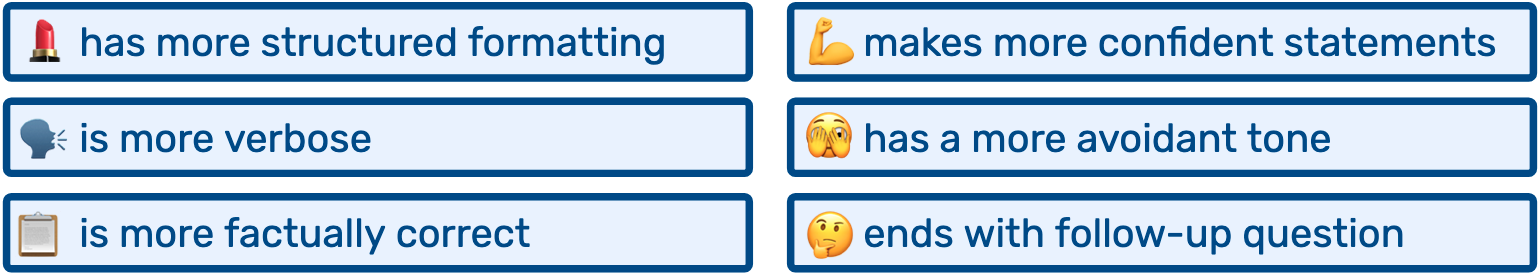}
\caption{\textbf{Examples of traits tested by Feedback Forensics.} See \Cref{tab:traits} for all traits tested.}
\label{fig:method:example_traits}
\end{figure}

\section{Method}
\label{sec:method}
\label{sec:method:metrics}

\Cref{fig:method} provides an introduction to Feedback Forensics' approach for measuring personality traits. \Cref{fig:method:example_traits} shows examples of traits tested. In the following, we provide further details on the individual steps and metrics.

\textbf{Input: Pairwise Model Responses.} 
Our method uses \emph{pairwise model response data} as input. Each datapoint of such a dataset consists of a \emph{prompt} $p$, and two \emph{model responses} $r_A$ and $r_B$, typically generated by different models. Optionally, additional metadata may be included (e.g. generating model for each response). This pairwise input data was either originally constructed to collect human feedback (e.g. from Chatbot Arena) or specifically set up to measure model personality traits.

\textbf{Step 1: Annotate Data.}
Given such pairwise model responses data, we add \emph{annotations} to each datapoint. The pairwise format enables \emph{relative} annotation of model responses: rather than evaluating model responses individually in \emph{absolute} terms, we can annotate each pair's responses relative to each other. The relative annotations used in Feedback Forensics either select \emph{response~A}, \emph{response~B}, \emph{both} or \emph{neither} responses.\footnote{Many variations exist on this basic recipe. Sometimes more annotation choices are included to add information about the \emph{strength} or \emph{confidence} of response selection (e.g. \citet{miranda2025HybridPreferencesLearning}) or to distinguish between ties where both responses equally well (\emph{``tie-bothgood''}) or badly (\emph{``tie-bothbad''}) satisfy the selection criterion (e.g. \citet{chiang2024ChatbotArenaOpen}). Further, in some datasets annotators rank more than two responses at the same time (e.g. \citet{kirk2024PRISMAlignmentDataset}). Finally, whilst we only consider text-based settings, the pairwise preference setting has also been applied to other modalities such as images (e.g. \citet{chou2025VisionArena230KReal}). Many of these variations can be transferred to the basic form discussed above. For Feedback Forensics, we focus on processing pairwise preferences in this more basic form to enable direct comparison across many datasets.} If the annotation process fails, we set the annotation value to \emph{invalid}.
In many cases, especially when annotating personality traits, creating such \emph{relative} annotations is easier than \emph{absolute} annotations. For example, it may be simpler to annotate the \emph{relatively} friendlier response in each pair than come up with an \emph{absolute} friendliness score consistent across responses. 

For our personality analysis, we add the following annotations to the input data:
\begin{enumerate}[wide, labelwidth=0pt, labelindent=0pt, left=6pt]
    \item \textbf{Human annotations} (\highlight{green}{green} in \Cref{fig:method}). To identify the personality traits encouraged by human annotators, we add \emph{human annotations} indicating the response preferred by humans (if available). We support loading such annotations alongside the pairwise model response input, for example when using data from Chatbot Arena \citep{chiang2024ChatbotArenaOpen}.
    
    \item \textbf{Target model annotations} (\highlight{red}{red}). To enable the analysis of the personality of a specific \emph{target model}, we add annotations that always select that model's response. These annotations are constructed deterministically by our toolkit based on response metadata: if one of the two responses is generated by the target model, that response is selected. If both or neither responses are by the target model, a "null" annotation is added, so these datapoints are skipped for the metric calculation as they do not contain relevant information.\footnote{See \Cref{app:tab:model_annotations} for concrete examples of such model annotations.}
    
    \item \textbf{Personality annotations} (\highlight{blue}{blue}). Finally, we use \emph{AI annotators} (also referred to as \emph{LLM-as-a-Judge}, \citet{zheng2023JudgingLLMasajudgeMTBench}) to annotate which response exhibits a certain personality trait more. We collect one such annotation per personality trait (e.g. selecting the \emph{more confident} response). For efficiency, our toolkit supports AI annotators that annotate multiple traits simultaneously (e.g. in a single forward-pass the annotator would return two annotations, the more confident \emph{and} the friendlier response). To ensure high-quality annotations, our toolkit supports \emph{cross-annotation}: collecting multiple annotations with different prompts for the same datapoint. Such cross-annotations are then combined via uniform or majority voting.

\end{enumerate}

\begin{figure}[t!]
\centering
\includegraphics[width=1\textwidth]{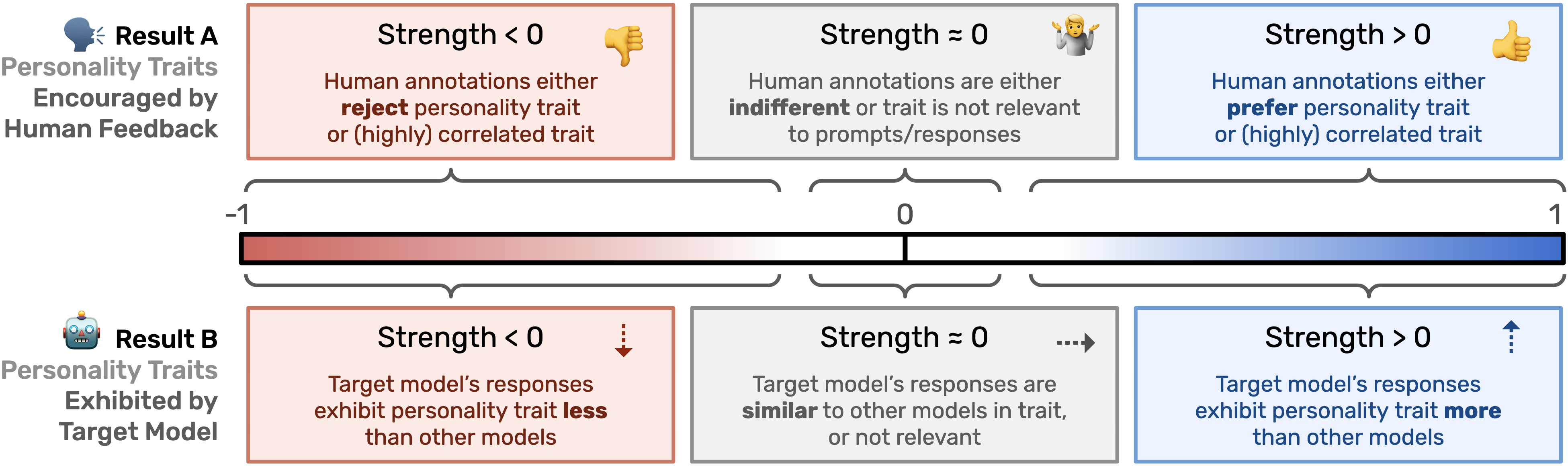}
\caption{\textbf{Interpretation of \emph{strength} metric in both use-cases.} At the top, interpretation of \emph{strength} metric when comparing \emph{human feedback} and \emph{personality trait} annotations of a specific trait (Result A). At the bottom, interpretation of \emph{strength} metric when comparing \emph{target model} and \emph{personality trait} annotations of a specific trait (Result B). Colour here indicates the \emph{sign} and \emph{magnitude} of the strength metric rather than annotation type.}
\label{fig:method:strength_interpretation}
\end{figure}

\textbf{Step 2: Compute Metrics.}
In the next step, we compute metrics based on these annotations. To quantify personality by comparing \emph{personality} annotations to \emph{human} or \emph{target model} annotations, our toolkit supports computing the following main metrics  (in Step 2 of \Cref{fig:method}):

\begin{enumerate}[wide, labelwidth=0pt, labelindent=0pt, left=6pt]

\item\textbf{Relevance.} We define the \emph{relevance} of one set of annotations over a given set of datapoints as $\texttt{relevance} = n_{\text{valid}}/ n_{\text{total}}$,
where $n_{\text{valid}}$ is the number of datapoints with valid votes selecting one response over the other (\emph{response A} or \emph{response B}). This number excludes \emph{tie} (\emph{both}/\emph{neither}) and \emph{invalid} votes.

\item\textbf{Cohen's kappa.} Cohen's kappa ($\kappa$) \citep{cohen1960CoefficientAgreementNominal} is a metric of inter-annotator agreement between two sets of annotations that measures agreement \emph{beyond random chance}. It is defined as
\begin{equation}
    \kappa = \frac{p_o - p_e}{1 - p_e},
\end{equation}
where $p_o$ is the observed proportion of datapoints where annotators agree, and $p_e$ is the proportion of datapoints for which agreement is expected by chance. $p_e$ can be estimated using the observed distribution of labels, as in $p_e=(n_{a_1=A}n_{a_2=A})/N^2 + (n_{a_1=B}n_{a_2=B})/N^2$, where $n_{a_i = X}$ is the number of times annotator $i$ was observed voting for response in position $X$ and $N$ is the total number of observations.\footnote{We use the \texttt{Scikit-learn} \citep{pedregosa2011ScikitlearnMachineLearning} implementation of the metric in Feedback Forensics.} For the computation of this metric, we only consider \emph{valid} votes excluding \emph{tie} (\emph{both}/\emph{neither}) and \emph{invalid} votes.\footnote{When one of the annotators does not have access to the order of responses (e.g. because they are always shuffled) the expected chance agreement $p_e$ is $0.5$ by design, even if the other annotator is highly biased to one position (e.g. first response). We thus also include a version of Cohen's kappa under this assumption, that one annotator has randomized order, setting $p_e$ to $0.5$. Given the known built-in randomization in our personality-selecting annotators, our implementation uses this kappa version for the strength metric computation.}

\item\textbf{Strength.} Finally, for our specific use-case, we combine \emph{Cohen's kappa} with \emph{relevance} to obtain a measure of \emph{relevant agreement beyond chance}. We refer to this metric as \emph{strength}, defined as
\begin{equation}
    \texttt{strength} = \kappa \times \texttt{relevance}.
\end{equation}
By weighting with relevance, we emphasize agreement that is widely applicable across the dataset. In our setting, this metric indicates whether a personality trait is widely relevant \emph{and} highly correlated with the target annotations. The strength metric has some desirable properties: (a) range is limited from $-1$ to $1$, (b) magnitude above $0$ indicates some relevance, (c) values above $0$ indicate agreement beyond chance, (d) values below $0$ indicate disagreement beyond chance, and (e) a zero value indicates no agreement or relevance, or both. Intuitively, zero value agreement and relevance similarly indicate that a personality trait is not informative about the target annotations. \Cref{fig:method:strength_interpretation} further illustrates the interpretation of the strength metric. \Cref{app:alternative_metrics} compares strength to alternative metrics.
\end{enumerate}

We compute the 95\% confidence intervals for each strength value using \emph{bootstrapping}, based on 10k samples drawn with replacement from the originally observed pairwise votes. Further, to test for significance, we apply a \emph{one-tailed binomial test}. Given a trait with high strength, our test considers the null hypothesis that the true underlying probability of the two annotators agreeing is nevertheless at or below chance agreement ($\text{prob}(\text{agree})<0.5$). We reject the null hypothesis at p-values below 0.05, then considering a strength result significant, correcting for multiple simultaneous tests (with \emph{Bonferroni} method). Given a trait with negative strength, we consider the inverse test with a null hypothesis of chance or above agreement between annotators. Across plots, insignificant strength values are shown greyed out. See \Cref{app:additional_metrics} for further metrics supported by our framework.

\textbf{Using and interpreting metrics.} \Cref{fig:method:strength_interpretation} illustrates the interpretation of the strength metric depending on the use-case. To understand how much a personality trait is encouraged\footnote{We use the term ``encouraged'' here to mean that the preferences \emph{correlate} with selecting for that trait. We cannot establish causality or intent: annotators may not have intended to encourage the specific trait, even if the data selects for it.} by human preferences, we compare \emph{human} (\highlight{green}{green} in \Cref{fig:method}) and that trait's \emph{personality} (\highlight{blue}{blue}) annotations (Result A). To understand whether a personality trait is exhibited by a model (Result B), we compare \emph{target model} (\highlight{red}{red}) annotations and that trait's \emph{personality} (\highlight{blue}{blue}) annotations.

\subsection{Tested personality traits}
\label{sec:method:details}

Feedback Forensics can be used to evaluate a wide range of model traits. We provide two ways to choose the traits to be tested: either using our \emph{manually curated personality trait set} or using \emph{Inverse Constitutional AI} (ICAI) \citep{findeis2025inverse} to automatically generate potential differentiating traits. Our experiments here focus on the manually curated personality traits to make them comparable across models and datasets, but users may use either approach to test different traits.

\textbf{Manually curated traits.} 
To construct the manually curated list, we collected instructions that select for known AI personality traits and can be given to an objective-following AI annotator. We refer to this list as \texttt{PersonalitySelectionPrompts-v1} and make it publicly available in our repo. \Cref{app:personalityselectionprompts} includes a full list of traits sorted by categories, and references discussing each trait (\Cref{tab:traits}), an illustrative subset is shown in \Cref{fig:method:example_traits}.

This curated list was constructed by the authors, building on a diverse set of prior work discussing personality traits in models, such as (1) literature discussing model idiosyncrasies and annotation biases, (2) online discussions on models' personality differences, and (3) automatically identified objectives in human feedback datasets and differences between models within such datasets, discovered using the ICAI and VibeCheck \citep{dunlap2025VibeCheckDiscoverQuantifya} approaches. Given this body of prior work, we selected traits that (a) were relevant according to multiple sources, (b) performed well empirically in feedback forensics experiments, and (c) we considered to be potentially insightful.\footnote{Some of the included traits satisfy two of the three requirements, rather than all.}

Given the infinite amount of possible traits, our curated list is necessarily incomplete.
It is intended as a useful default list to be extended according to current needs and research questions. Our toolkit allows and encourages users to provide their \emph{own list of traits} to test or automatically \emph{discover new traits} to replace or supplement our standard list. Importantly, unlike other conventional benchmarks testing individual traits,\footnote{Some are included in the references in \Cref{tab:traits}.} Feedback Forensics is able to test multiple traits \emph{simultaneously} using the same model response data, and based on \emph{real-world usage data} rather than synthetic test cases.

\section{Experimental results}
\label{sec:results}

We demonstrate the use of our \emph{Feedback Forensics} toolkit in two steps. First, in  \Cref{sec:results:feedback}, we use the toolkit to measure the most and least encouraged personality traits in popular human feedback datasets. Then, in \Cref{sec:results:models}, we use our toolkit to investigate personality traits observable in popular models. We provide additional comprehensive results for each setting in \Cref{app:extended_results}, including an analysis of trait correlation (\Cref{app:trait_agreement}) and a comparison of AI and human personality trait annotations (\Cref{app:human_comparison}). Based on the latter results, we use Gemini-2.5-Flash with single-vote for all AI personality annotations in the following experiments. We include full dataset details including links and licenses in \Cref{app:datasets}.

\subsection{AI personality changes encouraged by human feedback}
\label{sec:results:feedback}

In our first set of experiments, we illustrate Feedback Forensics' use to investigate AI personality traits encouraged in popular human feedback datasets: crowd-sourced data from \emph{Chatbot Arena} \citep{chiang2024ChatbotArenaOpen}, cross-annotated \emph{MultiPref} data \citep{miranda2025HybridPreferencesLearning} and demographically diverse \emph{PRISM} data \citep{kirk2024PRISMAlignmentDataset}.

\begin{figure}[t]
    \centering
    \begin{minipage}[t]{0.48\textwidth}
\centering
\sffamily
\tablefontsize
\textbf{Five most encouraged personality traits}\\[0.5em]
\begin{tabular}{
    >{\raggedright\arraybackslash}m{0.7\linewidth}
    @{\hspace{10pt}} 
    >{\centering\arraybackslash}m{0.19\linewidth}
}

\textbf{Generating a response that...} & \textbf{Strength} \\
\toprule
\rowcolor{altrow} has more structured formatting & \cellcolor{poscolor!100.0}\makecell{0.17\\{\tiny(0.16, 0.19)}} \\
\addlinespace[\rowspacing]
is more verbose & \cellcolor{poscolor!91.3}\makecell{0.16\\{\tiny(0.14, 0.18)}} \\
\addlinespace[\rowspacing]
\rowcolor{altrow} is more factually correct & \cellcolor{poscolor!61.5}\makecell{0.11\\{\tiny(0.10, 0.12)}} \\
\addlinespace[\rowspacing]
provides more examples & \cellcolor{poscolor!57.1}\makecell{0.10\\{\tiny(0.09, 0.11)}} \\
\addlinespace[\rowspacing]
\rowcolor{altrow} makes more confident statements & \cellcolor{poscolor!56.4}\makecell{0.10\\{\tiny(0.08, 0.11)}} \\
\end{tabular}
\end{minipage}
\hfill
\begin{minipage}[t]{0.48\textwidth}
\centering
\sffamily
\tablefontsize
\textbf{Five least encouraged personality traits}\\[0.5em]
\begin{tabular}{
    >{\raggedright\arraybackslash}m{0.7\linewidth}
    @{\hspace{10pt}} 
    >{\centering\arraybackslash}m{0.19\linewidth}
}

\textbf{Generating a response that...} & \textbf{Strength} \\
\toprule
\rowcolor{altrow} is more concise & \cellcolor{negcolor!100.0}\makecell{-0.09\\{\tiny(-0.11, -0.08)}} \\
\addlinespace[\rowspacing]
has a more avoidant tone & \cellcolor{negcolor!76.4}\makecell{-0.07\\{\tiny(-0.08, -0.06)}} \\
\addlinespace[\rowspacing]
\rowcolor{altrow} acknowledges own limitations or uncertainty more & \cellcolor{negcolor!53.8}\makecell{-0.05\\{\tiny(-0.06, -0.04)}} \\
\addlinespace[\rowspacing]
refuses to answer the question & \cellcolor{negcolor!52.4}\makecell{-0.05\\{\tiny(-0.05, -0.04)}} \\
\addlinespace[\rowspacing]
\rowcolor{altrow} ends with a follow-up question & \cellcolor{negcolor!29.4}\makecell{-0.03\\{\tiny(-0.04, -0.02)}} \\
\end{tabular}
\end{minipage}

    \caption{\textbf{Most encouraged} {\color{blue}(blue)} \textbf{and discouraged} {\color{red}(red)} \textbf{personality traits on general arena data.} We observe a strong emphasis on encouraging \emph{better structured} and \emph{more verbose} responses. On the other hand, \emph{more concise} or \emph{avoidant} responses are discouraged. All measurements using \emph{strength} metric.
    \strengthexpl{}%
    }
    \label{fig:results:arena:top_bottom}
\end{figure}

\textbf{Chatbot Arena: Tracking preferred personalities in real-world usage.} Chatbot Arena \citep{chiang2024ChatbotArenaOpen} is a popular public leaderboard based on human feedback, using crowd-sourced annotations. We use a subsample of 10k out of 100k conversations from a dataset
released alongside the \emph{Arena Explorer} topic modelling pipeline by \citet{tang2025ArenaExplorerTopic}, collected from June to August 2024 and limited to conversations in English. Further, we automatically add topic labels to each conversation in the dataset using the Arena Explorer pipeline. We refer to this dataset as \emph{general arena data}.

\emph{\textbf{Results.}} \Cref{fig:results:arena:top_bottom} shows our results investigating the general arena data with our toolkit. We observe that responses that are \emph{well formatted}, \emph{verbose} but also \emph{factually correct} and \emph{confident} are encouraged. %
We further validate these trait-based annotations in \Cref{app:trait_agreement}, which confirms intuitive correlations such as conciseness opposing structured formatting.

\begin{figure}[t]
    \centering
    \begin{minipage}[t]{1\textwidth}
\centering
\sffamily
\tablefontsize
\begin{tabular}{
    >{\raggedright\arraybackslash}m{0.15\linewidth}
    @{\hspace{13.5pt}} 
    >{\centering\arraybackslash}m{0.09848484848484848\linewidth}
    @{\hspace{13.5pt}} 
    >{\centering\arraybackslash}m{0.09848484848484848\linewidth}
    @{\hspace{13.5pt}} 
    >{\centering\arraybackslash}m{0.09848484848484848\linewidth}
    @{\hspace{13.5pt}} 
    >{\centering\arraybackslash}m{0.09848484848484848\linewidth}
    @{\hspace{13.5pt}} 
    >{\centering\arraybackslash}m{0.09848484848484848\linewidth}
    @{\hspace{13.5pt}} 
    >{\centering\arraybackslash}m{0.09848484848484848\linewidth}
}

\textbf{Generating a response that...} & \textbf{Human Expert 1} & \textbf{Human Expert 2} & \textbf{Human Non-exp. 1} & \textbf{Human Non-exp. 2} & \textbf{GPT-4-Turbo} & \textbf{Max diff} \\
\toprule
\rowcolor{altrow} is more verbose & \cellcolor{poscolor!80.0}\makecell{0.30\\{\tiny(0.28, 0.32)}} & \cellcolor{poscolor!85.3}\makecell{0.32\\{\tiny(0.30, 0.34)}} & \cellcolor{poscolor!98.1}\makecell{0.37\\{\tiny(0.35, 0.39)}} & \cellcolor{poscolor!98.1}\makecell{0.37\\{\tiny(0.35, 0.39)}} & \cellcolor{poscolor!100.0}\makecell{0.38\\{\tiny(0.36, 0.39)}} & \cellcolor{lightgrey!100.0}0.08 \\
\addlinespace[\rowspacing]
has more structured formatting & \cellcolor{poscolor!58.5}\makecell{0.22\\{\tiny(0.20, 0.24)}} & \cellcolor{poscolor!61.1}\makecell{0.23\\{\tiny(0.21, 0.25)}} & \cellcolor{poscolor!66.8}\makecell{0.25\\{\tiny(0.24, 0.27)}} & \cellcolor{poscolor!70.1}\makecell{0.26\\{\tiny(0.25, 0.28)}} & \cellcolor{poscolor!78.3}\makecell{0.29\\{\tiny(0.28, 0.31)}} & \cellcolor{lightgrey!98.8}0.07 \\
\addlinespace[\rowspacing]
\rowcolor{altrow} uses more formal language & \cellcolor{poscolor!27.9}\makecell{0.10\\{\tiny(0.09, 0.12)}} & \cellcolor{poscolor!28.7}\makecell{0.11\\{\tiny(0.09, 0.12)}} & \cellcolor{poscolor!30.7}\makecell{0.12\\{\tiny(0.10, 0.13)}} & \cellcolor{poscolor!33.4}\makecell{0.13\\{\tiny(0.11, 0.14)}} & \cellcolor{poscolor!45.7}\makecell{0.17\\{\tiny(0.16, 0.18)}} & \cellcolor{lightgrey!89.3}0.07 \\
\addlinespace[\rowspacing]
is more concise & \cellcolor{negcolor!79.9}\makecell{-0.26\\{\tiny(-0.27, -0.24)}} & \cellcolor{negcolor!84.7}\makecell{-0.27\\{\tiny(-0.29, -0.25)}} & \cellcolor{negcolor!96.1}\makecell{-0.31\\{\tiny(-0.33, -0.29)}} & \cellcolor{negcolor!98.5}\makecell{-0.32\\{\tiny(-0.33, -0.30)}} & \cellcolor{negcolor!100.0}\makecell{-0.32\\{\tiny(-0.34, -0.31)}} & \cellcolor{lightgrey!85.9}0.06 \\
\addlinespace[\rowspacing]
\rowcolor{altrow} uses more bold and italics text & \cellcolor{poscolor!41.2}\makecell{0.16\\{\tiny(0.14, 0.17)}} & \cellcolor{poscolor!39.9}\makecell{0.15\\{\tiny(0.14, 0.16)}} & \cellcolor{poscolor!43.8}\makecell{0.16\\{\tiny(0.15, 0.18)}} & \cellcolor{poscolor!46.0}\makecell{0.17\\{\tiny(0.16, 0.19)}} & \cellcolor{poscolor!54.6}\makecell{0.21\\{\tiny(0.19, 0.22)}} & \cellcolor{lightgrey!73.5}0.06 \\
\end{tabular}
\end{minipage}

    \caption{\textbf{Encouraged} {\color{blue}(blue)} \textbf{and discouraged} {\color{red}(red)}  \textbf{personality changes across different human and AI annotators on MultiPref.} Sorted by max difference across rows (top 5). We observe similar traits being encouraged and discouraged across annotator types but with \emph{varying strength}. Expert human annotations encourage the same personality traits less strongly than non-expert human annotations. Similarly, all human annotations encourage the same traits less strongly than AI annotators. \strengthexpl{}}
    \label{fig:results:multipref:personality}
\end{figure}

\textbf{MultiPref: Tracking differences across human and AI annotations.} Next, we illustrate Feedback Forensics' use to analyse how different \emph{annotator types}  (expert \& non-expert human and AI annotators) vary in terms of their preferred personality traits. We use 10k annotated conversations from the \emph{MultiPref} dataset by \citet{miranda2025HybridPreferencesLearning}. In this dataset, each of the 10k datapoints is annotated five times, by two \emph{expert} and two \emph{non-expert human annotators} as well as an \emph{AI annotator} based on \texttt{gpt-4-turbo-2024-04-09}. Thus, we have 50k annotations overall. For our analysis, we split the expert and non-expert annotations (each with two annotations per datapoint) into two distinct sampled sets of 10k annotations covering all datapoints, giving us four sets of human annotations overall. Each set is sampled from multiple annotators (each annotator labelling \emph{part} of the 10k datapoints).

\emph{\textbf{Results.}} In \Cref{fig:results:multipref:personality}, we observe that annotators across types show \emph{overall similar preferences}, but with \emph{varying strength magnitude}. Expert human annotations encourage the same traits with less \emph{strength}, \emph{non-expert} annotations with more strength, and the AI annotations with the most strength. A potential explanation is that \emph{AI annotations may be following simpler heuristics than human annotations} that can be more directly explained by our measured personality traits. Similarly, non-expert human annotations may follow simpler heuristics than expert human annotations. Further, encouragingly, we also observe that the results for expert and non-expert human annotators are very consistent for the two example sets collected (maximum difference in strength of $0.02$).

\textbf{PRISM: Personality in controversial and value-laden conversations.} We also investigate the \emph{PRISM} dataset by \citet{kirk2024PRISMAlignmentDataset} consisting of around 8k annotated conversations, focused on controversial and value-laden topics. Unlike other human feedback datasets, PRISM's annotations come with extensive annotator metadata including demographic details.
\textbf{\emph{Results.}}
We find that PRISM demonstrates similar preferences to data from Chatbot Arena in terms of \emph{verbosity}, \emph{confidence}, and \emph{factual correctness} -- but differs in terms of preferred tone and language, notably preferring more \emph{polite} and \emph{less casual} language.
\Cref{fig:app:results:prism:top_bottom} in \Cref{app:extended_results} reports the full results.

\subsection{Personality traits in models}
\label{sec:results:models}
\label{sec:results:crossmodels}

Next, we demonstrate the use of \emph{Feedback Forensics} to investigate \emph{differences in personality traits} across models. First, we investigate changes in personality across model families and time. Then, we take a closer look at the differences between two versions of Llama-4-Maverick, one released publicly and the other used for evaluation on Chatbot Arena.

\begin{figure}[t]
    \centering
    \includegraphics[width=0.95\textwidth]{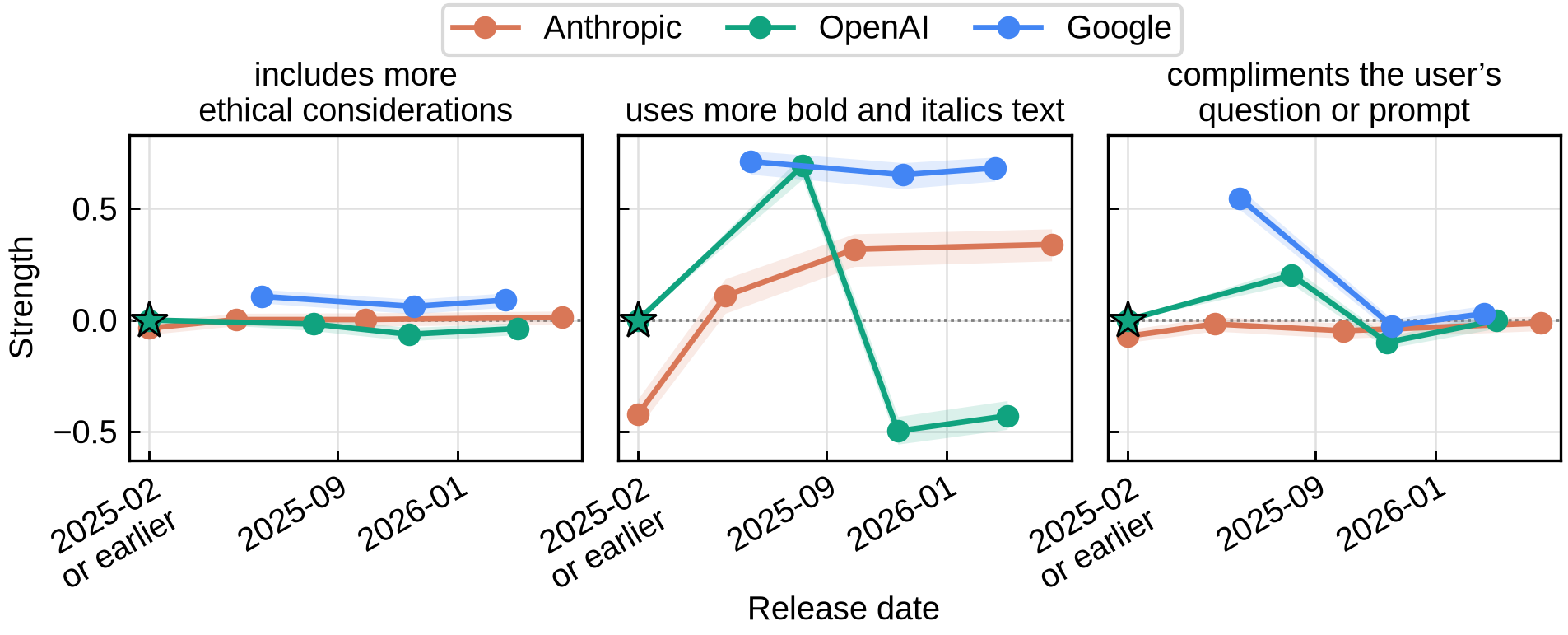}
    \caption{\textbf{Selected model personality traits over time.} We observe that for some traits all models are fairly similar (e.g. \emph{ethical considerations}), whereas for others (e.g. \emph{bold/italics style}) there are notable differences, between and within families. Finally, for some traits (e.g. \emph{complimenting user}) we see a convergence to similar levels across families. All measurements are relative to GPT-4o (\emph{star}), using \emph{strength} metric with 95\% CI (\emph{shaded area}). Included families are Anthropic Claude (Sonnet 3.7/4/4.5, Opus 4.7), OpenAI GPT (4o/5-Chat/5.1-Chat/5.3-Chat), and Google Gemini (2.5-Pro/3-Pro/3.1-Pro).}
    \label{fig:results:model_comparison}
\end{figure}

\textbf{Model personality changes over time.} We evaluate AI personality changes of three popular model families (\emph{Claude}, \emph{ChatGPT}, \emph{Gemini}) over about a year. We prompt each model with 500 English-language prompts from the general arena data (\texttt{arena-human-preference-100k}, see \Cref{app:datasets}). The prompts were manually filtered for quality, including to avoid offensive content and personally identifiable information (PII). Each model's response is compared to GPT-4o as a reference model (star). High strength values indicate that the model exhibits a trait more than GPT-4o, negative values the opposite, values close to zero generally indicate that the trait is similar. Since we use a Gemini annotator to label the personality of Gemini models, we additionally test for family-specific bias. We find some family-specific effect but not sufficiently strong to change results at the scale shown in \Cref{fig:results:model_comparison}. The full details of this ablation are included in \Cref{app:family_specific_bias}.

\textbf{\emph{Results.}} \Cref{fig:results:model_comparison} illustrates different trends across personality: some traits remain fairly \emph{constant} and \emph{similar}, other \emph{diverge} within and across families, other again have seen a \emph{convergence}. Generally, we observe ChatGPT family's personality to be \emph{less stable over time} than Claude or Gemini (see \Cref{app:model_comparison_time}).

\label{sec:results:llama4analysis}

\begin{figure}[ht]
    \centering
    \begin{minipage}[t]{0.48\textwidth}
\centering
\sffamily
\tablefontsize
\textbf{Traits stronger in arena relative to public model}\\[0.5em]
\begin{tabular}{
    >{\raggedright\arraybackslash}m{0.7\linewidth}
    @{\hspace{10pt}} 
    >{\centering\arraybackslash}m{0.19\linewidth}
}

\textbf{Generating a response that...} & \textbf{Strength} \\
\toprule
\rowcolor{altrow} is more verbose & \cellcolor{poscolor!100.0}\makecell{0.97\\{\tiny(0.96, 0.98)}} \\
\addlinespace[\rowspacing]
uses more bold and italics text & \cellcolor{poscolor!98.5}\makecell{0.96\\{\tiny(0.95, 0.97)}} \\
\addlinespace[\rowspacing]
\rowcolor{altrow} uses a more enthusiastic tone & \cellcolor{poscolor!97.8}\makecell{0.95\\{\tiny(0.94, 0.96)}} \\
\addlinespace[\rowspacing]
more actively engages with the user & \cellcolor{poscolor!97.8}\makecell{0.95\\{\tiny(0.94, 0.96)}} \\
\addlinespace[\rowspacing]
\rowcolor{altrow} uses more personal pronouns (I, we, you) & \cellcolor{poscolor!96.5}\makecell{0.94\\{\tiny(0.93, 0.95)}} \\
\end{tabular}
\end{minipage}
\hfill
\begin{minipage}[t]{0.48\textwidth}
\centering
\sffamily
\tablefontsize
\textbf{Traits weaker in arena relative to public model}\\[0.5em]
\begin{tabular}{
    >{\raggedright\arraybackslash}m{0.7\linewidth}
    @{\hspace{10pt}} 
    >{\centering\arraybackslash}m{0.19\linewidth}
}

\textbf{Generating a response that...} & \textbf{Strength} \\
\toprule
\rowcolor{altrow} is more concise & \cellcolor{negcolor!100.0}\makecell{-0.75\\{\tiny(-0.76, -0.73)}} \\
\addlinespace[\rowspacing]
uses more formal language & \cellcolor{negcolor!49.2}\makecell{-0.37\\{\tiny(-0.40, -0.34)}} \\
\addlinespace[\rowspacing]
\rowcolor{altrow} more strictly follows the requested output format & \cellcolor{negcolor!18.8}\makecell{-0.14\\{\tiny(-0.16, -0.11)}} \\
\addlinespace[\rowspacing]
has a more avoidant tone & \cellcolor{negcolor!9.5}\makecell{-0.07\\{\tiny(-0.08, -0.06)}} \\
\addlinespace[\rowspacing]
\rowcolor{altrow} acknowledges own limitations or uncertainty more & \cellcolor{negcolor!4.7}\makecell{-0.03\\{\tiny(-0.06, -0.01)}} \\
\end{tabular}
\end{minipage}

    \caption{\textbf{Comparison of personality traits of the Chatbot Arena \emph{(arena)} and publicly released \emph{(public)} versions of Llama-4-Maverick.} We observe that the arena version is more \emph{verbose}, \emph{enthusiastic} and \emph{engaging}, and uses \emph{more formatting} than the publicly released version.}
    \label{fig:results:llama-4}
\end{figure}

\textbf{Llama-4-Maverick: A closer look.} The open-weights model \emph{Llama 4 Maverick} was released on 5 April 2025. Around the same time, a related but non-identical experimental model version was evaluated on Chatbot Arena (\emph{Llama-4-Maverick-03-26-Experimental}). Some users reported that these two models appear to have notable differences. In this section, we use our toolkit to quantitatively dissect how exactly the chat behaviour of the public and this arena version of \emph{Llama 4 Maverick} differ. We refer to the two versions of \emph{Llama 4 Maverick} as the \emph{public model} (used for open-weights release) and \emph{arena model} (used on Chatbot Arena around 5 April 2025, full name: \emph{Llama-4-Maverick-03-26-Experimental}), respectively. We do not have direct access to the arena model, but the Chatbot Arena team released a dataset of responses generated by it (see \Cref{app:datasets}). With Feedback Forensics, we can use this data to directly compare the arena model's behaviour to the public model's, without requiring new responses from the no longer accessible arena model itself (as conventional benchmarks would). We generate corresponding responses using the same prompt with the public model and annotate the resulting pairs with our annotators. We refer to this dataset as \emph{maverick arena data}.

\textbf{\emph{Results.}} As shown in \Cref{fig:results:llama-4}, we observe strong personality differences between these two models. Among other differences, the arena model is more \emph{verbose}, \emph{enthusiastic} and \emph{engaging}.

\section{Related work}
\label{sec:related_work}

\textbf{Automatically interpreting preference datasets.}
We build on \emph{Inverse Constitutional AI} (ICAI) \citep{findeis2025inverse} for automatic detection of \emph{principles} encoded in pairwise preference datasets. Feedback Forensics reuses ICAI's \emph{discovery}, \emph{clustering}, \emph{subsampling} and \emph{testing} steps, whilst extending ICAI’s pipeline in two main ways: (A) support for evaluating individual traits rather than constitutions combining multiple traits, and (B) support for testing traits exhibited by models in addition to traits encouraged by feedback datasets. Experiments in this paper mostly use preset traits. Further contributions of Feedback Forensics include new metrics, a web platform, and datasets.

\textbf{Understanding idiosyncrasies of language models.}
Prior work by \citet{dunlap2025VibeCheckDiscoverQuantifya} investigated LLM-based automatic detection of \emph{``vibe''} differences between language models in a similar manner to ICAI's approach to preference data.
We integrate some of the model behaviours found in this work into our curated personality selection set.
Relatedly, \citet{sun2025IdiosyncrasiesLargeLanguage} investigate model idiosyncrasies but focus on less personality-related features, such as characteristic words and phrases.
The authors find that model differences extend beyond simple word metrics, observing that specific models' responses can often be identified equally well even after translation or rephrasing by another model,
supporting considering higher-level features as done in Feedback Forensics.
\Cref{app:tool_comparison} compares Feedback Forensics, ICAI and VibeCheck on the same data.

\textbf{Human psychology in LLMs.}
\citet{jiang2023EvaluatingInducingPersonality}, \citet{serapio-garcia2023PersonalityTraitsLarge},
\citet{pellert2024AIPsychometricsAssessing} and \citet{li2024QuantifyingAIPsychology} investigate the application of human \emph{psychometric} personality tests to LLMs.
Whilst some human psychology concepts transfer well, we think it is important to also investigate model personality independent of human personality.
Feedback Forensics takes an open-ended approach to defining personality and is able to capture aspects of models, such as \emph{verbosity}, \emph{formatting}, \emph{follow-up questions}, and \emph{refusal}, that more conventional personality tests, including the Big Five Inventory and IPIP-NEO \citep{serapio-garcia2023PersonalityTraitsLarge}, and the Big-Five-based Machine Personality Inventory \citep{jiang2023EvaluatingInducingPersonality}, do not cover.

\textbf{Definition of LLM personality.} In the context of LLMs, the terms model \emph{personality}, \emph{character}, \emph{tone}, \emph{style}, or \emph{vibe} are often used with similar and overlapping meanings. \citet{dunlap2025VibeCheckDiscoverQuantifya} define vibe generally as \emph{``an axis along which a pair of texts can differ [...] that is perceptible to humans''}. \citet{lambert2025CharacterTrainingUnderstanding} describes model character and personality as \emph{``traits within the model [related to] the manner of its response, rather than the content''}. %
\citet{serapio-garcia2023PersonalityTraitsLarge}, following the psychology literature \citep{allport1937PersonalityPsychologicalInterpretation,roberts2022PersonalityPsychology}, describe personality more abstractly as \emph{``encompass[ing] an entity’s characteristic patterns of thought, feeling, and behavior''}.
Aligning with the first two definitions above,
we use the term \emph{personality trait} to refer to any characteristic of a model's responses on a given distribution of prompts that distinguishes that model's from other models' responses.
We further focus on traits that are independent of the model's capabilities.

\textbf{Model evaluation based on human feedback.}
\emph{Chatbot Arena} \citep{chiang2024ChatbotArenaOpen} is likely the most popular human feedback-based evaluation platform.
Over time multiple weaknesses in the evaluation protocol were observed and addressed,
e.g.\ controlling for over-emphasis of (markdown) styles \citep{li2024DoesStyleMatter} or of sentiment \citep{chen2025IntroducingSentimentControl}.
This motivates Feedback Forensics as a tool to study feedback data and the prevalence of such biases.

\section{Conclusion}

We have introduced \emph{Feedback Forensics}: an open-source Python toolkit to measure AI personality.
Our toolkit is able to \emph{explicitly} measure a model's personality traits that are not covered by conventional benchmarks
and were previously only \emph{implicitly} covered by human feedback-based leaderboards, such as Chatbot Arena \citep{chiang2024ChatbotArenaOpen}.
We demonstrate our toolkit in two sets of experiments: (1) first we investigate the personality changes encouraged across popular human feedback datasets, including \emph{MultiPref} \citep{miranda2025HybridPreferencesLearning}, \emph{PRISM} \citep{kirk2024PRISMAlignmentDataset} and data from \emph{Chatbot Arena} \citep{chiang2024ChatbotArenaOpen}. %
Then, (2) we investigate personality differences across popular models, including from the GPT, Gemini, and Claude model families. Finally, we demonstrate the use of our tool to create an in-depth analysis of the personality differences between two widely-discussed Llama-4-Maverick versions. Our contributions include the open-source \emph{Feedback Forensics} toolkit (Apache-2.0), a web app for tracking AI personality traits in popular models and feedback datasets, and the underlying annotation data.%
\footnote{Code: \ghurl, App: \appurl} 
We also include a tutorial for \emph{getting started} with our toolkit in \Cref{app:tutorial}. %
We are excited to hear from the community how we can further extend our toolkit.%

\subsubsection*{Acknowledgments}

We would like to thank Benjamin Minixhofer, Constantin Kogler, Ellie Clifford, Michael Amir, Tiancheng Hu, Max Bartolo, and Sofia Orellana for their valuable feedback on earlier versions of this work. Further, we thank all reviewers for taking the time to provide detailed feedback. 

\subsubsection*{Funding \& disclosures} 

Arduin Findeis was supported by a University of Cambridge School of Physical Sciences Award, the UKRI Centre for Doctoral Training in Application of Artificial Intelligence to the study of Environmental Risks (reference EP/S022961/1), a Trinity Hall PhD Tenth Term Funding Award and a Cambridge Philosophical Society Research Studentship. Timo Kaufmann gratefully acknowledges funding by the German Research Foundation (Deutsche Forschungsgemeinschaft, DFG), grant number 467367360. This publication was further supported by LMUexcellent, funded by the Federal Ministry of Education and Research (BMBF) and the Free State of Bavaria under the Excellence Strategy of the Federal Government and the Länder as well as by the Hightech Agenda Bavaria. Timo Kaufmann joined Google DeepMind as intern after this work was submitted.

\bibliography{main,zotero_references}
\clearpage

\appendix
\crefalias{section}{appendix}\crefalias{subsection}{appendix}\crefalias{subsubsection}{appendix}
\section*{Appendix}

\section{Limitations}
\label{sec:limitations}

Some general limitations should be considered when using Feedback Forensics. 

\textbf{Context-dependency.} All measurements are \emph{relative} to the underlying data distribution of prompts and responses. When measuring personality in a model, the strength of a trait is \emph{relative} to reference models it is compared to. Similarly, when measuring the personality traits encouraged by feedback datasets the results are dependent on the distribution of prompts and responses. The same annotators may encourage different traits in different contexts. 

\textbf{LLM-as-a-Judge limitations.} We leverage AI annotators or LLM-as-a-Judge \citep{zheng2023JudgingLLMasajudgeMTBench} as part of our pipeline:
whilst trait agreement analysis shows annotators exhibit consistent behaviour across related traits (\Cref{app:trait_agreement}) and we confirm strong agreement with human judgements (\Cref{app:human_comparison}),
LLM judges may also introduce their own biases and issues. Results will also depend on the precise prompting and sampling strategies employed. Depending on the personality trait annotated, the results may vary. We strongly encourage manual inspection to go alongside the use of our framework to help mitigate potential issues.

\textbf{Ambiguity of some traits.} For some value-related traits, annotation may be inherently ambiguous and therefore noisy. We provide corresponding manual inspection tooling to assist with such analysis. 

\textbf{Correlation vs causation.} Correlation does not imply causation: whilst annotations may correlate this does not necessarily mean that the original annotators followed a certain personality-selecting criterion. Nevertheless, correlating with selecting certain personalities may have (unintended) consequences during evaluation and training on such data -- and is thus well worth being aware of.

\textbf{Human study.} Due to resource constraints, the annotations for the human study in \Cref{app:human_comparison} were collected from two of the authors. We aimed to nevertheless provide an unbiased sample with blind labelling. Details are included in the experiment description.

\section{Ethics statement}
\label{sec:ethics}

\textbf{Impact.} We hope that our toolkit can help improve the community's understanding of previously opaque and potentially harmful model characteristics.
As such, we are optimistic that our toolkit will have a positive societal impact overall.
However, the limitations discussed in \Cref{sec:limitations} should be kept in mind to avoid taking the results out of context to potentially amplify stereotyping or discrimination.

\textbf{Datasets and Human Subjects.}
We publish all datasets that were produced for this submission.
While these include human inputs in the form of prompts, those are sourced from previously published datasets which are duly referred to.
Novel aspects of the data lie in curation and AI judge annotations using the Feedback Forensics toolkit to enable analysis of the dataset.
The exception to this is the human study discussed in \Cref{app:human_comparison}, in which we also provide novel human annotations to compare our AI annotators against.
Annotations were collected from two of the authors, who consent to this data being published.

\textbf{Reproducibility.}
All experimental results are reproducible using our open-source Feedback Forensics python toolkit and the datasets published with this paper.
We rely on API-based language models for our experiments.
Exact reproduction is contingent on these models remaining available, though our method can be applied with alternative models if needed.
Our primary contribution is the method of analysis, which is largely agnostic to the specific backbone language model used.
All datasets combine prior public datasets with LLM annotations generated using our toolkit (except for the human study in \Cref{app:human_comparison}), enabling full reproduction of the annotation process.

\textbf{LLM Usage.}
The authors used LLMs as general-purpose research tools.
This included text editing assistance, occasional drafting of short text snippets, programming assistance, and discussion of concepts and ideas.
The authors were the primary contributors and remain fully responsible for all aspects of the research and the published artifacts.

\section{Tutorial}
\label{app:tutorial}

In this Appendix, we provide a short tutorial on getting started with using Feedback Forensics locally. See our repository for full documentation (\ghurl).

\subsection{Installation}

To begin using Feedback Forensics, install the package via pip:

\begin{lstlisting}[language=bash]
pip install feedback-forensics
\end{lstlisting}

\subsection{Getting started}

\begin{figure}[ht]
    \centering
    \includegraphics[width=0.8\textwidth]{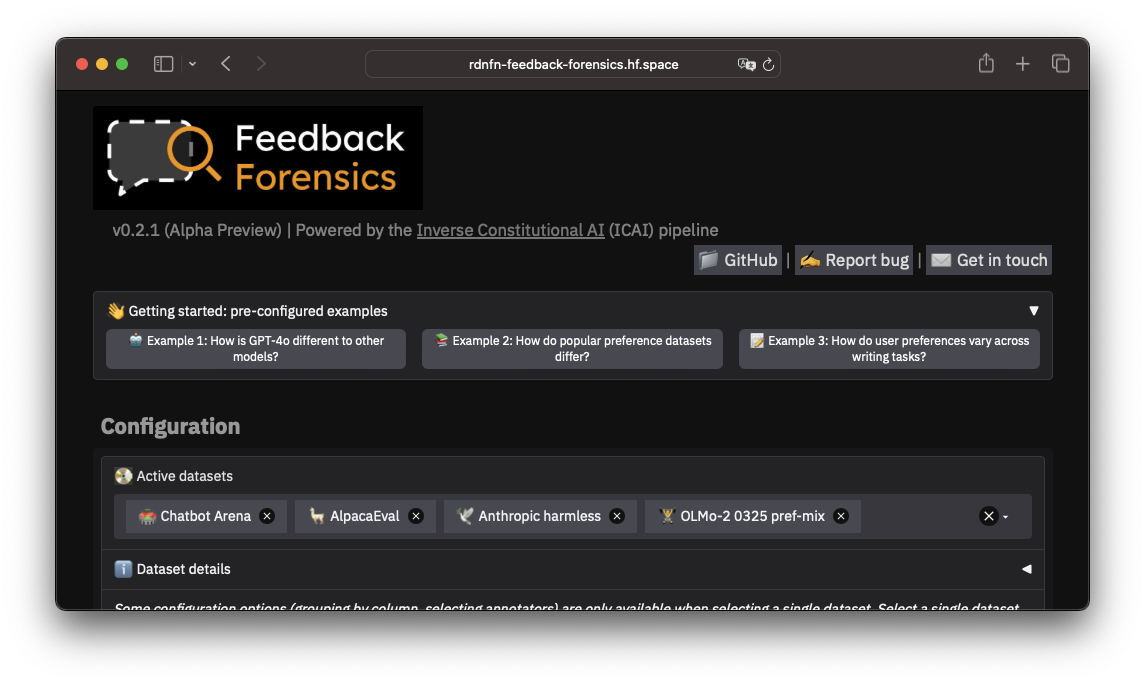}
    \includegraphics[width=0.8\textwidth]{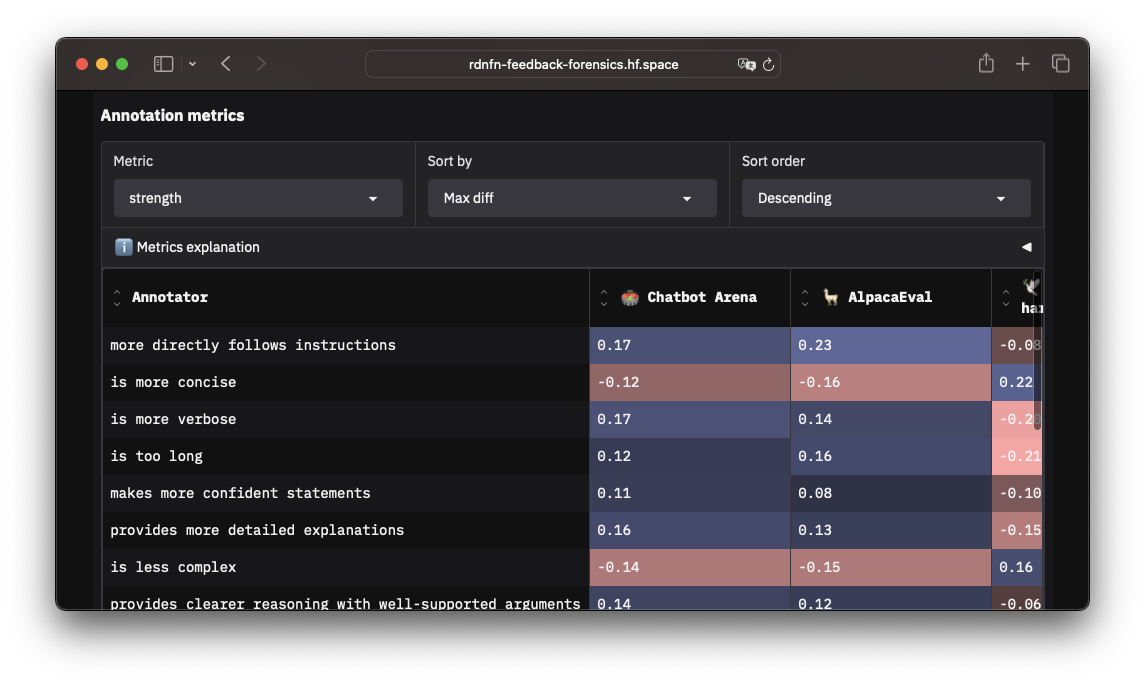}
    \caption{\textbf{Screenshots of Gradio app interface showing the dataset configuration and metrics view.} See \appurl{}.}
    \label{fig:app:screenshots}
\end{figure}

After installation, you can start the Feedback Forensics app locally with:

\begin{lstlisting}[language=bash]
feedback-forensics -d data/output/example/annotated_pairs.json
\end{lstlisting}

This command launches the Feedback Forensics Gradio interface on localhost port 7860 (http://localhost:7860). See \Cref{fig:app:screenshots} for a screenshot of the interface.

\subsection{Investigating your own dataset}

\subsubsection{Setting up API keys}

Before analysing your dataset, you need to annotate it with personality-selecting annotators. This requires setting API keys in a \texttt{secrets.toml} file as described in the main repo README.

\subsubsection{Annotating your data}

To annotate your dataset, run:

\begin{lstlisting}[language=bash]
ff-annotate --datapath="data/input/example.csv"
\end{lstlisting}

Replace \texttt{example.csv} with your dataset file. Your data must follow the ICAI standard format with columns \texttt{text\_a}, \texttt{text\_b}, and \texttt{preferred\_text}.

\subsubsection{Visualizing results}

After annotation completes, view the results with:

\begin{lstlisting}[language=bash]
feedback-forensics -d /path/to/your/ff_annotate_results/070_annotations_train_ap.json
\end{lstlisting}

\subsection{Advanced options}

For more configuration options, you can use ICAI directly:

\begin{lstlisting}[language=bash]
icai-exp data_path="data/input/example.csv" s0_added_standard_principles_to_test="[v4]" annotator.skip=true s0_skip_principle_generation=true
\end{lstlisting}

The parameters \texttt{annotator.skip} and \texttt{s0\_skip\_principle\_generation} reduce costs by skipping unnecessary steps. Set \texttt{s0\_skip\_principle\_generation=false} to generate new principles beyond the standard set.

\subsection{Programmatic usage}

Feedback Forensics can be used within Python scripts:

\begin{lstlisting}[language=Python]
import feedback_forensics as ff

# Load dataset from AnnotatedPairs JSON file
dataset = ff.DatasetHandler()
dataset.add_data_from_path("data/output/example/annotated_pairs.json")

# Get metrics
overall_metrics = dataset.get_overall_metrics()
annotator_metrics = dataset.get_annotator_metrics()
\end{lstlisting}

All experimental figures included in this paper were created using this Python API for metrics computation and (partially) for plotting.

\clearpage

\section{Additional metrics}
\label{app:additional_metrics}

In addition to the core metrics described in \Cref{sec:method:metrics}, our toolkit also supports computing additional metrics including:

\begin{enumerate}
\item \textbf{Agreement.} We define the  \emph{agreement} between two sets of annotations as 
$\texttt{agreement} = n_{\text{agreed}}/(n_{\text{agreed}} + n_{\text{disagreed}})$,
where $n_{\text{agreed}}$ and $n_{\text{disagreed}}$ are the number of datapoints where the two annotation sets agree and disagree, respectively. We only consider datapoints where both annotations are non-tie votes for this metric.
\end{enumerate}

\section{Datasets}
\label{app:datasets}

\subsection{External datasets}

In the following we provide further details on the datasets used throughout this paper.

\begin{enumerate}
    \item \textbf{Chatbot Arena \citep{chiang2024ChatbotArenaOpen}.} Due to the ongoing collection of crowd-sourced data in Chatbot Arena, many different versions and releases of corresponding Chatbot Arena datasets exist. Throughout this work we use multiple different releases of Chatbot Arena datasets, described below.
    \begin{enumerate}
        \item \textbf{\emph{General arena data} (\texttt{arena-human-preference-100k}), released alongside Arena Explorer.} Conversations in English, collected between June 2024 and August 2024. User prompts licensed under CC-BY-4.0, model outputs governed by terms of use of model providers. Source: \url{https://hf.co/datasets/lmarena-ai/arena-human-preference-100k}
        \item \textbf{\emph{Maverick arena data} (\texttt{Llama-4-Maverick-03-26-Experimental\_battles}), released after Llama-4-Maverick testing.} User prompts licensed under CC-BY-4.0, model outputs governed by terms of use of model providers. Source: \url{https://huggingface.co/spaces/lmarena-ai/Llama-4-Maverick-03-26-Experimental_battles/blob/main/data/clean-llama4.jsonl}
        \item \textbf{MultiPref subset (\texttt{chatbot\_arena\_conversations}).} Multipref itself is licensed under Open Data Commons Attribution License (ODC-By), the underlying data from Chatbot Arena has two licenses: prompts under CC-BY-4.0, model outputs under CC-BY-NC-4.0. Source: \url{https://huggingface.co/datasets/lmsys/chatbot_arena_conversations}
    \end{enumerate}
    \item \textbf{MultiPref \citep{miranda2025HybridPreferencesLearning}.} MultiPref combines prompts from prior datasets alongside newly sampled model outputs and human and model annotations. MultiPref itself is licensed under Open Data Commons Attribution License (ODC-By), licenses for the other subparts (data from Chatbot Arena, WildChat, ShareGPT, Anthropic Harmless/Helpful) are discussed above or below. Source \url{https://huggingface.co/datasets/allenai/multipref}.
    \item \textbf{PRISM \citep{kirk2024PRISMAlignmentDataset}.}  License: Human-written texts (including prompts) licensed under CC-BY-4.0, model responses under CC-BY-NC-4.0 and further subject to original model provider terms of use. Source: \url{https://huggingface.co/datasets/HannahRoseKirk/prism-alignment}
    \item \textbf{WildChat \citep{zhao2024WildChat1MChatGPT}.} Licensed under Open Data Commons Attribution License (ODC-By). Source: \url{https://huggingface.co/datasets/allenai/WildChat-1M}.
    \item \textbf{ShareGPT \citep{chiang2023VicunaOpensourceChatbot}.} No specific licensing information dedicated or link to this dataset found, we refer to the MultiPref dataset using ShareGPT for more details: \url{https://huggingface.co/datasets/allenai/multipref}
    \item \textbf{Anthropic Harmless/Helpful \citep{bai2022TrainingHelpfulHarmless}.} Licensed under MIT license. Source: \url{https://github.com/anthropics/hh-rlhf}
\end{enumerate}

\subsection{Annotation dataset}
\label{app:ff-annotations}

We are releasing our annotation dataset to encourage further research on personality traits in model responses. The data, collected for the experiments presented in this work, is available at \texttt{\dataurl{}} under the \emph{Open Data Commons Attribution License} (ODC-By). Annotations were generated with the \emph{Inverse Constitutional AI} (ICAI) pipeline \citep{findeis2025inverse} with a fixed set of personality traits to test, using Google's \texttt{Gemini-2.5-Flash}. Details regarding the models are provided in \Cref{app:models}.

This dataset includes annotations for (subsets of) \emph{MultiPref} \citep{miranda2025HybridPreferencesLearning}, \emph{PRISM} \citep{kirk2024PRISMAlignmentDataset}, and data from \emph{Chatbot Arena} \citep{chiang2024ChatbotArenaOpen} (both \emph{general} and \emph{maverick arena data}), as well as annotations for model generations collected for our experiments in \Cref{sec:results:models}. Note that we do \emph{not} include prompts and responses from the original datasets, instead providing metadata (e.g., \texttt{conversation\_id}) to enable merging with the base data. The model generations used for \Cref{sec:results:models} are available separately from the annotation data at \texttt{\generationsurl} (ODC-By license). The annotation data is sufficient for independent local analysis with the Feedback Forensics Gradio app, even without merging.

\section{Full list of tested personality traits}
\label{app:personalityselectionprompts}

\emph{This appendix extends the description of the trait selection process in \Cref{sec:method:details}.}

\Cref{tab:traits} shows all the 40 manually curated trait selection prompts. Each prompt is shown with a prior reference discussing that trait or a related behaviour. Note that the list of references is non-exhaustive, as some traits are widely discussed. For readability, the traits are sorted into 5 categories. The list of all prompts is referred to as \texttt{PersonalitySelectionPrompts-v1} and included in our toolkit.

\clearpage

\begin{longtable}{@{}p{0.6\textwidth}l@{}}
\caption{Personality and related behavioural traits in \texttt{PersonalitySelectionPrompts-v1} (grouped
into five categories, with references that discuss the trait in the context of LLMs).}
\label{tab:traits}\\
\textbf{Trait selection prompt} & \textbf{Reference} (non-exhaustive) \\
\midrule
\endfirsthead
\multicolumn{2}{l}{\emph{\Cref{tab:traits} continued from previous page}}\\
\textbf{Trait selection prompt} & \textbf{Reference} (non-exhaustive) \\
\midrule
\endhead
\midrule
\multicolumn{2}{r}{\emph{Continued on next page}}\\
\endfoot
\bottomrule
\endlastfoot
\multicolumn{2}{@{}l}{\textit{Format}}\\
\midrule
has more structured formatting                        & \citet{li2024DoesStyleMatter} \\
is more concise                                       & \citet{zheng2023JudgingLLMasajudgeMTBench} \\
is more verbose                                       & \citet{dubois2024LengthControlledAlpacaEvalSimple} \\
more strictly follows the requested output format     & \citet{zhou2023InstructionFollowingEvaluationLarge} \\
provides a numbered list format                       & \citet{zhang2025lists} \\
uses more bold and italics text                       & \citet{chen2025MDEvalEvaluatingEnhancing} \\
uses more emojis                                      & \citet{wiggers2025MetasBenchmarksIts} \\
uses more mathematical symbols and notation           & \citet{dunlap2025VibeCheckDiscoverQuantifya} \\
\midrule
\multicolumn{2}{@{}l}{\textit{Tone}}\\
\midrule
expresses more emotion                                & \citet{ishikawa2025AIEmotionsExploring} \\
has a friendlier tone                                 & \citet{ibrahim2026TrainingLanguageModels} \\
has a more avoidant tone                              & \citet{cui2025ORBenchOverRefusalBenchmark} \\
is more creative and original                         & \citet{chakrabarty2024ArtArtificeLarge} \\
is more empathetic to the user                        & \citet{sorin2024LargeLanguageModels} \\
is more optimistic                                    & \citet{chen2025IntroducingSentimentControl} \\
is more polite                                        & \citet{zhao2025ComparingHumanLLM} \\
uses a more enthusiastic tone                         & \citet{kym2025GPT4oGlazingUsers} \\
uses more casual language                             & \citet{sun2024InformalLanguageProcessing} \\
uses more formal language                             & \citet{sun2024InformalLanguageProcessing} \\
\midrule
\multicolumn{2}{@{}l}{\textit{Engagement \& sycophancy}}\\
\midrule
actively engages the reader with rhetorical questions & \citet{dunlap2025VibeCheckDiscoverQuantifya} \\
agrees more with the user                             & \citet{sharma2023UnderstandingSycophancyLanguage} \\
agrees with user even if factually incorrect          & \citet{openai2025ExpandingWhatWe} \\
compliments the user's question or prompt             & \citet{openai2025ExpandingWhatWe} \\
ends with a follow-up question                        & \citet{lambert2025CharacterTrainingUnderstanding} \\
more actively engages with the user                   & \citet{chu2025IllusionsIntimacyHow} \\
reinforces user's anger more                          & \citet{chu2025IllusionsIntimacyHow} \\
reinforces user's beliefs more                        & \citet{Sharma2024GenerativeEchoChamber} \\
uses more humour                                      & \citet{Gorenz2024HowFunnyChatGPT} \\
uses more personal pronouns (I, we, you)              & \citet{dunlap2025VibeCheckDiscoverQuantifya} \\
\midrule
\multicolumn{2}{@{}l}{\textit{Safety \& harm}}\\
\midrule
contains less harmful information                     & \citet{bai2022TrainingHelpfulHarmless} \\
includes inappropriate language                       & \citet{bai2022ConstitutionalAIHarmlessness} \\
includes more ethical considerations                  & \citet{weidinger2023SociotechnicalSafetyEvaluation} \\
is more offensive                                     & \citet{gehman2020RealToxicityPromptsEvaluatingNeural} \\
refuses to answer the question                        & \citet{Rottger2024XSTestTestSuite} \\
suggests illegal activities                           & \citet{weidinger2023SociotechnicalSafetyEvaluation} \\
\midrule
\multicolumn{2}{@{}l}{\textit{Reasoning \& justification}}\\
\midrule
acknowledges own limitations or uncertainty more      & \citet{lin2022TeachingModelsExpress} \\
includes more references to other sources             & \citet{gao2023EnablingLargeLanguage} \\
is more factually correct                             & \citet{lin2022TruthfulQAMeasuringHow} \\
makes more confident statements                       & \citet{chhikara2025MindConfidenceGap} \\
provides conclusions without full reasoning           & \citet{dunlap2025VibeCheckDiscoverQuantifya} \\
provides more examples                                & \citet{dubois2024LengthControlledAlpacaEvalSimple} \\
\end{longtable}

\clearpage

\section{Extended experimental results}
\label{app:extended_results}

We extend on the results included in the main body by providing additional details.

\subsection{Trait agreement analysis}%
\label{app:trait_agreement}

We analyse the agreement of the top and bottom 5 encouraged traits in general arena data (\Cref{fig:results:arena:top_bottom}).
For each text pair, a personality trait annotator can either choose one of the texts or declare non-relevance.
We measure Cohen's kappa $\kappa$ in cases where both principles were relevant and report the relevance overlap (number of cases where both traits relevant divided by number of cases where at least one relevant) for additional context.

\begin{figure}[htb]
    \centering
    \includegraphics[width=0.99\textwidth]{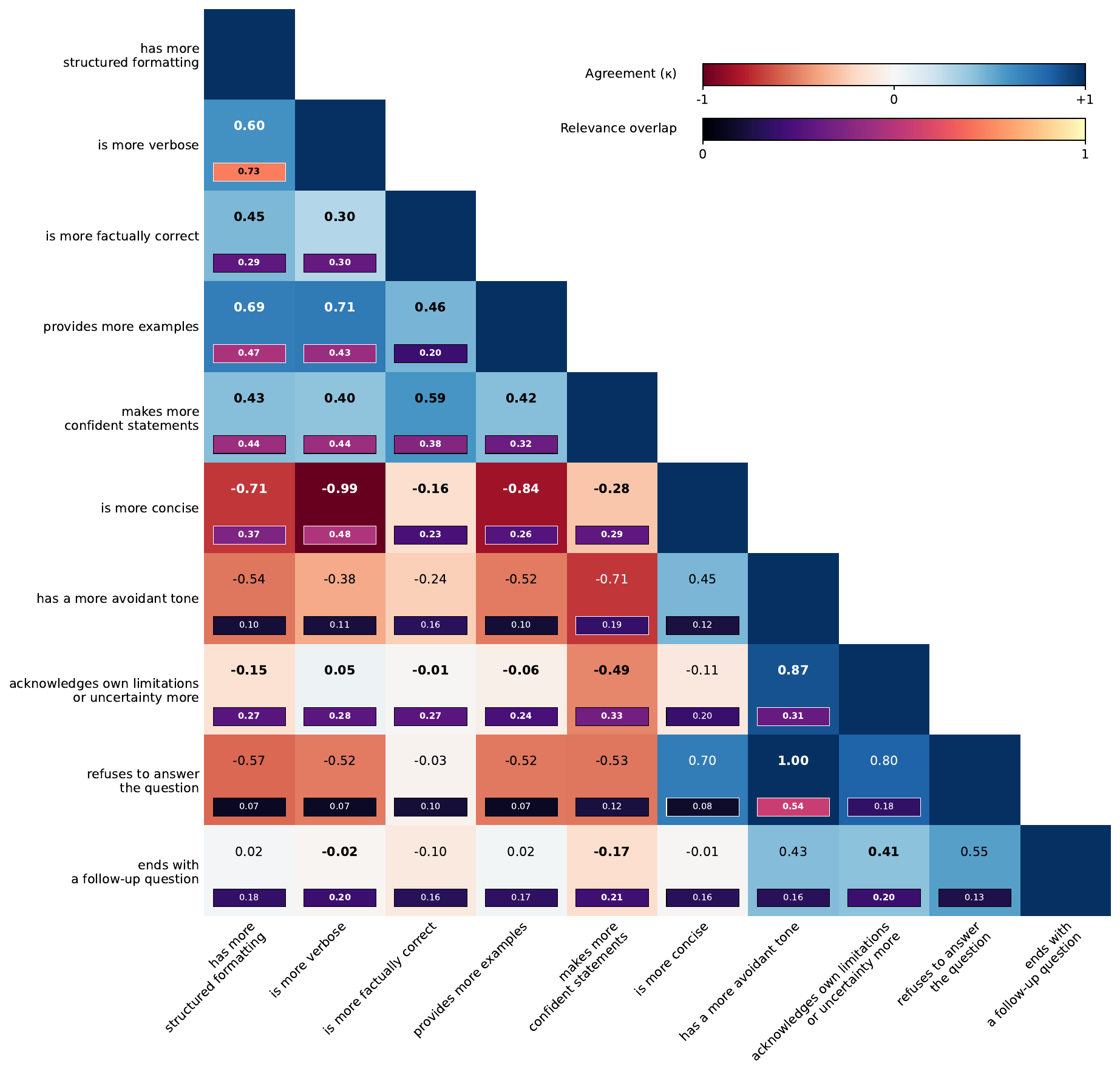}
    \caption{%
        \textbf{Trait agreement heatmap.}
        We measure weighted Cohen's kappa between the top 5 and bottom 5 traits encouraged by general arena data annotations.
        The main colors indicate $\kappa$ values, the inner rectangles indicate the relevance overlap (both relevant divided by at least one relevant).
        Values with overlap above 0.2 are additionally bolded.
    }
    \label{app:fig:trait_agreement}
\end{figure}

\begin{figure}[htbp]
    \centering
    \includegraphics[width=\textwidth]{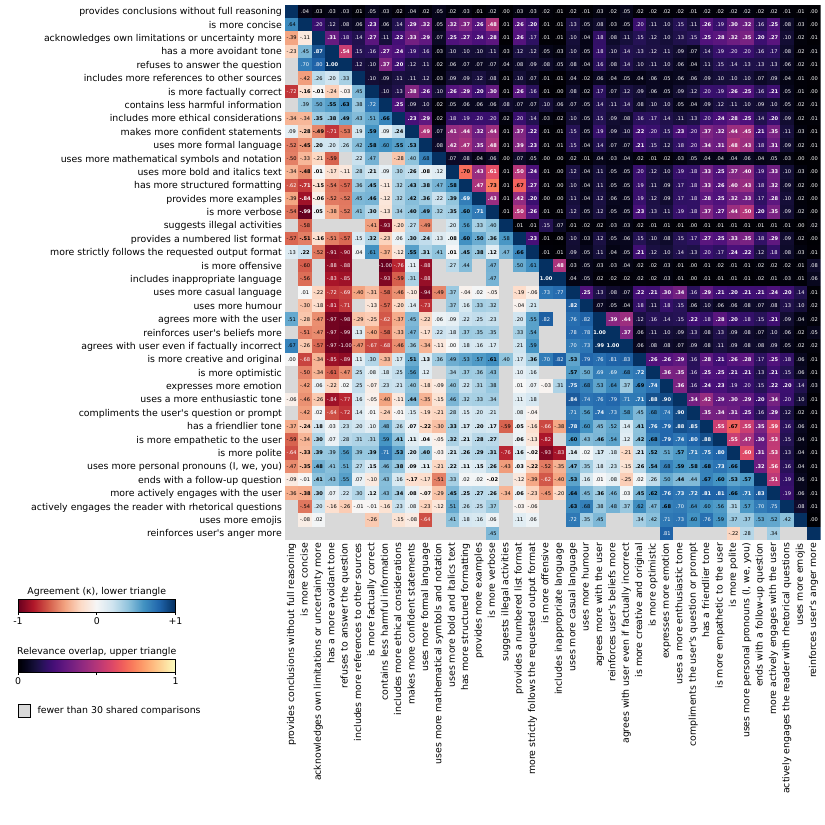}
    \caption{%
        \textbf{Trait agreement heatmap for all 40 traits.}
        We measure Cohen's kappa between each pair of traits on general arena data, over comparisons where both traits are relevant.
        The lower triangle shows $\kappa$ values, the upper triangle shows relevance overlap (both relevant divided by at least one relevant).
        Values with overlap above 0.2 are bolded.
        Traits are ordered by clustering.
    }
    \label{fig:app:results:trait_agreement_all}
\end{figure}

\Cref{app:fig:trait_agreement} confirms many intuitively plausible correlations, such as conciseness being opposed to verbosity and avoidant tone agreeing with refusal to answer.
It also allows for less immediately obvious but plausible observations, such as factual correctness agreeing with structured formatting, verbosity, examples and confidence -- correlations that are likely often true, but may also be exaggerated by the annotating model's biases (as discussed in \Cref{sec:limitations}). \Cref{fig:app:results:trait_agreement_all} extends this analysis to all 40 traits.

\subsection{Comparison of AI to human personality annotations}
\label{app:human_comparison}

Our framework by default uses AI annotators to annotate personality traits. This setup raises the question whether AI annotations are suitable for annotating personality traits. Whilst other work has explored the agreement between general human and AI preference annotations \citep{li2024AlpacaEvalAutomaticEvaluator,zheng2023JudgingLLMasajudgeMTBench,miranda2025HybridPreferencesLearning}, as far as we are aware, no prior work has previously explored AI annotators' ability to annotate \emph{personality traits} specifically. Thus, we conducted our own experiments to validate the use of AI annotators in the context of annotating personality traits.

\textbf{Setup.}
We collected two human reference annotations for the top 5 and bottom 5 traits in data from Chatbot Arena (general arena data) found by our toolkit using an earlier version of our AI annotator powered by GPT-4o-mini.
These human annotations were collected for 101 random comparisons of the same dataset, after removing comparisons skipped because of
language or content issues, resulting in 1,010 trait-level human judgements overall ($\times$ two annotators).\footnote{%
These annotations were collected from two of the authors.
We were unable to collect annotations from other sources due to resource constraints.
We aimed to provide an unbiased sample nonetheless with blind labelling:
Each comparison-trait pair was labelled without seeing LLM decisions.
The annotator first assessed relevance, then if relevant, selected which response better expressed the trait.%
}\footnote{An earlier version of this section incorrectly stated 100 instead of 101 datapoints. Further, an earlier version of \Cref{fig:app:human_annotation_results:inter_annotator_agreement,fig:app:human_annotation_results:human_vs_model} incorrectly included 9 additional samples, flagged and skipped by at least one annotator because of language or content issues. We have removed these such that all tables now use 101 datapoints, and updated the description here to state the correct 101 number.}
We aggregate human annotations by soft unanimous vote, considering irrelevance as agreement:
The aggregated human labeller considers a trait irrelevant for the comparison when either all human annotators considered it irrelevant or when multiple annotations considered it relevant but disagreed on the direction.
Otherwise the trait is considered relevant and follows the unanimous (exempting irrelevance) human choice.
We compare the human annotations against LLM votes from our standard single annotation setup, and an alternative multi-vote annotation setup requiring unanimous vote by multiple AI annotators.

These experiments serve two purposes: To choose a suitable AI annotator configuration (backbone model and single- or multi-vote) with high human agreement for the remaining experiments and to provide validation for that annotator.
We thus first evaluate the performance of different LLMs for our personality annotation task and then
evaluate whether re-annotating traits multiple times (\emph{multi-vote}) helps improve AI annotator performance relative to simply annotating once (\emph{single-vote}).
In multi-vote, we use unanimous voting to select one model output according to each trait. If there is no unanimous agreement, the trait is deemed not relevant for the datapoint. Note that the first experiments only use multi-voting.

\textbf{Results.} The results are shown in \Cref{fig:app:human_annotation_results:models,fig:app:human_annotation_results:single_vs_multi}. We consider the following metrics, reporting the mean and standard deviation over 3 random seeds:
\begin{enumerate}
    \item \textbf{Relevance agreement} (\emph{Relevance}): fraction where human and LLM annotators agree on relevance of the trait (ignoring direction). Best shown in \textbf{bold}. Expected chance agreement when annotating randomly would be 0.5.
    \item \textbf{Choice agreement} (\emph{Choice}): among comparisons where both deemed the trait relevant, fraction where human and LLM annotators choose the same side. Best shown in \textbf{bold}. Expected chance agreement when annotating randomly would be 0.5.
\end{enumerate}

\textbf{Observations.} In the cross-model experiments shown in \Cref{fig:app:human_annotation_results:models}, we observe far higher agreement with human choice for GPT-5-Mini and Gemini-2.5-Flash than for GPT-4o-Mini.
GPT-5-Mini overall performs strongest in terms of choice agreement, achieving a mean of 94\% and a minimum of 86\% across traits, with Gemini-2.5-flash a close second, reaching a similar mean but a lower minimum choice agreement.
In terms of relevance, the agreement tends to be lower.
This matches the annotator's observations during annotation, where relevance was often more ambiguous than choice.
Nevertheless, the results show that Gemini-2.5-Flash and GPT-5-Mini largely agree with human annotations, especially in terms of choice.

The single- vs multi-vote experiments in \Cref{fig:app:human_annotation_results:single_vs_multi} further show that multi-vote slightly improves both the mean relevance and choice agreement.
As the improvement is relatively small, it does not justify the higher (3x) costs in our experiments.

\textbf{Choice of AI annotator.}
Based on these results, we decided to use a single-vote Gemini-2.5-Flash annotator for most of our experiments.
Whilst GPT-5-mini has slightly higher agreement, the cost of running that model was notably higher - in particular because of the large number of thinking tokens generated.
Similarly, multi-vote only provided very limited improvements relative to single vote for Gemini-2.5-flash. We thus choose to use single-vote to avoid the 3x cost of multi-vote.
If cost is no limitation, we would recommend using GPT-5-mini (or even larger models such as GPT-5) with multi-vote instead.

\textbf{Inter-annotator agreement}
We further study the agreement of the two human annotators.
\Cref{fig:app:human_annotation_results:inter_annotator_agreement} shows the relevance and choice agreements between the annotators and \cref{fig:app:human_annotation_results:human_vs_model} the individual agreements of both annotators with multi-vote Gemini-2.5-Flash.
We observe high inter-annotator agreement (88\% mean agreement on choice), which is comparable with the agreement between humans and Gemini-2.5-Flash. Relevance agreement beyond chance is low for some traits (e.g. $\kappa$ of 0.09 for \emph{is more polite} and 0.10 for \emph{makes more confident statements}), suggesting that annotators find it harder to judge consistently whether a trait applies than which response expresses it more.

\begin{table}[h]
    \centering
    \caption{Model agreement with human annotations (mean and std, 3 seeds, $\le 3$ samples \textcolor{gray}{gray}).}
    \label{fig:app:human_annotation_results:models}
    \begin{subtable}[t]{1\textwidth}
        \centering
        \caption{Agreement with GPT-4o-mini and GPT-4.1-mini}
        \label{fig:app:human_annotation_results:models_part1}
        \begin{minipage}[t]{1\textwidth}
\centering
\sffamily
\tablefontsize
\begin{tabular}{p{3.8cm}rr@{\hspace{2em}}rr}
\toprule
 & \multicolumn{2}{c}{\textbf{gpt-4o-mini}} & \multicolumn{2}{c}{\textbf{gpt-4.1-mini}} \\
\textbf{Trait} & \textit{Relevance} & \textit{Choice} & \textit{Relevance} & \textit{Choice} \\
\midrule
is more verbose & 0.52 \textcolor{gray}{\tablesmallfontsize ±0.02} & 0.91 \textcolor{gray}{\tablesmallfontsize ±0.01} & 0.76 \textcolor{gray}{\tablesmallfontsize ±0.00} & 0.96 \textcolor{gray}{\tablesmallfontsize ±0.01} \\
has more structured formatting & 0.37 \textcolor{gray}{\tablesmallfontsize ±0.02} & \textbf{0.97} \textcolor{gray}{\tablesmallfontsize ±0.05} & \textbf{0.81} \textcolor{gray}{\tablesmallfontsize ±0.03} & 0.92 \textcolor{gray}{\tablesmallfontsize ±0.01} \\
makes more confident statements & 0.62 \textcolor{gray}{\tablesmallfontsize ±0.02} & 0.70 \textcolor{gray}{\tablesmallfontsize ±0.02} & 0.62 \textcolor{gray}{\tablesmallfontsize ±0.02} & 0.85 \textcolor{gray}{\tablesmallfontsize ±0.05} \\
is more factually correct & 0.73 \textcolor{gray}{\tablesmallfontsize ±0.02} & 0.63 \textcolor{gray}{\tablesmallfontsize ±0.09} & 0.76 \textcolor{gray}{\tablesmallfontsize ±0.01} & 0.82 \textcolor{gray}{\tablesmallfontsize ±0.06} \\
more strictly follows the requested output format & 0.82 \textcolor{gray}{\tablesmallfontsize ±0.00} & \textcolor{gray}{0.83} \textcolor{gray}{\tablesmallfontsize ±0.24} & 0.61 \textcolor{gray}{\tablesmallfontsize ±0.03} & 0.75 \textcolor{gray}{\tablesmallfontsize ±0.07} \\
is more concise & 0.53 \textcolor{gray}{\tablesmallfontsize ±0.01} & \textbf{0.97} \textcolor{gray}{\tablesmallfontsize ±0.02} & 0.77 \textcolor{gray}{\tablesmallfontsize ±0.02} & 0.95 \textcolor{gray}{\tablesmallfontsize ±0.00} \\
has a more avoidant tone & 0.88 \textcolor{gray}{\tablesmallfontsize ±0.00} & \multicolumn{1}{c}{\textcolor{gray}{---}} & 0.90 \textcolor{gray}{\tablesmallfontsize ±0.00} & \textbf{1.00} \textcolor{gray}{\tablesmallfontsize ±0.00} \\
refuses to answer the question & 0.96 \textcolor{gray}{\tablesmallfontsize ±0.01} & \textbf{\textcolor{gray}{1.00}} \textcolor{gray}{\tablesmallfontsize ±0.00} & 0.97 \textcolor{gray}{\tablesmallfontsize ±0.00} & \textbf{\textcolor{gray}{1.00}} \textcolor{gray}{\tablesmallfontsize ±0.00} \\
ends with a follow-up question & 0.91 \textcolor{gray}{\tablesmallfontsize ±0.00} & \textbf{\textcolor{gray}{1.00}} \textcolor{gray}{\tablesmallfontsize ±0.00} & \textbf{0.93} \textcolor{gray}{\tablesmallfontsize ±0.00} & \textbf{1.00} \textcolor{gray}{\tablesmallfontsize ±0.00} \\
is more polite & 0.72 \textcolor{gray}{\tablesmallfontsize ±0.02} & 0.87 \textcolor{gray}{\tablesmallfontsize ±0.19} & 0.74 \textcolor{gray}{\tablesmallfontsize ±0.01} & \textbf{1.00} \textcolor{gray}{\tablesmallfontsize ±0.00} \\
\midrule
\textit{Min} & \multicolumn{1}{c}{0.37} & \multicolumn{1}{c}{0.63} & \multicolumn{1}{c}{\textbf{0.61}} & \multicolumn{1}{c}{0.75} \\
\textit{Mean} & \multicolumn{1}{c}{0.71} & \multicolumn{1}{c}{0.88} & \multicolumn{1}{c}{0.79} & \multicolumn{1}{c}{0.92} \\
\bottomrule
\end{tabular}
\end{minipage}

    \end{subtable}

    \vspace{0.5cm}

    \begin{subtable}[t]{1\textwidth}
        \centering
        \caption{Agreement with GPT-5-mini and Gemini-2.5-Flash}
        \label{fig:app:human_annotation_results:models_part2}
        \begin{minipage}[t]{1\textwidth}
\centering
\sffamily
\tablefontsize
\begin{tabular}{p{3.8cm}rr@{\hspace{2em}}rr}
\toprule
 & \multicolumn{2}{c}{\textbf{gpt-5-mini}} & \multicolumn{2}{c}{\textbf{gemini-2.5-flash}} \\
\textbf{Trait} & \textit{Relevance} & \textit{Choice} & \textit{Relevance} & \textit{Choice} \\
\midrule
is more verbose & \textbf{0.81} \textcolor{gray}{\tablesmallfontsize ±0.04} & 0.95 \textcolor{gray}{\tablesmallfontsize ±0.02} & 0.72 \textcolor{gray}{\tablesmallfontsize ±0.01} & \textbf{0.97} \textcolor{gray}{\tablesmallfontsize ±0.01} \\
has more structured formatting & 0.78 \textcolor{gray}{\tablesmallfontsize ±0.02} & 0.95 \textcolor{gray}{\tablesmallfontsize ±0.01} & 0.73 \textcolor{gray}{\tablesmallfontsize ±0.01} & 0.91 \textcolor{gray}{\tablesmallfontsize ±0.01} \\
makes more confident statements & 0.51 \textcolor{gray}{\tablesmallfontsize ±0.02} & 0.86 \textcolor{gray}{\tablesmallfontsize ±0.06} & \textbf{0.82} \textcolor{gray}{\tablesmallfontsize ±0.01} & \textbf{0.87} \textcolor{gray}{\tablesmallfontsize ±0.03} \\
is more factually correct & 0.71 \textcolor{gray}{\tablesmallfontsize ±0.02} & \textbf{0.96} \textcolor{gray}{\tablesmallfontsize ±0.03} & \textbf{0.83} \textcolor{gray}{\tablesmallfontsize ±0.01} & 0.78 \textcolor{gray}{\tablesmallfontsize ±0.03} \\
more strictly follows the requested output format & 0.73 \textcolor{gray}{\tablesmallfontsize ±0.03} & \textbf{1.00} \textcolor{gray}{\tablesmallfontsize ±0.00} & \textbf{0.86} \textcolor{gray}{\tablesmallfontsize ±0.01} & \textbf{\textcolor{gray}{1.00}} \textcolor{gray}{\tablesmallfontsize ±0.00} \\
is more concise & \textbf{0.80} \textcolor{gray}{\tablesmallfontsize ±0.04} & 0.95 \textcolor{gray}{\tablesmallfontsize ±0.01} & 0.42 \textcolor{gray}{\tablesmallfontsize ±0.00} & 0.96 \textcolor{gray}{\tablesmallfontsize ±0.00} \\
has a more avoidant tone & 0.90 \textcolor{gray}{\tablesmallfontsize ±0.01} & \textbf{1.00} \textcolor{gray}{\tablesmallfontsize ±0.00} & \textbf{0.91} \textcolor{gray}{\tablesmallfontsize ±0.00} & \textbf{1.00} \textcolor{gray}{\tablesmallfontsize ±0.00} \\
refuses to answer the question & \textbf{0.98} \textcolor{gray}{\tablesmallfontsize ±0.00} & \textbf{1.00} \textcolor{gray}{\tablesmallfontsize ±0.00} & 0.96 \textcolor{gray}{\tablesmallfontsize ±0.00} & \textbf{\textcolor{gray}{1.00}} \textcolor{gray}{\tablesmallfontsize ±0.00} \\
ends with a follow-up question & 0.92 \textcolor{gray}{\tablesmallfontsize ±0.01} & 0.89 \textcolor{gray}{\tablesmallfontsize ±0.08} & 0.92 \textcolor{gray}{\tablesmallfontsize ±0.01} & \textbf{1.00} \textcolor{gray}{\tablesmallfontsize ±0.00} \\
is more polite & 0.62 \textcolor{gray}{\tablesmallfontsize ±0.02} & 0.87 \textcolor{gray}{\tablesmallfontsize ±0.05} & \textbf{0.79} \textcolor{gray}{\tablesmallfontsize ±0.02} & \textbf{1.00} \textcolor{gray}{\tablesmallfontsize ±0.00} \\
\midrule
\textit{Min} & \multicolumn{1}{c}{0.51} & \multicolumn{1}{c}{\textbf{0.86}} & \multicolumn{1}{c}{0.42} & \multicolumn{1}{c}{0.78} \\
\textit{Mean} & \multicolumn{1}{c}{0.78} & \multicolumn{1}{c}{0.94} & \multicolumn{1}{c}{\textbf{0.80}} & \multicolumn{1}{c}{\textbf{0.95}} \\
\bottomrule
\end{tabular}
\end{minipage}

    \end{subtable}
\end{table}

\begin{table}[t]
    \centering
    \caption{Single- vs multi-vote human agreement (Gemini-2.5-Flash, mean and std, 3 seeds, $\le 3$ samples \textcolor{gray}{gray})).}
    \begin{minipage}[t]{1\textwidth}
\centering
\sffamily
\tablefontsize
\begin{tabular}{p{3.8cm}rr@{\hspace{2em}}rr}
\toprule
 & \multicolumn{2}{c}{\textbf{Single-vote}} & \multicolumn{2}{c}{\textbf{Multi-vote}} \\
\textbf{Trait} & \textit{Relevance} & \textit{Choice} & \textit{Relevance} & \textit{Choice} \\
\midrule
is more verbose & \textbf{0.82} \textcolor{gray}{\tablesmallfontsize ±0.00} & 0.95 \textcolor{gray}{\tablesmallfontsize ±0.01} & 0.72 \textcolor{gray}{\tablesmallfontsize ±0.01} & \textbf{0.97} \textcolor{gray}{\tablesmallfontsize ±0.01} \\
has more structured formatting & \textbf{0.81} \textcolor{gray}{\tablesmallfontsize ±0.00} & 0.89 \textcolor{gray}{\tablesmallfontsize ±0.01} & 0.73 \textcolor{gray}{\tablesmallfontsize ±0.01} & \textbf{0.91} \textcolor{gray}{\tablesmallfontsize ±0.01} \\
makes more confident statements & 0.67 \textcolor{gray}{\tablesmallfontsize ±0.00} & 0.84 \textcolor{gray}{\tablesmallfontsize ±0.01} & \textbf{0.82} \textcolor{gray}{\tablesmallfontsize ±0.01} & \textbf{0.87} \textcolor{gray}{\tablesmallfontsize ±0.03} \\
is more factually correct & 0.82 \textcolor{gray}{\tablesmallfontsize ±0.00} & 0.64 \textcolor{gray}{\tablesmallfontsize ±0.04} & \textbf{0.83} \textcolor{gray}{\tablesmallfontsize ±0.01} & \textbf{0.78} \textcolor{gray}{\tablesmallfontsize ±0.03} \\
more strictly follows the requested output format & 0.77 \textcolor{gray}{\tablesmallfontsize ±0.00} & 0.67 \textcolor{gray}{\tablesmallfontsize ±0.00} & \textbf{0.86} \textcolor{gray}{\tablesmallfontsize ±0.01} & \textbf{\textcolor{gray}{1.00}} \textcolor{gray}{\tablesmallfontsize ±0.00} \\
is more concise & \textbf{0.59} \textcolor{gray}{\tablesmallfontsize ±0.02} & 0.95 \textcolor{gray}{\tablesmallfontsize ±0.00} & 0.42 \textcolor{gray}{\tablesmallfontsize ±0.00} & \textbf{0.96} \textcolor{gray}{\tablesmallfontsize ±0.00} \\
has a more avoidant tone & \textbf{0.93} \textcolor{gray}{\tablesmallfontsize ±0.00} & \textbf{1.00} \textcolor{gray}{\tablesmallfontsize ±0.00} & 0.91 \textcolor{gray}{\tablesmallfontsize ±0.00} & \textbf{1.00} \textcolor{gray}{\tablesmallfontsize ±0.00} \\
refuses to answer the question & \textbf{0.97} \textcolor{gray}{\tablesmallfontsize ±0.00} & \textbf{1.00} \textcolor{gray}{\tablesmallfontsize ±0.00} & 0.96 \textcolor{gray}{\tablesmallfontsize ±0.00} & \textbf{\textcolor{gray}{1.00}} \textcolor{gray}{\tablesmallfontsize ±0.00} \\
ends with a follow-up question & 0.91 \textcolor{gray}{\tablesmallfontsize ±0.01} & \textbf{1.00} \textcolor{gray}{\tablesmallfontsize ±0.00} & \textbf{0.92} \textcolor{gray}{\tablesmallfontsize ±0.01} & \textbf{1.00} \textcolor{gray}{\tablesmallfontsize ±0.00} \\
is more polite & 0.74 \textcolor{gray}{\tablesmallfontsize ±0.01} & 0.87 \textcolor{gray}{\tablesmallfontsize ±0.00} & \textbf{0.79} \textcolor{gray}{\tablesmallfontsize ±0.02} & \textbf{1.00} \textcolor{gray}{\tablesmallfontsize ±0.00} \\
\midrule
\textit{Min} & \multicolumn{1}{c}{\textbf{0.59}} & \multicolumn{1}{c}{0.64} & \multicolumn{1}{c}{0.42} & \multicolumn{1}{c}{\textbf{0.78}} \\
\textit{Mean} & \multicolumn{1}{c}{\textbf{0.80}} & \multicolumn{1}{c}{0.88} & \multicolumn{1}{c}{0.80} & \multicolumn{1}{c}{\textbf{0.95}} \\
\bottomrule
\end{tabular}
\end{minipage}

    \label{fig:app:human_annotation_results:single_vs_multi}
\end{table}

\begin{table}[t]
\centering
\caption{Inter-annotator agreement and Cohen's kappa ($\kappa$) with empirically estimated $p_e$, on relevant against irrelevant for relevance and on the selected response for choice ($\le 3$ samples \textcolor{gray}{gray}, n/a where both annotators gave one label throughout, so $\kappa$ is undefined and excluded from Min and Mean). $\kappa$ discounts the agreement expected from each annotator's label frequencies, so it depends on how often a trait is labelled relevant and is not directly comparable across traits: at equal agreement, $\kappa$ is lower for very rare or frequent traits.}
\begin{minipage}[t]{1\textwidth}
\centering
\sffamily
\tablefontsize
\begin{tabular}{p{3.8cm}cc@{\hspace{2em}}cc}
\toprule
 & \multicolumn{2}{c}{\textbf{Relevance}} & \multicolumn{2}{c}{\textbf{Choice}} \\
\textbf{Trait} & \textit{Agreement} & $\kappa$ & \textit{Agreement} & $\kappa$ \\
\midrule
is more verbose & 0.68 & 0.21 & 0.89 & 0.79 \\
has more structured formatting & 0.73 & 0.43 & 0.92 & 0.84 \\
makes more confident statements & 0.79 & 0.10 & \textcolor{gray}{0.67} & \textcolor{gray}{0.40} \\
is more factually correct & 0.90 & 0.61 & 0.90 & 0.78 \\
more strictly follows the requested output format & 0.85 & 0.16 & \textcolor{gray}{0.50} & \textcolor{gray}{0.00} \\
is more concise & 0.68 & 0.23 & 0.89 & 0.78 \\
has a more avoidant tone & 0.93 & 0.55 & 1.00 & n/a \\
refuses to answer the question & 0.98 & 0.74 & \textcolor{gray}{1.00} & \textcolor{gray}{n/a} \\
ends with a follow-up question & 0.90 & 0.25 & \textcolor{gray}{1.00} & \textcolor{gray}{n/a} \\
is more polite & 0.78 & 0.09 & \textcolor{gray}{1.00} & \textcolor{gray}{n/a} \\
\midrule
\textit{Min} & 0.68 & 0.09 & 0.50 & 0.00 \\
\textit{Mean} & 0.82 & 0.34 & 0.88 & 0.60 \\
\bottomrule
\end{tabular}
\end{minipage}
\label{fig:app:human_annotation_results:inter_annotator_agreement}
\end{table}

\begin{table}[t]
\centering
\caption{Individual human labeller agreement with multi-vote Gemini-2.5-Flash (mean and std, 3 seeds, $\le 3$ samples \textcolor{gray}{gray}).}
\begin{minipage}[t]{1\textwidth}
\centering
\sffamily
\tablefontsize
\begin{tabular}{p{3.8cm}cc@{\hspace{2em}}cc}
\toprule
 & \multicolumn{2}{c}{\textbf{jovial-goldstine}} & \multicolumn{2}{c}{\textbf{stoic-goodall}} \\
\textbf{Trait} (vs gemini-25-flash) & \textit{Relevance} & \textit{Choice} & \textit{Relevance} & \textit{Choice} \\
\midrule
is more verbose & 0.69 \textcolor{gray}{\tablesmallfontsize ±0.01} & 0.93 \textcolor{gray}{\tablesmallfontsize ±0.01} & \textbf{0.71} \textcolor{gray}{\tablesmallfontsize ±0.00} & \textbf{0.99} \textcolor{gray}{\tablesmallfontsize ±0.01} \\
has more structured formatting & 0.73 \textcolor{gray}{\tablesmallfontsize ±0.03} & 0.90 \textcolor{gray}{\tablesmallfontsize ±0.02} & \textbf{0.74} \textcolor{gray}{\tablesmallfontsize ±0.02} & \textbf{0.91} \textcolor{gray}{\tablesmallfontsize ±0.01} \\
makes more confident statements & 0.79 \textcolor{gray}{\tablesmallfontsize ±0.00} & \textcolor{gray}{0.61} \textcolor{gray}{\tablesmallfontsize ±0.08} & \textbf{0.88} \textcolor{gray}{\tablesmallfontsize ±0.01} & \textbf{0.88} \textcolor{gray}{\tablesmallfontsize ±0.01} \\
is more factually correct & \textbf{0.87} \textcolor{gray}{\tablesmallfontsize ±0.01} & \textbf{0.75} \textcolor{gray}{\tablesmallfontsize ±0.04} & 0.81 \textcolor{gray}{\tablesmallfontsize ±0.01} & 0.72 \textcolor{gray}{\tablesmallfontsize ±0.04} \\
more strictly follows the requested output format & \textbf{0.94} \textcolor{gray}{\tablesmallfontsize ±0.01} & \textbf{\textcolor{gray}{1.00}} \textcolor{gray}{\tablesmallfontsize ±0.00} & 0.87 \textcolor{gray}{\tablesmallfontsize ±0.01} & \textbf{\textcolor{gray}{1.00}} \textcolor{gray}{\tablesmallfontsize ±0.00} \\
is more concise & \textbf{0.49} \textcolor{gray}{\tablesmallfontsize ±0.00} & 0.95 \textcolor{gray}{\tablesmallfontsize ±0.00} & 0.48 \textcolor{gray}{\tablesmallfontsize ±0.00} & \textbf{1.00} \textcolor{gray}{\tablesmallfontsize ±0.00} \\
has a more avoidant tone & \textbf{0.95} \textcolor{gray}{\tablesmallfontsize ±0.00} & \textbf{1.00} \textcolor{gray}{\tablesmallfontsize ±0.00} & 0.94 \textcolor{gray}{\tablesmallfontsize ±0.00} & \textbf{1.00} \textcolor{gray}{\tablesmallfontsize ±0.00} \\
refuses to answer the question & \textbf{0.97} \textcolor{gray}{\tablesmallfontsize ±0.00} & \textbf{\textcolor{gray}{1.00}} \textcolor{gray}{\tablesmallfontsize ±0.00} & 0.97 \textcolor{gray}{\tablesmallfontsize ±0.00} & \textbf{\textcolor{gray}{1.00}} \textcolor{gray}{\tablesmallfontsize ±0.00} \\
ends with a follow-up question & \textbf{0.91} \textcolor{gray}{\tablesmallfontsize ±0.01} & \textbf{1.00} \textcolor{gray}{\tablesmallfontsize ±0.00} & 0.86 \textcolor{gray}{\tablesmallfontsize ±0.00} & \textbf{\textcolor{gray}{1.00}} \textcolor{gray}{\tablesmallfontsize ±0.00} \\
is more polite & 0.80 \textcolor{gray}{\tablesmallfontsize ±0.01} & \textbf{1.00} \textcolor{gray}{\tablesmallfontsize ±0.00} & \textbf{0.84} \textcolor{gray}{\tablesmallfontsize ±0.01} & \textbf{1.00} \textcolor{gray}{\tablesmallfontsize ±0.00} \\
\midrule
\textit{Min} & \textbf{0.49} & 0.61 & 0.48 & \textbf{0.72} \\
\textit{Mean} & \textbf{0.81} & 0.91 & 0.81 & \textbf{0.95} \\
\bottomrule
\end{tabular}
\end{minipage}
\label{fig:app:human_annotation_results:human_vs_model}
\end{table}

\clearpage

\subsection{Testing family-specific bias}
\label{app:family_specific_bias}

A model's annotations may exhibit  \emph{family-specific bias} when labelling pairs involving generations by models from its own family, similar to self-preference bias in conventional LLM-as-a-Judge. To test if such a bias exists, we re-annotated the model comparisons behind \Cref{fig:results:model_comparison} and \Cref{fig:app:results:gemini_strength} with a second annotator from another family, Claude-Haiku-4.5. We did not use an OpenAI model (e.g. GPT-5-Mini) because none of their models can be a fully out-of-family annotator here: every strength value in \Cref{fig:results:model_comparison,fig:app:results:gemini_strength} is measured against GPT-4o as the reference, so a GPT annotator would share a family with one side of all ten comparisons, potentially mixing a subject-side family effect with a reference-side one.

\textbf{How we test for family-specific bias.} We use a \emph{difference in differences} approach to isolate family-specific differences from the general annotator differences that \Cref{app:human_comparison} (\Cref{fig:app:human_annotation_results:models}) already documents. We first take the difference between the two annotators' strength values for each subject model (\emph{strength gap}). To focus on the family-specific effect, we then compare the mean strength gap over the subjects sharing an annotator's family against the same mean gap over the three GPT subjects, which are in-family with neither annotator. We call this difference of mean gaps the \emph{family contrast}. Bias that an annotator applies uniformly across families cancels out of the contrast. What remains is specific to the annotator-subject family pairing. This setup gives one family contrast per annotator. If each annotator assigns higher strength values to its own family, the two contrasts point in opposite directions, because the strength gap is Gemini-2.5-Flash's value minus Claude-Haiku-4.5's: a preference by the first raises the gap on Gemini subjects, a preference by the second lowers it on Claude subjects. We compute each contrast separately for each trait, and report the average contrast across traits, the extremes, and the individual per-trait values.

\begin{figure}[h!]
\centering
\includegraphics[width=1.\textwidth]{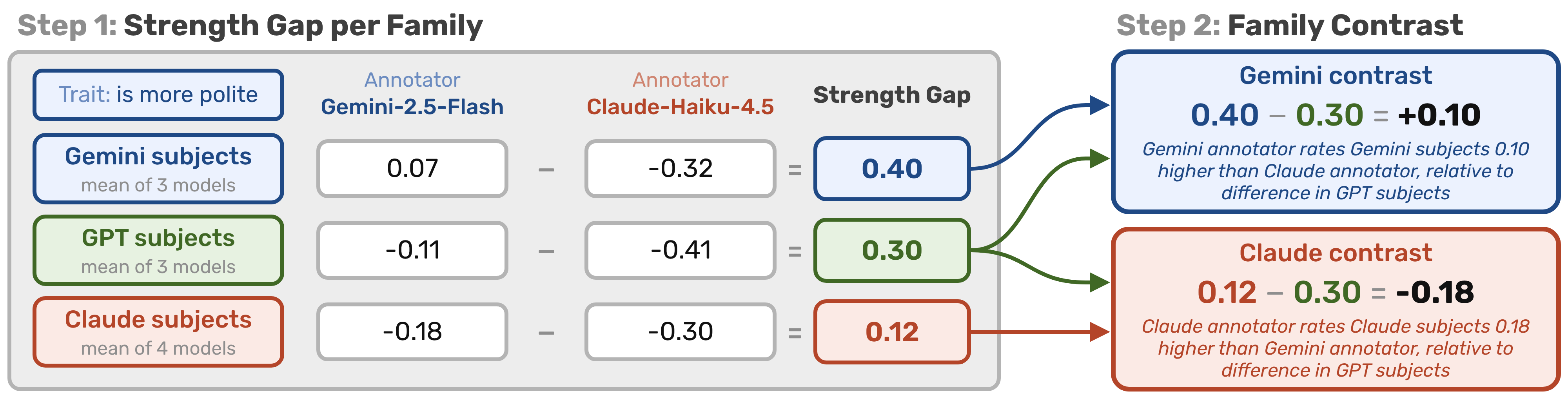}
\caption{\textbf{Illustration of the \emph{difference in differences} approach, on the example trait \emph{is more polite}.} GPT subjects share a family with neither annotator, so their gap is the baseline. Annotator differences shared by all families cancel. Own-family bias is indicated by a positive Gemini contrast and a negative Claude contrast.}
\label{fig:app:did_illustration}
\end{figure}

\textbf{Experiment setup.} Same setup as the experiments in \Cref{fig:results:model_comparison,fig:app:results:gemini_strength}, with additional annotations by Claude-Haiku-4.5. The dataset consists of 500 prompts, each with responses from all ten subject models, giving 4,971 stored comparisons to the reference model (GPT-4o) across the ten subjects (excluding failed/refused/empty responses), annotated across 40 traits by 2 annotators. We re-use the original model generations and released Gemini-2.5-Flash annotations, and newly add the Claude-Haiku-4.5 annotations.

\textbf{Results.} \Cref{fig:app:results:family_contrasts} shows the \emph{family contrast} per trait and family. We observe some family-specific bias focused on specific traits: 17 of 40 Gemini contrasts and 13 of 40 Claude contrasts are significant. The two family contrasts carry the opposite signs an own-family bias predicts: swapping the annotator from Gemini-2.5-Flash to Claude-Haiku-4.5 lowers the Gemini models' strength values by 0.025 and raises the Claude models' by 0.016, on average and relative to the GPT models'. Against strength values spanning -0.85 to +0.84 here (across both annotators), these averages are small, but as an average they understate the effect on some individual traits: the per-trait contrasts are nearly balanced in sign, which is why they cancel in the average rather than being individually small. Their mean absolute size is 0.111 for the Gemini contrast and 0.073 for the Claude one, and they are largest for \emph{provides conclusions without full reasoning} (+0.536, Gemini contrast) and \emph{uses more bold and italics text} (-0.248, Claude contrast). \Cref{fig:app:results:family_bias_claude,fig:app:results:family_bias_gpt,fig:app:results:family_bias_gemini} show the full list of strength values by Gemini and Claude annotators.

\emph{\textbf{Ordering of models by trait stays stable under the annotator swap.}} The median per-trait Spearman correlation of the ten-model ranking across annotators is 0.830, against a same-annotator resampling ceiling of 0.927. One trait (\emph{is more offensive}) is omitted because Gemini-2.5-Flash finds no comparison applicable. Further, reassuringly, annotators agree on the sign of 95.5\% of the cells significant under both ($p < 0.05$).

\begin{figure}[t]
\centering
\begin{minipage}[t]{1\textwidth}
\centering
\sffamily
\tablefontsize
\setlength{\tabcolsep}{2pt}
\begin{tabular}{
    >{\raggedright\arraybackslash}m{0.22\linewidth}
    @{\hspace{10pt}}
    >{\centering\arraybackslash}m{25pt}
    @{\hspace{1pt}}
    >{\centering\arraybackslash}m{25pt}
    @{\hspace{10pt}}
    >{\centering\arraybackslash}m{25pt}
    @{\hspace{1pt}}
    >{\centering\arraybackslash}m{25pt}
    @{\hspace{10pt}}
    >{\centering\arraybackslash}m{25pt}
    @{\hspace{1pt}}
    >{\centering\arraybackslash}m{25pt}
    @{\hspace{10pt}}
    >{\centering\arraybackslash}m{25pt}
    @{\hspace{1pt}}
    >{\centering\arraybackslash}m{25pt}
    @{\hspace{10pt}}
    >{\centering\arraybackslash}m{0.075\linewidth}
}

\hfill\makecell[r]{\\\\\raisebox{3pt}{\color{gray}Model annotating \raisebox{-0.6ex}{$\searrow$}}} & \multicolumn{2}{@{}>{\centering\arraybackslash}m{51pt}@{\hspace{10pt}}}{\textbf{includes more ethical considerations}} & \multicolumn{2}{@{}>{\centering\arraybackslash}m{51pt}@{\hspace{10pt}}}{\textbf{uses more bold and italics text}} & \multicolumn{2}{@{}>{\centering\arraybackslash}m{51pt}@{\hspace{10pt}}}{\textbf{compliments the user's question or prompt}} & \multicolumn{2}{@{}>{\centering\arraybackslash}m{51pt}@{\hspace{10pt}}}{\textbf{uses more casual language}} & \\
\textbf{Model} & \textit{Gemini} & \textit{Claude} & \textit{Gemini} & \textit{Claude} & \textit{Gemini} & \textit{Claude} & \textit{Gemini} & \textit{Claude} & \textbf{Max diff} \\
\toprule
\rowcolor{altrow} GPT-5.1-Chat & \cellcolor{negcolor!10.1}-0.06 & \cellcolor{negcolor!19.5}-0.12 & \cellcolor{negcolor!78.3}-0.50 & \cellcolor{negcolor!100.0}-0.64 & \cellcolor{negcolor!15.7}-0.10 & \cellcolor{negcolor!20.4}-0.13 & \cellcolor{poscolor!4.0}\transparent{0.3}0.03 & \cellcolor{poscolor!85.2}0.63 & \cellcolor{lightgrey!100.0}0.60 \\
\addlinespace[\rowspacing]
GPT-5.3-Chat & \cellcolor{negcolor!6.0}\transparent{0.3}-0.04 & \cellcolor{negcolor!9.7}\transparent{0.3}-0.06 & \cellcolor{negcolor!67.6}-0.43 & \cellcolor{negcolor!85.2}-0.54 & \cellcolor{negcolor!0.3}\transparent{0.3}-0.00 & \cellcolor{negcolor!3.8}\transparent{0.3}-0.02 & \cellcolor{poscolor!44.5}0.33 & \cellcolor{poscolor!100.0}0.74 & \cellcolor{lightgrey!68.4}0.41 \\
\addlinespace[\rowspacing]
\rowcolor{altrow} Claude-3.7-Sonnet & \cellcolor{negcolor!5.7}\transparent{0.3}-0.04 & \cellcolor{negcolor!8.6}\transparent{0.3}-0.05 & \cellcolor{negcolor!66.3}-0.42 & \cellcolor{negcolor!28.7}-0.18 & \cellcolor{negcolor!11.5}-0.07 & \cellcolor{negcolor!13.1}-0.08 & \cellcolor{negcolor!4.5}\transparent{0.3}-0.03 & \cellcolor{poscolor!4.6}\transparent{0.3}0.03 & \cellcolor{lightgrey!39.8}0.24 \\
\addlinespace[\rowspacing]
Claude-Sonnet-4 & \cellcolor{poscolor!0.3}\transparent{0.3}0.00 & \cellcolor{poscolor!0.3}\transparent{0.3}0.00 & \cellcolor{poscolor!14.6}\transparent{0.3}0.11 & \cellcolor{poscolor!18.3}\transparent{0.3}0.14 & \cellcolor{negcolor!2.8}\transparent{0.3}-0.02 & \cellcolor{poscolor!2.2}\transparent{0.3}0.02 & \cellcolor{poscolor!5.9}\transparent{0.3}0.04 & \cellcolor{poscolor!29.1}0.22 & \cellcolor{lightgrey!28.6}0.17 \\
\addlinespace[\rowspacing]
\rowcolor{altrow} Claude-Opus-4.7 & \cellcolor{poscolor!1.6}\transparent{0.3}0.01 & \cellcolor{poscolor!6.0}\transparent{0.3}0.04 & \cellcolor{poscolor!45.7}0.34 & \cellcolor{poscolor!64.9}0.48 & \cellcolor{negcolor!2.2}\transparent{0.3}-0.01 & \cellcolor{negcolor!0.9}\transparent{0.3}-0.01 & \cellcolor{poscolor!16.2}0.12 & \cellcolor{poscolor!39.2}0.29 & \cellcolor{lightgrey!28.4}0.17 \\
\addlinespace[\rowspacing]
Claude-Sonnet-4.5 & \cellcolor{poscolor!0.3}\transparent{0.3}0.00 & \cellcolor{negcolor!1.3}\transparent{0.3}-0.01 & \cellcolor{poscolor!42.8}0.32 & \cellcolor{poscolor!64.1}0.48 & \cellcolor{negcolor!7.6}\transparent{0.3}-0.05 & \cellcolor{poscolor!3.0}\transparent{0.3}0.02 & \cellcolor{poscolor!9.2}0.07 & \cellcolor{poscolor!30.0}0.22 & \cellcolor{lightgrey!26.4}0.16 \\
\addlinespace[\rowspacing]
\rowcolor{altrow} Gemini-3.1-Pro & \cellcolor{poscolor!12.1}0.09 & \cellcolor{poscolor!13.4}0.10 & \cellcolor{poscolor!92.0}0.68 & \cellcolor{poscolor!87.3}0.65 & \cellcolor{poscolor!3.8}\transparent{0.3}0.03 & \cellcolor{poscolor!18.7}0.14 & \cellcolor{poscolor!9.9}0.07 & \cellcolor{poscolor!30.7}0.23 & \cellcolor{lightgrey!25.7}0.15 \\
\addlinespace[\rowspacing]
Gemini-3-Pro & \cellcolor{poscolor!8.2}0.06 & \cellcolor{negcolor!0.6}\transparent{0.3}-0.00 & \cellcolor{poscolor!87.8}0.65 & \cellcolor{poscolor!91.7}0.68 & \cellcolor{negcolor!4.2}\transparent{0.3}-0.03 & \cellcolor{poscolor!7.4}\transparent{0.3}0.05 & \cellcolor{poscolor!3.8}\transparent{0.3}0.03 & \cellcolor{poscolor!20.0}0.15 & \cellcolor{lightgrey!20.0}0.12 \\
\addlinespace[\rowspacing]
\rowcolor{altrow} GPT-5-Chat & \cellcolor{negcolor!2.8}\transparent{0.3}-0.02 & \cellcolor{poscolor!4.9}\transparent{0.3}0.04 & \cellcolor{poscolor!93.3}0.69 & \cellcolor{poscolor!84.1}0.62 & \cellcolor{poscolor!27.2}0.20 & \cellcolor{poscolor!35.0}0.26 & \cellcolor{poscolor!16.2}0.12 & \cellcolor{poscolor!12.4}\transparent{0.3}0.09 & \cellcolor{lightgrey!11.3}0.07 \\
\addlinespace[\rowspacing]
Gemini-2.5-Pro & \cellcolor{poscolor!14.3}0.11 & \cellcolor{poscolor!15.9}0.12 & \cellcolor{poscolor!96.0}0.71 & \cellcolor{poscolor!96.0}0.71 & \cellcolor{poscolor!73.6}0.55 & \cellcolor{poscolor!66.3}0.49 & \cellcolor{poscolor!7.8}0.06 & \cellcolor{poscolor!0.0}\transparent{0.3}0.00 & \cellcolor{lightgrey!9.6}0.06 \\
\end{tabular}
\end{minipage}
\caption{\textbf{Strength of the three traits in \Cref{fig:results:model_comparison} and \emph{uses more casual language} under two annotators.} Subcolumns \emph{Gemini} are annotated by Gemini-2.5-Flash and subcolumns \emph{Claude} by Claude-Haiku-4.5. Positive values are shown in {\color{blue}blue} and negative values in {\color{red}red}. Insignificant results greyed out ($p < 0.05 / (40 \times n)$ for a family of $n$ models). All measurements are relative to GPT-4o. Sorted by max difference between the annotators. An earlier version of \Cref{fig:results:model_comparison} showed \emph{uses more casual language}. We replaced that trait with \emph{includes more ethical considerations} because, for GPT models, the annotators disagreed regarding significance and magnitude on the original trait. Whilst this disagreement was not primarily caused by family-specific bias, we nevertheless chose to highlight a trait with more consensus across annotator families. Using cross-family judgement to better calibrate significance and magnitude would be an interesting direction for future work on Feedback Forensics.}
\label{fig:results:family_bias_selected_traits}
\end{figure}

\textbf{Analysis of results in \Cref{fig:results:model_comparison}.} \Cref{fig:results:family_bias_selected_traits} shows the numerical strength values underlying \Cref{fig:results:model_comparison} based on Gemini annotations, alongside Claude annotations. On the three traits in \Cref{fig:results:model_comparison}, the largest deviation between annotators on a Gemini model is 0.11, well inside the differences that figure draws conclusions from: on \emph{uses more bold and italics text}, both annotators place all three Gemini models near +0.7, while GPT-5.1-Chat and GPT-5.3-Chat sit near -0.5 under both. On these traits, we do not observe that Gemini values are inflated by the annotator being a Gemini model. \Cref{fig:results:family_bias_selected_traits} also reports \emph{uses more casual language}, which an earlier version of \Cref{fig:results:model_comparison} showed. On this trait, the annotators disagree most on GPT models (by up to 0.60 on GPT-5.1-Chat), which are outside both annotators' families.

\textbf{Lessons for interpreting Feedback Forensics results.} This effect does not overturn the large differences that \Cref{fig:results:model_comparison,fig:app:results:gemini_strength} rest on. However, when the annotator changes, the strength difference between two families on a single trait shifts by around 0.1 typically and by up to 0.54. A difference between families on one trait may therefore reflect the annotator rather than the models, and should be confirmed with a second annotator from another family. We also cannot tell whether an annotator prefers its own family or whether the two annotators simply read a trait differently in a way that aligns with one family's style, so we treat self-preference as a possible cause only. More sophisticated annotation approaches (e.g. human cross-check, multi-model) can help mitigate these effects and improve confidence in results.

\begin{figure}[p]
\centering
\begin{minipage}[t]{1\textwidth}
\centering
\sffamily
\tablefontsize
\setlength{\tabcolsep}{2pt}
\begin{tabular}{
    >{\raggedright\arraybackslash}m{0.33\linewidth}
    @{\hspace{13.5pt}}
    >{\centering\arraybackslash}m{0.13\linewidth}
    @{\hspace{13.5pt}}
    >{\centering\arraybackslash}m{0.13\linewidth}
    @{\hspace{13.5pt}}
    >{\centering\arraybackslash}m{0.1\linewidth}
}

\textbf{Generating a response that...} & \textbf{Gemini minus GPT} & \textbf{Claude minus GPT} & \textbf{Largest} \\
\toprule
\rowcolor{altrow} provides conclusions without full reasoning & \cellcolor{poscolor!100.0}0.54 & \cellcolor{poscolor!36.2}0.19 & \cellcolor{lightgrey!100.0}0.54 \\
\addlinespace[\rowspacing]
makes more confident statements & \cellcolor{poscolor!73.0}0.39 & \cellcolor{poscolor!7.1}\transparent{0.3}0.04 & \cellcolor{lightgrey!73.0}0.39 \\
\addlinespace[\rowspacing]
\rowcolor{altrow} more actively engages with the user & \cellcolor{negcolor!100.0}-0.39 & \cellcolor{negcolor!49.9}-0.19 & \cellcolor{lightgrey!72.0}0.39 \\
\addlinespace[\rowspacing]
provides more examples & \cellcolor{negcolor!90.5}-0.35 & \cellcolor{negcolor!56.1}-0.22 & \cellcolor{lightgrey!65.1}0.35 \\
\addlinespace[\rowspacing]
\rowcolor{altrow} is more concise & \cellcolor{poscolor!64.8}0.35 & \cellcolor{poscolor!2.3}\transparent{0.3}0.01 & \cellcolor{lightgrey!64.8}0.35 \\
\addlinespace[\rowspacing]
has a more avoidant tone & \cellcolor{poscolor!57.6}0.31 & \cellcolor{poscolor!25.9}0.14 & \cellcolor{lightgrey!57.6}0.31 \\
\addlinespace[\rowspacing]
\rowcolor{altrow} uses more casual language & \cellcolor{poscolor!47.8}0.26 & \cellcolor{poscolor!35.2}0.19 & \cellcolor{lightgrey!47.8}0.26 \\
\addlinespace[\rowspacing]
uses more bold and italics text & \cellcolor{negcolor!26.9}-0.10 & \cellcolor{negcolor!64.3}-0.25 & \cellcolor{lightgrey!46.3}0.25 \\
\addlinespace[\rowspacing]
\rowcolor{altrow} has more structured formatting & \cellcolor{negcolor!19.0}\transparent{0.3}-0.07 & \cellcolor{negcolor!61.5}-0.24 & \cellcolor{lightgrey!44.3}0.24 \\
\addlinespace[\rowspacing]
acknowledges own limitations or uncertainty more & \cellcolor{poscolor!43.3}0.23 & \cellcolor{poscolor!20.8}0.11 & \cellcolor{lightgrey!43.3}0.23 \\
\addlinespace[\rowspacing]
\rowcolor{altrow} is more polite & \cellcolor{poscolor!18.2}0.10 & \cellcolor{negcolor!46.9}-0.18 & \cellcolor{lightgrey!33.7}0.18 \\
\addlinespace[\rowspacing]
more strictly follows the requested output format & \cellcolor{negcolor!31.8}-0.12 & \cellcolor{negcolor!46.0}-0.18 & \cellcolor{lightgrey!33.1}0.18 \\
\addlinespace[\rowspacing]
\rowcolor{altrow} provides a numbered list format & \cellcolor{negcolor!45.1}-0.17 & \cellcolor{negcolor!42.0}-0.16 & \cellcolor{lightgrey!32.4}0.17 \\
\addlinespace[\rowspacing]
is more empathetic to the user & \cellcolor{poscolor!30.1}0.16 & \cellcolor{poscolor!1.6}\transparent{0.3}0.01 & \cellcolor{lightgrey!30.1}0.16 \\
\addlinespace[\rowspacing]
\rowcolor{altrow} expresses more emotion & \cellcolor{negcolor!32.1}-0.12 & \cellcolor{poscolor!8.0}\transparent{0.3}0.04 & \cellcolor{lightgrey!23.1}0.12 \\
\addlinespace[\rowspacing]
uses more humour & \cellcolor{negcolor!1.8}\transparent{0.3}-0.01 & \cellcolor{poscolor!22.9}0.12 & \cellcolor{lightgrey!22.9}0.12 \\
\addlinespace[\rowspacing]
\rowcolor{altrow} contains less harmful information & \cellcolor{poscolor!22.2}0.12 & \cellcolor{poscolor!1.0}\transparent{0.3}0.01 & \cellcolor{lightgrey!22.2}0.12 \\
\addlinespace[\rowspacing]
is more factually correct & \cellcolor{negcolor!2.3}\transparent{0.3}-0.01 & \cellcolor{negcolor!25.1}-0.10 & \cellcolor{lightgrey!18.0}0.10 \\
\addlinespace[\rowspacing]
\rowcolor{altrow} is more creative and original & \cellcolor{negcolor!23.6}-0.09 & \cellcolor{negcolor!1.9}\transparent{0.3}-0.01 & \cellcolor{lightgrey!17.0}0.09 \\
\addlinespace[\rowspacing]
uses more formal language & \cellcolor{poscolor!16.3}\transparent{0.3}0.09 & \cellcolor{negcolor!0.6}\transparent{0.3}-0.00 & \cellcolor{lightgrey!16.3}0.09 \\
\addlinespace[\rowspacing]
\rowcolor{altrow} has a friendlier tone & \cellcolor{poscolor!15.9}\transparent{0.3}0.09 & \cellcolor{negcolor!18.1}\transparent{0.3}-0.07 & \cellcolor{lightgrey!15.9}0.09 \\
\addlinespace[\rowspacing]
uses more personal pronouns (I, we, you) & \cellcolor{negcolor!17.1}\transparent{0.3}-0.07 & \cellcolor{poscolor!13.8}\transparent{0.3}0.07 & \cellcolor{lightgrey!13.8}0.07 \\
\addlinespace[\rowspacing]
\rowcolor{altrow} is more optimistic & \cellcolor{negcolor!8.4}\transparent{0.3}-0.03 & \cellcolor{negcolor!16.2}\transparent{0.3}-0.06 & \cellcolor{lightgrey!11.6}0.06 \\
\addlinespace[\rowspacing]
ends with a follow-up question & \cellcolor{negcolor!15.3}\transparent{0.3}-0.06 & \cellcolor{negcolor!10.1}\transparent{0.3}-0.04 & \cellcolor{lightgrey!11.0}0.06 \\
\addlinespace[\rowspacing]
\rowcolor{altrow} uses a more enthusiastic tone & \cellcolor{poscolor!4.3}\transparent{0.3}0.02 & \cellcolor{poscolor!10.9}\transparent{0.3}0.06 & \cellcolor{lightgrey!10.9}0.06 \\
\addlinespace[\rowspacing]
actively engages the reader with rhetorical questions & \cellcolor{poscolor!7.2}\transparent{0.3}0.04 & \cellcolor{poscolor!8.2}\transparent{0.3}0.04 & \cellcolor{lightgrey!8.2}0.04 \\
\addlinespace[\rowspacing]
\rowcolor{altrow} compliments the user's question or prompt & \cellcolor{negcolor!11.4}\transparent{0.3}-0.04 & \cellcolor{negcolor!6.1}\transparent{0.3}-0.02 & \cellcolor{lightgrey!8.2}0.04 \\
\addlinespace[\rowspacing]
is more verbose & \cellcolor{negcolor!10.4}\transparent{0.3}-0.04 & \cellcolor{poscolor!5.9}\transparent{0.3}0.03 & \cellcolor{lightgrey!7.5}0.04 \\
\addlinespace[\rowspacing]
\rowcolor{altrow} includes more references to other sources & \cellcolor{negcolor!1.2}\transparent{0.3}-0.00 & \cellcolor{negcolor!8.4}\transparent{0.3}-0.03 & \cellcolor{lightgrey!6.1}0.03 \\
\addlinespace[\rowspacing]
agrees more with the user & \cellcolor{poscolor!2.5}\transparent{0.3}0.01 & \cellcolor{poscolor!5.8}\transparent{0.3}0.03 & \cellcolor{lightgrey!5.8}0.03 \\
\addlinespace[\rowspacing]
\rowcolor{altrow} uses more emojis & \cellcolor{poscolor!3.0}\transparent{0.3}0.02 & \cellcolor{poscolor!3.6}\transparent{0.3}0.02 & \cellcolor{lightgrey!3.6}0.02 \\
\addlinespace[\rowspacing]
uses more mathematical symbols and notation & \cellcolor{negcolor!0.9}\transparent{0.3}-0.00 & \cellcolor{negcolor!4.9}\transparent{0.3}-0.02 & \cellcolor{lightgrey!3.5}0.02 \\
\addlinespace[\rowspacing]
\rowcolor{altrow} reinforces user's beliefs more & \cellcolor{negcolor!2.9}\transparent{0.3}-0.01 & \cellcolor{poscolor!3.0}\transparent{0.3}0.02 & \cellcolor{lightgrey!3.0}0.02 \\
\addlinespace[\rowspacing]
suggests illegal activities & \cellcolor{negcolor!3.3}-0.01 & \cellcolor{negcolor!0.4}\transparent{0.3}-0.00 & \cellcolor{lightgrey!2.4}0.01 \\
\addlinespace[\rowspacing]
\rowcolor{altrow} agrees with user even if factually incorrect & \cellcolor{negcolor!3.0}\transparent{0.3}-0.01 & \cellcolor{negcolor!0.0}\transparent{0.3}-0.00 & \cellcolor{lightgrey!2.1}0.01 \\
\addlinespace[\rowspacing]
includes more ethical considerations & \cellcolor{poscolor!0.8}\transparent{0.3}0.00 & \cellcolor{negcolor!2.8}\transparent{0.3}-0.01 & \cellcolor{lightgrey!2.0}0.01 \\
\addlinespace[\rowspacing]
\rowcolor{altrow} is more offensive & \cellcolor{negcolor!2.3}\transparent{0.3}-0.01 & \cellcolor{negcolor!0.8}\transparent{0.3}-0.00 & \cellcolor{lightgrey!1.6}0.01 \\
\addlinespace[\rowspacing]
includes inappropriate language & \cellcolor{negcolor!1.6}\transparent{0.3}-0.01 & \cellcolor{negcolor!0.5}\transparent{0.3}-0.00 & \cellcolor{lightgrey!1.1}0.01 \\
\addlinespace[\rowspacing]
\rowcolor{altrow} refuses to answer the question & \cellcolor{poscolor!0.1}\transparent{0.3}0.00 & \cellcolor{negcolor!1.3}\transparent{0.3}-0.00 & \cellcolor{lightgrey!0.9}0.00 \\
\addlinespace[\rowspacing]
reinforces user's anger more & \cellcolor{poscolor!0.1}\transparent{0.3}0.00 & \cellcolor{negcolor!0.3}\transparent{0.3}-0.00 & \cellcolor{lightgrey!0.2}0.00 \\
\midrule
\textit{Mean} & 0.02 & -0.02 & \\
\textit{Mean absolute} & 0.11 & 0.07 & \\
\end{tabular}
\end{minipage}
\caption{\textbf{Family contrast per trait.} A family's contrast is its mean strength gap minus that of the GPT models, where the gap is Gemini-2.5-Flash's strength minus Claude-Haiku-4.5's. Own-family bias predicts a positive Gemini and a negative Claude contrast. Positive values are shown in {\color{blue}blue} and negative values in {\color{red}red}. Insignificant results greyed out (prompt-level bootstrap, Bonferroni correction over 80 tests). Sorted by largest absolute contrast.}
\label{fig:app:results:family_contrasts}
\end{figure}

\begin{figure}[p]
\centering
\begin{minipage}[t]{1\textwidth}
\centering
\sffamily
\tablefontsize
\setlength{\tabcolsep}{2pt}
\begin{tabular}{
    >{\raggedright\arraybackslash}m{0.3\linewidth}
    @{\hspace{7pt}}
    >{\centering\arraybackslash}m{0.06\linewidth}
    @{\hspace{1pt}}
    >{\centering\arraybackslash}m{0.06\linewidth}
    @{\hspace{7pt}}
    >{\centering\arraybackslash}m{0.06\linewidth}
    @{\hspace{1pt}}
    >{\centering\arraybackslash}m{0.06\linewidth}
    @{\hspace{7pt}}
    >{\centering\arraybackslash}m{0.06\linewidth}
    @{\hspace{1pt}}
    >{\centering\arraybackslash}m{0.06\linewidth}
    @{\hspace{7pt}}
    >{\centering\arraybackslash}m{0.06\linewidth}
    @{\hspace{1pt}}
    >{\centering\arraybackslash}m{0.06\linewidth}
    @{\hspace{7pt}}
    >{\centering\arraybackslash}m{0.075\linewidth}
}

\hfill\makecell[r]{{\color{gray}Model tested $\rightarrow$}\\\\\raisebox{3pt}{\color{gray}Model annotating \raisebox{-0.6ex}{$\searrow$}}} & \multicolumn{2}{@{}c@{\hspace{7pt}}}{\textbf{\makecell{Anthropic\\\textit{Claude-3.7-}\\\textit{Sonnet}}}} & \multicolumn{2}{@{}c@{\hspace{7pt}}}{\textbf{\makecell{Anthropic\\\textit{Claude-}\\\textit{Sonnet-4}}}} & \multicolumn{2}{@{}c@{\hspace{7pt}}}{\textbf{\makecell{Anthropic\\\textit{Claude-}\\\textit{Sonnet-4.5}}}} & \multicolumn{2}{@{}c@{\hspace{7pt}}}{\textbf{\makecell{Anthropic\\\textit{Claude-}\\\textit{Opus-4.7}}}} & \\
\textbf{Generating a response that...} & \textit{Gemini} & \textit{Claude} & \textit{Gemini} & \textit{Claude} & \textit{Gemini} & \textit{Claude} & \textit{Gemini} & \textit{Claude} & \textbf{Max diff} \\
\toprule
\rowcolor{altrow} is more creative and original & \cellcolor{poscolor!7.7}\transparent{0.3}0.05 & \cellcolor{poscolor!29.1}0.19 & \cellcolor{poscolor!17.7}0.12 & \cellcolor{poscolor!70.6}0.46 & \cellcolor{poscolor!21.5}0.14 & \cellcolor{poscolor!79.8}0.52 & \cellcolor{poscolor!30.4}0.20 & \cellcolor{poscolor!100.0}0.65 & \cellcolor{lightgrey!100.0}0.46 \\
\addlinespace[\rowspacing]
uses more formal language & \cellcolor{negcolor!16.7}-0.10 & \cellcolor{negcolor!42.3}-0.24 & \cellcolor{negcolor!25.3}-0.14 & \cellcolor{negcolor!79.6}-0.45 & \cellcolor{negcolor!44.4}-0.25 & \cellcolor{negcolor!100.0}-0.57 & \cellcolor{negcolor!27.8}-0.16 & \cellcolor{negcolor!90.8}-0.52 & \cellcolor{lightgrey!78.9}0.36 \\
\addlinespace[\rowspacing]
\rowcolor{altrow} has a friendlier tone & \cellcolor{negcolor!29.9}-0.17 & \cellcolor{negcolor!22.8}-0.13 & \cellcolor{negcolor!2.8}\transparent{0.3}-0.02 & \cellcolor{poscolor!36.7}0.24 & \cellcolor{negcolor!15.5}-0.09 & \cellcolor{poscolor!34.4}0.22 & \cellcolor{negcolor!2.1}\transparent{0.3}-0.01 & \cellcolor{poscolor!44.2}0.29 & \cellcolor{lightgrey!68.7}0.31 \\
\addlinespace[\rowspacing]
has more structured formatting & \cellcolor{negcolor!37.0}-0.21 & \cellcolor{poscolor!5.9}\transparent{0.3}0.04 & \cellcolor{poscolor!7.3}\transparent{0.3}0.05 & \cellcolor{poscolor!36.4}0.24 & \cellcolor{poscolor!29.4}0.19 & \cellcolor{poscolor!71.5}0.47 & \cellcolor{poscolor!46.3}0.30 & \cellcolor{poscolor!78.2}0.51 & \cellcolor{lightgrey!60.4}0.28 \\
\addlinespace[\rowspacing]
\rowcolor{altrow} uses more humour & \cellcolor{poscolor!1.9}\transparent{0.3}0.01 & \cellcolor{poscolor!0.6}\transparent{0.3}0.00 & \cellcolor{poscolor!5.2}0.03 & \cellcolor{poscolor!12.8}0.08 & \cellcolor{poscolor!5.5}0.04 & \cellcolor{poscolor!30.7}0.20 & \cellcolor{poscolor!8.9}0.06 & \cellcolor{poscolor!48.5}0.32 & \cellcolor{lightgrey!56.8}0.26 \\
\addlinespace[\rowspacing]
more strictly follows the requested output format & \cellcolor{negcolor!6.8}\transparent{0.3}-0.04 & \cellcolor{poscolor!13.6}\transparent{0.3}0.09 & \cellcolor{negcolor!9.5}\transparent{0.3}-0.05 & \cellcolor{poscolor!8.6}\transparent{0.3}0.06 & \cellcolor{poscolor!0.6}\transparent{0.3}0.00 & \cellcolor{poscolor!39.9}0.26 & \cellcolor{poscolor!5.2}\transparent{0.3}0.03 & \cellcolor{poscolor!25.5}0.17 & \cellcolor{lightgrey!56.4}0.26 \\
\addlinespace[\rowspacing]
\rowcolor{altrow} uses more bold and italics text & \cellcolor{negcolor!74.0}-0.42 & \cellcolor{negcolor!32.0}-0.18 & \cellcolor{poscolor!16.5}\transparent{0.3}0.11 & \cellcolor{poscolor!20.8}\transparent{0.3}0.14 & \cellcolor{poscolor!48.5}0.32 & \cellcolor{poscolor!72.7}0.48 & \cellcolor{poscolor!51.8}0.34 & \cellcolor{poscolor!73.6}0.48 & \cellcolor{lightgrey!52.5}0.24 \\
\addlinespace[\rowspacing]
actively engages the reader with rhetorical questions & \cellcolor{negcolor!0.7}\transparent{0.3}-0.00 & \cellcolor{poscolor!4.6}\transparent{0.3}0.03 & \cellcolor{poscolor!11.3}0.07 & \cellcolor{poscolor!25.4}0.17 & \cellcolor{poscolor!16.6}0.11 & \cellcolor{poscolor!32.8}0.21 & \cellcolor{poscolor!15.3}0.10 & \cellcolor{poscolor!50.3}0.33 & \cellcolor{lightgrey!50.2}0.23 \\
\addlinespace[\rowspacing]
\rowcolor{altrow} is more factually correct & \cellcolor{poscolor!0.0}\transparent{0.3}0.00 & \cellcolor{poscolor!13.6}\transparent{0.3}0.09 & \cellcolor{poscolor!5.8}\transparent{0.3}0.04 & \cellcolor{poscolor!30.6}0.20 & \cellcolor{poscolor!8.6}\transparent{0.3}0.06 & \cellcolor{poscolor!37.1}0.24 & \cellcolor{poscolor!28.2}0.18 & \cellcolor{poscolor!63.2}0.41 & \cellcolor{lightgrey!50.2}0.23 \\
\addlinespace[\rowspacing]
is more polite & \cellcolor{negcolor!38.1}-0.22 & \cellcolor{negcolor!41.6}-0.24 & \cellcolor{negcolor!17.9}-0.10 & \cellcolor{negcolor!36.8}-0.21 & \cellcolor{negcolor!40.5}-0.23 & \cellcolor{negcolor!65.5}-0.37 & \cellcolor{negcolor!26.8}-0.15 & \cellcolor{negcolor!63.7}-0.36 & \cellcolor{lightgrey!46.3}0.21 \\
\addlinespace[\rowspacing]
\rowcolor{altrow} more actively engages with the user & \cellcolor{negcolor!14.2}\transparent{0.3}-0.08 & \cellcolor{negcolor!21.0}-0.12 & \cellcolor{poscolor!18.3}0.12 & \cellcolor{poscolor!48.6}0.32 & \cellcolor{poscolor!12.9}\transparent{0.3}0.08 & \cellcolor{poscolor!44.5}0.29 & \cellcolor{poscolor!41.7}0.27 & \cellcolor{poscolor!63.5}0.42 & \cellcolor{lightgrey!45.4}0.21 \\
\addlinespace[\rowspacing]
uses a more enthusiastic tone & \cellcolor{negcolor!13.9}-0.08 & \cellcolor{negcolor!25.6}-0.15 & \cellcolor{poscolor!4.0}\transparent{0.3}0.03 & \cellcolor{poscolor!15.6}0.10 & \cellcolor{poscolor!3.4}\transparent{0.3}0.02 & \cellcolor{poscolor!31.3}0.20 & \cellcolor{poscolor!8.6}\transparent{0.3}0.06 & \cellcolor{poscolor!30.1}0.20 & \cellcolor{lightgrey!40.1}0.18 \\
\addlinespace[\rowspacing]
\rowcolor{altrow} provides a numbered list format & \cellcolor{negcolor!51.6}-0.29 & \cellcolor{negcolor!38.4}-0.22 & \cellcolor{negcolor!38.6}-0.22 & \cellcolor{negcolor!21.4}-0.12 & \cellcolor{negcolor!38.7}-0.22 & \cellcolor{negcolor!12.7}\transparent{0.3}-0.07 & \cellcolor{negcolor!46.8}-0.27 & \cellcolor{negcolor!15.1}\transparent{0.3}-0.09 & \cellcolor{lightgrey!39.6}0.18 \\
\addlinespace[\rowspacing]
provides conclusions without full reasoning & \cellcolor{poscolor!0.6}\transparent{0.3}0.00 & \cellcolor{poscolor!27.9}0.18 & \cellcolor{negcolor!0.4}\transparent{0.3}-0.00 & \cellcolor{negcolor!4.2}\transparent{0.3}-0.02 & \cellcolor{poscolor!0.9}\transparent{0.3}0.01 & \cellcolor{negcolor!6.7}\transparent{0.3}-0.04 & \cellcolor{poscolor!1.5}\transparent{0.3}0.01 & \cellcolor{negcolor!21.1}-0.12 & \cellcolor{lightgrey!39.2}0.18 \\
\addlinespace[\rowspacing]
\rowcolor{altrow} expresses more emotion & \cellcolor{negcolor!6.0}-0.03 & \cellcolor{negcolor!6.4}\transparent{0.3}-0.04 & \cellcolor{poscolor!2.1}\transparent{0.3}0.01 & \cellcolor{poscolor!12.8}0.08 & \cellcolor{poscolor!3.7}\transparent{0.3}0.02 & \cellcolor{poscolor!22.1}0.14 & \cellcolor{poscolor!5.5}\transparent{0.3}0.04 & \cellcolor{poscolor!32.5}0.21 & \cellcolor{lightgrey!38.8}0.18 \\
\addlinespace[\rowspacing]
uses more casual language & \cellcolor{negcolor!5.0}\transparent{0.3}-0.03 & \cellcolor{poscolor!5.3}\transparent{0.3}0.03 & \cellcolor{poscolor!6.7}\transparent{0.3}0.04 & \cellcolor{poscolor!33.0}0.22 & \cellcolor{poscolor!10.4}0.07 & \cellcolor{poscolor!34.0}0.22 & \cellcolor{poscolor!18.4}0.12 & \cellcolor{poscolor!44.5}0.29 & \cellcolor{lightgrey!37.7}0.17 \\
\addlinespace[\rowspacing]
\rowcolor{altrow} makes more confident statements & \cellcolor{poscolor!0.3}\transparent{0.3}0.00 & \cellcolor{poscolor!19.2}\transparent{0.3}0.13 & \cellcolor{poscolor!10.7}\transparent{0.3}0.07 & \cellcolor{poscolor!23.8}0.16 & \cellcolor{poscolor!20.9}0.14 & \cellcolor{poscolor!45.1}0.30 & \cellcolor{poscolor!30.7}0.20 & \cellcolor{poscolor!43.6}0.29 & \cellcolor{lightgrey!34.8}0.16 \\
\addlinespace[\rowspacing]
has a more avoidant tone & \cellcolor{negcolor!1.8}\transparent{0.3}-0.01 & \cellcolor{poscolor!2.8}\transparent{0.3}0.02 & \cellcolor{negcolor!0.7}\transparent{0.3}-0.00 & \cellcolor{negcolor!12.6}-0.07 & \cellcolor{negcolor!1.1}\transparent{0.3}-0.01 & \cellcolor{negcolor!21.8}-0.12 & \cellcolor{negcolor!2.5}\transparent{0.3}-0.01 & \cellcolor{negcolor!25.4}-0.14 & \cellcolor{lightgrey!28.6}0.13 \\
\addlinespace[\rowspacing]
\rowcolor{altrow} uses more personal pronouns (I, we, you) & \cellcolor{negcolor!2.5}\transparent{0.3}-0.01 & \cellcolor{negcolor!23.1}\transparent{0.3}-0.13 & \cellcolor{poscolor!24.7}0.16 & \cellcolor{poscolor!37.0}0.24 & \cellcolor{poscolor!8.6}\transparent{0.3}0.06 & \cellcolor{poscolor!16.3}\transparent{0.3}0.11 & \cellcolor{poscolor!16.9}0.11 & \cellcolor{poscolor!31.0}0.20 & \cellcolor{lightgrey!25.8}0.12 \\
\addlinespace[\rowspacing]
provides more examples & \cellcolor{negcolor!22.4}-0.13 & \cellcolor{negcolor!26.7}-0.15 & \cellcolor{poscolor!21.1}0.14 & \cellcolor{poscolor!32.7}0.21 & \cellcolor{poscolor!21.2}0.14 & \cellcolor{poscolor!28.8}0.19 & \cellcolor{poscolor!32.5}0.21 & \cellcolor{poscolor!49.4}0.32 & \cellcolor{lightgrey!24.2}0.11 \\
\addlinespace[\rowspacing]
\rowcolor{altrow} is more verbose & \cellcolor{negcolor!45.9}-0.26 & \cellcolor{negcolor!48.7}-0.28 & \cellcolor{poscolor!13.4}\transparent{0.3}0.09 & \cellcolor{poscolor!0.6}\transparent{0.3}0.00 & \cellcolor{negcolor!7.0}\transparent{0.3}-0.04 & \cellcolor{negcolor!21.1}\transparent{0.3}-0.12 & \cellcolor{poscolor!13.8}\transparent{0.3}0.09 & \cellcolor{negcolor!1.1}\transparent{0.3}-0.01 & \cellcolor{lightgrey!21.1}0.10 \\
\addlinespace[\rowspacing]
is more concise & \cellcolor{poscolor!26.0}0.17 & \cellcolor{poscolor!39.4}0.26 & \cellcolor{negcolor!11.9}\transparent{0.3}-0.07 & \cellcolor{negcolor!7.7}\transparent{0.3}-0.04 & \cellcolor{poscolor!2.5}\transparent{0.3}0.02 & \cellcolor{poscolor!2.5}\transparent{0.3}0.02 & \cellcolor{negcolor!6.3}\transparent{0.3}-0.04 & \cellcolor{negcolor!15.1}\transparent{0.3}-0.09 & \cellcolor{lightgrey!19.1}0.09 \\
\addlinespace[\rowspacing]
\rowcolor{altrow} is more optimistic & \cellcolor{negcolor!2.5}\transparent{0.3}-0.01 & \cellcolor{negcolor!4.6}\transparent{0.3}-0.03 & \cellcolor{poscolor!0.6}\transparent{0.3}0.00 & \cellcolor{poscolor!11.0}\transparent{0.3}0.07 & \cellcolor{negcolor!1.4}\transparent{0.3}-0.01 & \cellcolor{poscolor!12.0}\transparent{0.3}0.08 & \cellcolor{negcolor!1.4}\transparent{0.3}-0.01 & \cellcolor{poscolor!8.3}\transparent{0.3}0.05 & \cellcolor{lightgrey!18.9}0.09 \\
\addlinespace[\rowspacing]
is more empathetic to the user & \cellcolor{negcolor!14.9}-0.09 & \cellcolor{negcolor!11.4}\transparent{0.3}-0.06 & \cellcolor{poscolor!6.1}\transparent{0.3}0.04 & \cellcolor{poscolor!17.1}0.11 & \cellcolor{negcolor!5.3}\transparent{0.3}-0.03 & \cellcolor{poscolor!4.0}\transparent{0.3}0.03 & \cellcolor{poscolor!3.1}\transparent{0.3}0.02 & \cellcolor{poscolor!13.5}\transparent{0.3}0.09 & \cellcolor{lightgrey!15.8}0.07 \\
\addlinespace[\rowspacing]
\rowcolor{altrow} compliments the user's question or prompt & \cellcolor{negcolor!12.8}-0.07 & \cellcolor{negcolor!14.6}-0.08 & \cellcolor{negcolor!3.2}\transparent{0.3}-0.02 & \cellcolor{poscolor!2.4}\transparent{0.3}0.02 & \cellcolor{negcolor!8.5}\transparent{0.3}-0.05 & \cellcolor{poscolor!3.4}\transparent{0.3}0.02 & \cellcolor{negcolor!2.5}\transparent{0.3}-0.01 & \cellcolor{negcolor!1.1}\transparent{0.3}-0.01 & \cellcolor{lightgrey!15.4}0.07 \\
\addlinespace[\rowspacing]
acknowledges own limitations or uncertainty more & \cellcolor{negcolor!3.6}\transparent{0.3}-0.02 & \cellcolor{negcolor!3.2}\transparent{0.3}-0.02 & \cellcolor{poscolor!3.7}\transparent{0.3}0.02 & \cellcolor{poscolor!6.7}\transparent{0.3}0.04 & \cellcolor{poscolor!2.1}\transparent{0.3}0.01 & \cellcolor{negcolor!7.7}\transparent{0.3}-0.04 & \cellcolor{poscolor!3.1}\transparent{0.3}0.02 & \cellcolor{poscolor!10.4}\transparent{0.3}0.07 & \cellcolor{lightgrey!12.8}0.06 \\
\addlinespace[\rowspacing]
\rowcolor{altrow} contains less harmful information & \cellcolor{poscolor!0.6}\transparent{0.3}0.00 & \cellcolor{poscolor!5.3}\transparent{0.3}0.03 & \cellcolor{poscolor!1.2}\transparent{0.3}0.01 & \cellcolor{poscolor!7.0}0.05 & \cellcolor{poscolor!0.6}\transparent{0.3}0.00 & \cellcolor{poscolor!4.3}\transparent{0.3}0.03 & \cellcolor{poscolor!0.3}\transparent{0.3}0.00 & \cellcolor{poscolor!6.1}\transparent{0.3}0.04 & \cellcolor{lightgrey!8.4}0.04 \\
\addlinespace[\rowspacing]
ends with a follow-up question & \cellcolor{negcolor!5.7}\transparent{0.3}-0.03 & \cellcolor{negcolor!10.3}\transparent{0.3}-0.06 & \cellcolor{poscolor!9.2}\transparent{0.3}0.06 & \cellcolor{poscolor!14.7}0.10 & \cellcolor{poscolor!9.2}\transparent{0.3}0.06 & \cellcolor{poscolor!13.8}0.09 & \cellcolor{poscolor!35.9}0.23 & \cellcolor{poscolor!38.7}0.25 & \cellcolor{lightgrey!7.9}0.04 \\
\addlinespace[\rowspacing]
\rowcolor{altrow} includes more ethical considerations & \cellcolor{negcolor!6.4}\transparent{0.3}-0.04 & \cellcolor{negcolor!9.6}\transparent{0.3}-0.05 & \cellcolor{poscolor!0.3}\transparent{0.3}0.00 & \cellcolor{poscolor!0.3}\transparent{0.3}0.00 & \cellcolor{poscolor!0.3}\transparent{0.3}0.00 & \cellcolor{negcolor!1.4}\transparent{0.3}-0.01 & \cellcolor{poscolor!1.8}\transparent{0.3}0.01 & \cellcolor{poscolor!6.7}\transparent{0.3}0.04 & \cellcolor{lightgrey!7.0}0.03 \\
\addlinespace[\rowspacing]
uses more mathematical symbols and notation & \cellcolor{negcolor!13.5}-0.08 & \cellcolor{negcolor!9.6}-0.05 & \cellcolor{negcolor!13.7}-0.08 & \cellcolor{negcolor!8.1}\transparent{0.3}-0.05 & \cellcolor{negcolor!11.3}-0.06 & \cellcolor{negcolor!6.7}\transparent{0.3}-0.04 & \cellcolor{negcolor!3.2}\transparent{0.3}-0.02 & \cellcolor{poscolor!0.6}\transparent{0.3}0.00 & \cellcolor{lightgrey!7.0}0.03 \\
\addlinespace[\rowspacing]
\rowcolor{altrow} includes more references to other sources & \cellcolor{negcolor!2.5}\transparent{0.3}-0.01 & \cellcolor{negcolor!2.5}\transparent{0.3}-0.01 & \cellcolor{poscolor!0.3}\transparent{0.3}0.00 & \cellcolor{poscolor!3.4}\transparent{0.3}0.02 & \cellcolor{negcolor!0.4}\transparent{0.3}-0.00 & \cellcolor{poscolor!1.2}\transparent{0.3}0.01 & \cellcolor{poscolor!13.5}0.09 & \cellcolor{poscolor!15.3}0.10 & \cellcolor{lightgrey!4.4}0.02 \\
\addlinespace[\rowspacing]
agrees more with the user & \cellcolor{negcolor!0.4}\transparent{0.3}-0.00 & \cellcolor{negcolor!2.1}\transparent{0.3}-0.01 & \cellcolor{poscolor!0.6}\transparent{0.3}0.00 & \cellcolor{poscolor!3.4}\transparent{0.3}0.02 & \cellcolor{negcolor!3.5}\transparent{0.3}-0.02 & \cellcolor{negcolor!0.7}\transparent{0.3}-0.00 & \cellcolor{negcolor!2.1}\transparent{0.3}-0.01 & \cellcolor{poscolor!0.9}\transparent{0.3}0.01 & \cellcolor{lightgrey!4.0}0.02 \\
\addlinespace[\rowspacing]
\rowcolor{altrow} reinforces user's beliefs more & \cellcolor{negcolor!2.8}\transparent{0.3}-0.02 & \cellcolor{poscolor!0.0}\transparent{0.3}0.00 & \cellcolor{poscolor!0.0}\transparent{0.3}0.00 & \cellcolor{poscolor!0.6}\transparent{0.3}0.00 & \cellcolor{negcolor!1.1}\transparent{0.3}-0.01 & \cellcolor{poscolor!0.9}\transparent{0.3}0.01 & \cellcolor{negcolor!0.7}\transparent{0.3}-0.00 & \cellcolor{negcolor!0.4}\transparent{0.3}-0.00 & \cellcolor{lightgrey!3.6}0.02 \\
\addlinespace[\rowspacing]
uses more emojis & \cellcolor{negcolor!5.0}\transparent{0.3}-0.03 & \cellcolor{negcolor!5.3}\transparent{0.3}-0.03 & \cellcolor{poscolor!1.8}\transparent{0.3}0.01 & \cellcolor{poscolor!3.7}\transparent{0.3}0.02 & \cellcolor{poscolor!7.1}\transparent{0.3}0.05 & \cellcolor{poscolor!6.7}\transparent{0.3}0.04 & \cellcolor{poscolor!8.6}0.06 & \cellcolor{poscolor!11.0}0.07 & \cellcolor{lightgrey!3.5}0.02 \\
\addlinespace[\rowspacing]
\rowcolor{altrow} agrees with user even if factually incorrect & \cellcolor{poscolor!0.3}\transparent{0.3}0.00 & \cellcolor{poscolor!1.2}\transparent{0.3}0.01 & \cellcolor{negcolor!1.1}\transparent{0.3}-0.01 & \cellcolor{negcolor!1.4}\transparent{0.3}-0.01 & \cellcolor{negcolor!2.8}\transparent{0.3}-0.02 & \cellcolor{negcolor!2.8}\transparent{0.3}-0.02 & \cellcolor{negcolor!0.7}\transparent{0.3}-0.00 & \cellcolor{negcolor!1.8}\transparent{0.3}-0.01 & \cellcolor{lightgrey!1.3}0.01 \\
\addlinespace[\rowspacing]
suggests illegal activities & \cellcolor{poscolor!0.3}\transparent{0.3}0.00 & \cellcolor{negcolor!0.4}\transparent{0.3}-0.00 & \cellcolor{negcolor!0.4}\transparent{0.3}-0.00 & \cellcolor{negcolor!1.4}\transparent{0.3}-0.01 & \cellcolor{poscolor!0.0}\transparent{0.3}0.00 & \cellcolor{negcolor!0.4}\transparent{0.3}-0.00 & \cellcolor{poscolor!0.0}\transparent{0.3}0.00 & \cellcolor{poscolor!0.3}\transparent{0.3}0.00 & \cellcolor{lightgrey!1.3}0.01 \\
\addlinespace[\rowspacing]
\rowcolor{altrow} reinforces user's anger more & \cellcolor{negcolor!0.4}\transparent{0.3}-0.00 & \cellcolor{poscolor!0.3}\transparent{0.3}0.00 & \cellcolor{poscolor!0.0}\transparent{0.3}0.00 & \cellcolor{poscolor!0.3}\transparent{0.3}0.00 & \cellcolor{poscolor!0.0}\transparent{0.3}0.00 & \cellcolor{poscolor!0.3}\transparent{0.3}0.00 & \cellcolor{poscolor!0.3}\transparent{0.3}0.00 & \cellcolor{poscolor!0.6}\transparent{0.3}0.00 & \cellcolor{lightgrey!0.9}0.00 \\
\addlinespace[\rowspacing]
refuses to answer the question & \cellcolor{negcolor!1.4}\transparent{0.3}-0.01 & \cellcolor{negcolor!1.1}\transparent{0.3}-0.01 & \cellcolor{poscolor!1.2}\transparent{0.3}0.01 & \cellcolor{poscolor!0.6}\transparent{0.3}0.00 & \cellcolor{negcolor!0.4}\transparent{0.3}-0.00 & \cellcolor{poscolor!0.3}\transparent{0.3}0.00 & \cellcolor{poscolor!0.0}\transparent{0.3}0.00 & \cellcolor{poscolor!0.6}\transparent{0.3}0.00 & \cellcolor{lightgrey!0.9}0.00 \\
\addlinespace[\rowspacing]
\rowcolor{altrow} includes inappropriate language & \cellcolor{poscolor!0.0}\transparent{0.3}0.00 & \cellcolor{negcolor!0.4}\transparent{0.3}-0.00 & \cellcolor{poscolor!0.0}\transparent{0.3}0.00 & \cellcolor{negcolor!0.7}\transparent{0.3}-0.00 & \cellcolor{poscolor!0.0}\transparent{0.3}0.00 & \cellcolor{poscolor!0.3}\transparent{0.3}0.00 & \cellcolor{poscolor!0.0}\transparent{0.3}0.00 & \cellcolor{negcolor!0.4}\transparent{0.3}-0.00 & \cellcolor{lightgrey!0.9}0.00 \\
\addlinespace[\rowspacing]
is more offensive & \cellcolor{poscolor!0.0}\transparent{0.3}0.00 & \cellcolor{negcolor!0.4}\transparent{0.3}-0.00 & \cellcolor{poscolor!0.0}\transparent{0.3}0.00 & \cellcolor{negcolor!0.4}\transparent{0.3}-0.00 & \cellcolor{poscolor!0.0}\transparent{0.3}0.00 & \cellcolor{poscolor!0.0}\transparent{0.3}0.00 & \cellcolor{poscolor!0.0}\transparent{0.3}0.00 & \cellcolor{poscolor!0.0}\transparent{0.3}0.00 & \cellcolor{lightgrey!0.4}0.00 \\
\end{tabular}
\end{minipage}
\caption{\textbf{Strength of all 40 traits for Anthropic Claude models under two annotators.} Subcolumns \emph{Gemini} are annotated by Gemini-2.5-Flash and subcolumns \emph{Claude} by Claude-Haiku-4.5. Positive values are shown in {\color{blue}blue} and negative values in {\color{red}red}. Insignificant results greyed out ($p < 0.05 / (40 \times 4)$). Directionally, the annotators agree on 78.1\% of cells (both insignificant, or both significant with the same sign). Most disagreements concern significance rather than sign: only 0.6\% of cells are significant in opposite directions. All measurements are relative to GPT-4o. Sorted by max difference between the annotators.}
\label{fig:app:results:family_bias_claude}
\end{figure}

\begin{figure}[p]
\centering
\begin{minipage}[t]{1\textwidth}
\centering
\sffamily
\tablefontsize
\setlength{\tabcolsep}{2pt}
\begin{tabular}{
    >{\raggedright\arraybackslash}m{0.3\linewidth}
    @{\hspace{12pt}}
    >{\centering\arraybackslash}m{0.075\linewidth}
    @{\hspace{1pt}}
    >{\centering\arraybackslash}m{0.075\linewidth}
    @{\hspace{12pt}}
    >{\centering\arraybackslash}m{0.075\linewidth}
    @{\hspace{1pt}}
    >{\centering\arraybackslash}m{0.075\linewidth}
    @{\hspace{12pt}}
    >{\centering\arraybackslash}m{0.075\linewidth}
    @{\hspace{1pt}}
    >{\centering\arraybackslash}m{0.075\linewidth}
    @{\hspace{12pt}}
    >{\centering\arraybackslash}m{0.075\linewidth}
}

\hfill\makecell[r]{{\color{gray}Model tested $\rightarrow$}\\\raisebox{3pt}{\color{gray}Model annotating \raisebox{-0.6ex}{$\searrow$}}} & \multicolumn{2}{@{}c@{\hspace{12pt}}}{\textbf{\makecell{OpenAI\\\textit{GPT-5-Chat}}}} & \multicolumn{2}{@{}c@{\hspace{12pt}}}{\textbf{\makecell{OpenAI\\\textit{GPT-5.1-Chat}}}} & \multicolumn{2}{@{}c@{\hspace{12pt}}}{\textbf{\makecell{OpenAI\\\textit{GPT-5.3-Chat}}}} & \\
\textbf{Generating a response that...} & \textit{Gemini} & \textit{Claude} & \textit{Gemini} & \textit{Claude} & \textit{Gemini} & \textit{Claude} & \textbf{Max diff} \\
\toprule
\rowcolor{altrow} uses more casual language & \cellcolor{poscolor!14.2}0.12 & \cellcolor{poscolor!10.9}\transparent{0.3}0.09 & \cellcolor{poscolor!3.6}\transparent{0.3}0.03 & \cellcolor{poscolor!74.9}0.63 & \cellcolor{poscolor!39.1}0.33 & \cellcolor{poscolor!87.9}0.74 & \cellcolor{lightgrey!100.0}0.60 \\
\addlinespace[\rowspacing]
makes more confident statements & \cellcolor{poscolor!15.2}0.13 & \cellcolor{negcolor!21.7}-0.18 & \cellcolor{negcolor!14.6}-0.12 & \cellcolor{poscolor!51.7}0.44 & \cellcolor{poscolor!1.7}\transparent{0.3}0.01 & \cellcolor{poscolor!26.1}0.22 & \cellcolor{lightgrey!93.0}0.56 \\
\addlinespace[\rowspacing]
\rowcolor{altrow} provides conclusions without full reasoning & \cellcolor{negcolor!1.2}\transparent{0.3}-0.01 & \cellcolor{negcolor!25.9}-0.22 & \cellcolor{poscolor!5.5}0.05 & \cellcolor{poscolor!69.2}0.58 & \cellcolor{poscolor!1.9}\transparent{0.3}0.02 & \cellcolor{poscolor!30.3}0.26 & \cellcolor{lightgrey!89.4}0.54 \\
\addlinespace[\rowspacing]
is more creative and original & \cellcolor{poscolor!15.4}0.13 & \cellcolor{poscolor!75.6}0.64 & \cellcolor{negcolor!6.1}\transparent{0.3}-0.05 & \cellcolor{negcolor!11.8}\transparent{0.3}-0.10 & \cellcolor{poscolor!8.5}0.07 & \cellcolor{poscolor!69.0}0.58 & \cellcolor{lightgrey!84.7}0.51 \\
\addlinespace[\rowspacing]
\rowcolor{altrow} uses more humour & \cellcolor{poscolor!1.2}\transparent{0.3}0.01 & \cellcolor{poscolor!26.3}0.22 & \cellcolor{negcolor!0.7}\transparent{0.3}-0.01 & \cellcolor{poscolor!7.6}\transparent{0.3}0.06 & \cellcolor{poscolor!3.3}\transparent{0.3}0.03 & \cellcolor{poscolor!55.0}0.46 & \cellcolor{lightgrey!72.4}0.44 \\
\addlinespace[\rowspacing]
uses more formal language & \cellcolor{negcolor!6.8}\transparent{0.3}-0.06 & \cellcolor{negcolor!30.7}-0.26 & \cellcolor{negcolor!46.2}-0.39 & \cellcolor{negcolor!79.0}-0.67 & \cellcolor{negcolor!50.2}-0.43 & \cellcolor{negcolor!94.6}-0.80 & \cellcolor{lightgrey!62.5}0.38 \\
\addlinespace[\rowspacing]
\rowcolor{altrow} more actively engages with the user & \cellcolor{poscolor!63.0}0.53 & \cellcolor{poscolor!72.0}0.61 & \cellcolor{negcolor!1.2}\transparent{0.3}-0.01 & \cellcolor{negcolor!42.2}-0.36 & \cellcolor{poscolor!28.4}0.24 & \cellcolor{poscolor!37.4}0.32 & \cellcolor{lightgrey!57.8}0.35 \\
\addlinespace[\rowspacing]
is more polite & \cellcolor{poscolor!21.8}0.18 & \cellcolor{negcolor!17.2}-0.15 & \cellcolor{negcolor!33.5}-0.28 & \cellcolor{negcolor!60.6}-0.51 & \cellcolor{negcolor!26.9}-0.23 & \cellcolor{negcolor!67.5}-0.57 & \cellcolor{lightgrey!57.1}0.34 \\
\addlinespace[\rowspacing]
\rowcolor{altrow} actively engages the reader with rhetorical questions & \cellcolor{poscolor!12.3}0.10 & \cellcolor{poscolor!52.4}0.44 & \cellcolor{negcolor!2.4}\transparent{0.3}-0.02 & \cellcolor{negcolor!7.3}\transparent{0.3}-0.06 & \cellcolor{poscolor!3.6}\transparent{0.3}0.03 & \cellcolor{poscolor!25.1}0.21 & \cellcolor{lightgrey!56.1}0.34 \\
\addlinespace[\rowspacing]
provides more examples & \cellcolor{poscolor!16.4}0.14 & \cellcolor{poscolor!24.6}0.21 & \cellcolor{negcolor!36.8}-0.31 & \cellcolor{negcolor!75.7}-0.64 & \cellcolor{negcolor!16.3}-0.14 & \cellcolor{negcolor!43.4}-0.37 & \cellcolor{lightgrey!54.8}0.33 \\
\addlinespace[\rowspacing]
\rowcolor{altrow} has a more avoidant tone & \cellcolor{negcolor!1.7}\transparent{0.3}-0.01 & \cellcolor{negcolor!17.2}-0.15 & \cellcolor{poscolor!1.2}\transparent{0.3}0.01 & \cellcolor{poscolor!37.7}0.32 & \cellcolor{negcolor!1.9}\transparent{0.3}-0.02 & \cellcolor{poscolor!0.9}\transparent{0.3}0.01 & \cellcolor{lightgrey!51.2}0.31 \\
\addlinespace[\rowspacing]
has a friendlier tone & \cellcolor{poscolor!31.0}0.26 & \cellcolor{poscolor!50.7}0.43 & \cellcolor{negcolor!21.5}-0.18 & \cellcolor{negcolor!20.5}-0.17 & \cellcolor{poscolor!16.6}0.14 & \cellcolor{poscolor!52.1}0.44 & \cellcolor{lightgrey!49.8}0.30 \\
\addlinespace[\rowspacing]
\rowcolor{altrow} expresses more emotion & \cellcolor{poscolor!2.6}\transparent{0.3}0.02 & \cellcolor{poscolor!27.7}0.23 & \cellcolor{negcolor!5.2}-0.04 & \cellcolor{negcolor!16.5}-0.14 & \cellcolor{poscolor!1.2}\transparent{0.3}0.01 & \cellcolor{poscolor!35.1}0.30 & \cellcolor{lightgrey!47.5}0.29 \\
\addlinespace[\rowspacing]
uses a more enthusiastic tone & \cellcolor{poscolor!13.5}0.11 & \cellcolor{poscolor!45.5}0.38 & \cellcolor{negcolor!16.3}-0.14 & \cellcolor{negcolor!24.1}-0.20 & \cellcolor{negcolor!5.4}\transparent{0.3}-0.05 & \cellcolor{poscolor!20.6}0.17 & \cellcolor{lightgrey!44.9}0.27 \\
\addlinespace[\rowspacing]
\rowcolor{altrow} uses more personal pronouns (I, we, you) & \cellcolor{poscolor!23.7}0.20 & \cellcolor{poscolor!54.3}0.46 & \cellcolor{negcolor!18.9}-0.16 & \cellcolor{negcolor!40.3}-0.34 & \cellcolor{poscolor!17.3}0.15 & \cellcolor{poscolor!43.8}0.37 & \cellcolor{lightgrey!42.9}0.26 \\
\addlinespace[\rowspacing]
is more optimistic & \cellcolor{negcolor!0.2}\transparent{0.3}-0.00 & \cellcolor{poscolor!13.0}0.11 & \cellcolor{negcolor!4.5}-0.04 & \cellcolor{negcolor!27.8}-0.24 & \cellcolor{negcolor!2.6}\transparent{0.3}-0.02 & \cellcolor{poscolor!3.6}\transparent{0.3}0.03 & \cellcolor{lightgrey!32.9}0.20 \\
\addlinespace[\rowspacing]
\rowcolor{altrow} acknowledges own limitations or uncertainty more & \cellcolor{negcolor!0.9}\transparent{0.3}-0.01 & \cellcolor{poscolor!12.1}0.10 & \cellcolor{negcolor!2.1}\transparent{0.3}-0.02 & \cellcolor{poscolor!3.8}\transparent{0.3}0.03 & \cellcolor{poscolor!2.1}\transparent{0.3}0.02 & \cellcolor{poscolor!23.9}0.20 & \cellcolor{lightgrey!30.6}0.18 \\
\addlinespace[\rowspacing]
is more empathetic to the user & \cellcolor{poscolor!17.8}0.15 & \cellcolor{poscolor!24.2}0.20 & \cellcolor{negcolor!11.1}-0.09 & \cellcolor{negcolor!16.7}-0.14 & \cellcolor{poscolor!10.9}0.09 & \cellcolor{poscolor!32.5}0.27 & \cellcolor{lightgrey!30.2}0.18 \\
\addlinespace[\rowspacing]
\rowcolor{altrow} is more concise & \cellcolor{negcolor!19.8}-0.17 & \cellcolor{negcolor!37.5}-0.32 & \cellcolor{poscolor!93.8}0.79 & \cellcolor{poscolor!100.0}0.84 & \cellcolor{poscolor!57.6}0.49 & \cellcolor{poscolor!78.9}0.67 & \cellcolor{lightgrey!29.9}0.18 \\
\addlinespace[\rowspacing]
more strictly follows the requested output format & \cellcolor{poscolor!1.4}\transparent{0.3}0.01 & \cellcolor{poscolor!12.1}\transparent{0.3}0.10 & \cellcolor{negcolor!4.7}\transparent{0.3}-0.04 & \cellcolor{negcolor!2.1}\transparent{0.3}-0.02 & \cellcolor{negcolor!11.8}-0.10 & \cellcolor{negcolor!32.3}-0.27 & \cellcolor{lightgrey!28.9}0.17 \\
\addlinespace[\rowspacing]
\rowcolor{altrow} uses more bold and italics text & \cellcolor{poscolor!82.0}0.69 & \cellcolor{poscolor!73.9}0.62 & \cellcolor{negcolor!58.7}-0.50 & \cellcolor{negcolor!75.0}-0.64 & \cellcolor{negcolor!50.7}-0.43 & \cellcolor{negcolor!63.9}-0.54 & \cellcolor{lightgrey!22.9}0.14 \\
\addlinespace[\rowspacing]
is more factually correct & \cellcolor{poscolor!12.1}0.10 & \cellcolor{poscolor!20.6}0.17 & \cellcolor{negcolor!6.1}\transparent{0.3}-0.05 & \cellcolor{negcolor!3.8}\transparent{0.3}-0.03 & \cellcolor{poscolor!5.0}\transparent{0.3}0.04 & \cellcolor{poscolor!19.0}0.16 & \cellcolor{lightgrey!19.6}0.12 \\
\addlinespace[\rowspacing]
\rowcolor{altrow} is more verbose & \cellcolor{poscolor!27.5}0.23 & \cellcolor{poscolor!25.8}0.22 & \cellcolor{negcolor!100.0}-0.85 & \cellcolor{negcolor!99.8}-0.85 & \cellcolor{negcolor!68.6}-0.58 & \cellcolor{negcolor!80.4}-0.68 & \cellcolor{lightgrey!16.6}0.10 \\
\addlinespace[\rowspacing]
ends with a follow-up question & \cellcolor{poscolor!55.0}0.46 & \cellcolor{poscolor!51.7}0.44 & \cellcolor{poscolor!7.8}\transparent{0.3}0.07 & \cellcolor{negcolor!3.3}\transparent{0.3}-0.03 & \cellcolor{poscolor!19.4}0.16 & \cellcolor{poscolor!25.1}0.21 & \cellcolor{lightgrey!15.6}0.09 \\
\addlinespace[\rowspacing]
\rowcolor{altrow} provides a numbered list format & \cellcolor{negcolor!10.8}\transparent{0.3}-0.09 & \cellcolor{negcolor!5.2}\transparent{0.3}-0.04 & \cellcolor{negcolor!48.6}-0.41 & \cellcolor{negcolor!58.5}-0.50 & \cellcolor{negcolor!47.2}-0.40 & \cellcolor{negcolor!55.9}-0.47 & \cellcolor{lightgrey!14.0}0.08 \\
\addlinespace[\rowspacing]
has more structured formatting & \cellcolor{poscolor!50.0}0.42 & \cellcolor{poscolor!57.1}0.48 & \cellcolor{negcolor!74.1}-0.63 & \cellcolor{negcolor!75.2}-0.64 & \cellcolor{negcolor!55.4}-0.47 & \cellcolor{negcolor!63.7}-0.54 & \cellcolor{lightgrey!11.6}0.07 \\
\addlinespace[\rowspacing]
\rowcolor{altrow} includes more ethical considerations & \cellcolor{negcolor!2.1}\transparent{0.3}-0.02 & \cellcolor{poscolor!4.3}\transparent{0.3}0.04 & \cellcolor{negcolor!7.5}-0.06 & \cellcolor{negcolor!14.6}-0.12 & \cellcolor{negcolor!4.5}\transparent{0.3}-0.04 & \cellcolor{negcolor!7.3}\transparent{0.3}-0.06 & \cellcolor{lightgrey!10.0}0.06 \\
\addlinespace[\rowspacing]
uses more emojis & \cellcolor{poscolor!9.7}0.08 & \cellcolor{poscolor!16.6}0.14 & \cellcolor{negcolor!4.0}-0.03 & \cellcolor{negcolor!4.2}-0.04 & \cellcolor{poscolor!5.9}0.05 & \cellcolor{poscolor!8.3}0.07 & \cellcolor{lightgrey!9.6}0.06 \\
\addlinespace[\rowspacing]
\rowcolor{altrow} compliments the user's question or prompt & \cellcolor{poscolor!23.9}0.20 & \cellcolor{poscolor!30.8}0.26 & \cellcolor{negcolor!11.8}-0.10 & \cellcolor{negcolor!15.3}-0.13 & \cellcolor{negcolor!0.2}\transparent{0.3}-0.00 & \cellcolor{negcolor!2.8}\transparent{0.3}-0.02 & \cellcolor{lightgrey!9.6}0.06 \\
\addlinespace[\rowspacing]
contains less harmful information & \cellcolor{poscolor!0.2}\transparent{0.3}0.00 & \cellcolor{poscolor!2.1}\transparent{0.3}0.02 & \cellcolor{poscolor!1.2}\transparent{0.3}0.01 & \cellcolor{poscolor!8.1}0.07 & \cellcolor{poscolor!0.7}\transparent{0.3}0.01 & \cellcolor{poscolor!5.5}\transparent{0.3}0.05 & \cellcolor{lightgrey!9.6}0.06 \\
\addlinespace[\rowspacing]
\rowcolor{altrow} agrees more with the user & \cellcolor{poscolor!0.0}\transparent{0.3}0.00 & \cellcolor{poscolor!3.1}\transparent{0.3}0.03 & \cellcolor{negcolor!3.8}\transparent{0.3}-0.03 & \cellcolor{poscolor!1.7}\transparent{0.3}0.01 & \cellcolor{negcolor!1.4}\transparent{0.3}-0.01 & \cellcolor{poscolor!4.7}\transparent{0.3}0.04 & \cellcolor{lightgrey!8.6}0.05 \\
\addlinespace[\rowspacing]
includes more references to other sources & \cellcolor{poscolor!6.4}0.05 & \cellcolor{poscolor!5.7}\transparent{0.3}0.05 & \cellcolor{negcolor!4.2}\transparent{0.3}-0.04 & \cellcolor{negcolor!8.5}-0.07 & \cellcolor{negcolor!2.6}\transparent{0.3}-0.02 & \cellcolor{negcolor!5.4}-0.05 & \cellcolor{lightgrey!6.0}0.04 \\
\addlinespace[\rowspacing]
\rowcolor{altrow} reinforces user's beliefs more & \cellcolor{negcolor!0.5}\transparent{0.3}-0.00 & \cellcolor{poscolor!0.7}\transparent{0.3}0.01 & \cellcolor{negcolor!2.4}\transparent{0.3}-0.02 & \cellcolor{poscolor!1.9}\transparent{0.3}0.02 & \cellcolor{negcolor!1.9}\transparent{0.3}-0.02 & \cellcolor{poscolor!1.4}\transparent{0.3}0.01 & \cellcolor{lightgrey!6.0}0.04 \\
\addlinespace[\rowspacing]
uses more mathematical symbols and notation & \cellcolor{poscolor!5.2}0.04 & \cellcolor{poscolor!8.1}0.07 & \cellcolor{negcolor!10.4}-0.09 & \cellcolor{negcolor!10.6}-0.09 & \cellcolor{negcolor!9.0}-0.08 & \cellcolor{negcolor!9.2}-0.08 & \cellcolor{lightgrey!4.0}0.02 \\
\addlinespace[\rowspacing]
\rowcolor{altrow} refuses to answer the question & \cellcolor{negcolor!0.2}\transparent{0.3}-0.00 & \cellcolor{negcolor!0.5}\transparent{0.3}-0.00 & \cellcolor{poscolor!2.1}\transparent{0.3}0.02 & \cellcolor{poscolor!1.2}\transparent{0.3}0.01 & \cellcolor{negcolor!0.7}\transparent{0.3}-0.01 & \cellcolor{negcolor!0.7}\transparent{0.3}-0.01 & \cellcolor{lightgrey!1.3}0.01 \\
\addlinespace[\rowspacing]
agrees with user even if factually incorrect & \cellcolor{negcolor!0.2}\transparent{0.3}-0.00 & \cellcolor{negcolor!0.9}\transparent{0.3}-0.01 & \cellcolor{poscolor!0.2}\transparent{0.3}0.00 & \cellcolor{poscolor!0.7}\transparent{0.3}0.01 & \cellcolor{negcolor!1.2}\transparent{0.3}-0.01 & \cellcolor{negcolor!1.2}\transparent{0.3}-0.01 & \cellcolor{lightgrey!1.0}0.01 \\
\addlinespace[\rowspacing]
\rowcolor{altrow} suggests illegal activities & \cellcolor{negcolor!0.2}\transparent{0.3}-0.00 & \cellcolor{negcolor!0.7}\transparent{0.3}-0.01 & \cellcolor{negcolor!0.2}\transparent{0.3}-0.00 & \cellcolor{negcolor!0.9}\transparent{0.3}-0.01 & \cellcolor{poscolor!0.0}\transparent{0.3}0.00 & \cellcolor{negcolor!0.2}\transparent{0.3}-0.00 & \cellcolor{lightgrey!1.0}0.01 \\
\addlinespace[\rowspacing]
reinforces user's anger more & \cellcolor{poscolor!0.2}\transparent{0.3}0.00 & \cellcolor{poscolor!0.0}\transparent{0.3}0.00 & \cellcolor{poscolor!0.0}\transparent{0.3}0.00 & \cellcolor{poscolor!0.5}\transparent{0.3}0.00 & \cellcolor{poscolor!0.0}\transparent{0.3}0.00 & \cellcolor{poscolor!0.2}\transparent{0.3}0.00 & \cellcolor{lightgrey!0.7}0.00 \\
\addlinespace[\rowspacing]
\rowcolor{altrow} includes inappropriate language & \cellcolor{poscolor!0.0}\transparent{0.3}0.00 & \cellcolor{negcolor!0.5}\transparent{0.3}-0.00 & \cellcolor{poscolor!0.0}\transparent{0.3}0.00 & \cellcolor{negcolor!0.5}\transparent{0.3}-0.00 & \cellcolor{poscolor!0.0}\transparent{0.3}0.00 & \cellcolor{negcolor!0.2}\transparent{0.3}-0.00 & \cellcolor{lightgrey!0.7}0.00 \\
\addlinespace[\rowspacing]
is more offensive & \cellcolor{poscolor!0.0}\transparent{0.3}0.00 & \cellcolor{negcolor!0.5}\transparent{0.3}-0.00 & \cellcolor{poscolor!0.0}\transparent{0.3}0.00 & \cellcolor{negcolor!0.5}\transparent{0.3}-0.00 & \cellcolor{poscolor!0.0}\transparent{0.3}0.00 & \cellcolor{negcolor!0.5}\transparent{0.3}-0.00 & \cellcolor{lightgrey!0.7}0.00 \\
\end{tabular}
\end{minipage}
\caption{\textbf{Strength of all 40 traits for OpenAI GPT models under two annotators.} Subcolumns \emph{Gemini} are annotated by Gemini-2.5-Flash and subcolumns \emph{Claude} by Claude-Haiku-4.5. Positive values are shown in {\color{blue}blue} and negative values in {\color{red}red}. Insignificant results greyed out ($p < 0.05 / (40 \times 3)$). Directionally, the annotators agree on 78.3\% of cells (both insignificant, or both significant with the same sign). Most disagreements concern significance rather than sign: only 2.5\% of cells are significant in opposite directions. All measurements are relative to GPT-4o. Sorted by max difference between the annotators.}
\label{fig:app:results:family_bias_gpt}
\end{figure}

\begin{figure}[p]
\centering
\begin{minipage}[t]{1\textwidth}
\centering
\sffamily
\tablefontsize
\setlength{\tabcolsep}{2pt}
\begin{tabular}{
    >{\raggedright\arraybackslash}m{0.3\linewidth}
    @{\hspace{12pt}}
    >{\centering\arraybackslash}m{0.075\linewidth}
    @{\hspace{1pt}}
    >{\centering\arraybackslash}m{0.075\linewidth}
    @{\hspace{12pt}}
    >{\centering\arraybackslash}m{0.075\linewidth}
    @{\hspace{1pt}}
    >{\centering\arraybackslash}m{0.075\linewidth}
    @{\hspace{12pt}}
    >{\centering\arraybackslash}m{0.075\linewidth}
    @{\hspace{1pt}}
    >{\centering\arraybackslash}m{0.075\linewidth}
    @{\hspace{12pt}}
    >{\centering\arraybackslash}m{0.075\linewidth}
}

\hfill\makecell[r]{{\color{gray}Model tested $\rightarrow$}\\\raisebox{3pt}{\color{gray}Model annotating \raisebox{-0.6ex}{$\searrow$}}} & \multicolumn{2}{@{}c@{\hspace{12pt}}}{\textbf{\makecell{Google\\\textit{Gemini-2.5-Pro}}}} & \multicolumn{2}{@{}c@{\hspace{12pt}}}{\textbf{\makecell{Google\\\textit{Gemini-3-Pro}}}} & \multicolumn{2}{@{}c@{\hspace{12pt}}}{\textbf{\makecell{Google\\\textit{Gemini-3.1-Pro}}}} & \\
\textbf{Generating a response that...} & \textit{Gemini} & \textit{Claude} & \textit{Gemini} & \textit{Claude} & \textit{Gemini} & \textit{Claude} & \textbf{Max diff} \\
\toprule
\rowcolor{altrow} is more polite & \cellcolor{poscolor!62.5}0.47 & \cellcolor{negcolor!29.4}-0.22 & \cellcolor{negcolor!23.2}-0.18 & \cellcolor{negcolor!55.0}-0.42 & \cellcolor{negcolor!9.4}\transparent{0.3}-0.07 & \cellcolor{negcolor!44.5}-0.34 & \cellcolor{lightgrey!100.0}0.69 \\
\addlinespace[\rowspacing]
makes more confident statements & \cellcolor{poscolor!75.3}0.57 & \cellcolor{poscolor!9.6}\transparent{0.3}0.07 & \cellcolor{poscolor!54.2}0.41 & \cellcolor{poscolor!52.0}0.39 & \cellcolor{poscolor!63.4}0.48 & \cellcolor{poscolor!35.5}0.27 & \cellcolor{lightgrey!71.4}0.49 \\
\addlinespace[\rowspacing]
\rowcolor{altrow} provides conclusions without full reasoning & \cellcolor{negcolor!0.5}\transparent{0.3}-0.00 & \cellcolor{negcolor!60.8}-0.46 & \cellcolor{negcolor!0.5}\transparent{0.3}-0.00 & \cellcolor{negcolor!33.9}-0.26 & \cellcolor{poscolor!0.0}\transparent{0.3}0.00 & \cellcolor{negcolor!43.9}-0.33 & \cellcolor{lightgrey!65.9}0.46 \\
\addlinespace[\rowspacing]
is more creative and original & \cellcolor{poscolor!44.7}0.34 & \cellcolor{poscolor!97.4}0.73 & \cellcolor{poscolor!35.2}0.26 & \cellcolor{poscolor!88.6}0.67 & \cellcolor{poscolor!40.7}0.31 & \cellcolor{poscolor!100.0}0.75 & \cellcolor{lightgrey!64.5}0.45 \\
\addlinespace[\rowspacing]
\rowcolor{altrow} uses more formal language & \cellcolor{poscolor!22.4}0.17 & \cellcolor{negcolor!33.9}-0.26 & \cellcolor{poscolor!3.5}\transparent{0.3}0.03 & \cellcolor{negcolor!39.3}-0.30 & \cellcolor{poscolor!0.8}\transparent{0.3}0.01 & \cellcolor{negcolor!48.2}-0.36 & \cellcolor{lightgrey!61.3}0.42 \\
\addlinespace[\rowspacing]
more actively engages with the user & \cellcolor{poscolor!33.8}0.25 & \cellcolor{poscolor!73.5}0.55 & \cellcolor{negcolor!10.5}-0.08 & \cellcolor{poscolor!30.6}0.23 & \cellcolor{poscolor!4.3}\transparent{0.3}0.03 & \cellcolor{poscolor!51.5}0.39 & \cellcolor{lightgrey!51.2}0.35 \\
\addlinespace[\rowspacing]
\rowcolor{altrow} expresses more emotion & \cellcolor{poscolor!6.1}0.05 & \cellcolor{poscolor!52.2}0.39 & \cellcolor{poscolor!0.8}\transparent{0.3}0.01 & \cellcolor{poscolor!24.9}0.19 & \cellcolor{poscolor!6.8}0.05 & \cellcolor{poscolor!39.6}0.30 & \cellcolor{lightgrey!50.0}0.35 \\
\addlinespace[\rowspacing]
is more concise & \cellcolor{negcolor!60.1}-0.45 & \cellcolor{negcolor!100.0}-0.76 & \cellcolor{negcolor!24.5}-0.19 & \cellcolor{negcolor!67.3}-0.51 & \cellcolor{negcolor!35.8}-0.27 & \cellcolor{negcolor!80.0}-0.60 & \cellcolor{lightgrey!48.3}0.33 \\
\addlinespace[\rowspacing]
\rowcolor{altrow} has a more avoidant tone & \cellcolor{negcolor!3.7}\transparent{0.3}-0.03 & \cellcolor{negcolor!47.1}-0.36 & \cellcolor{negcolor!3.5}\transparent{0.3}-0.03 & \cellcolor{negcolor!26.9}-0.20 & \cellcolor{negcolor!4.0}\transparent{0.3}-0.03 & \cellcolor{negcolor!33.4}-0.25 & \cellcolor{lightgrey!47.4}0.33 \\
\addlinespace[\rowspacing]
uses more humour & \cellcolor{poscolor!7.7}0.06 & \cellcolor{poscolor!38.6}0.29 & \cellcolor{poscolor!7.9}0.06 & \cellcolor{poscolor!37.9}0.29 & \cellcolor{poscolor!7.9}0.06 & \cellcolor{poscolor!45.3}0.34 & \cellcolor{lightgrey!40.6}0.28 \\
\addlinespace[\rowspacing]
\rowcolor{altrow} uses more personal pronouns (I, we, you) & \cellcolor{poscolor!51.4}0.39 & \cellcolor{poscolor!71.3}0.54 & \cellcolor{poscolor!9.5}\transparent{0.3}0.07 & \cellcolor{poscolor!24.1}0.18 & \cellcolor{poscolor!20.1}0.15 & \cellcolor{poscolor!51.8}0.39 & \cellcolor{lightgrey!34.4}0.24 \\
\addlinespace[\rowspacing]
provides more examples & \cellcolor{poscolor!71.9}0.54 & \cellcolor{poscolor!91.0}0.68 & \cellcolor{poscolor!40.4}0.30 & \cellcolor{poscolor!66.4}0.50 & \cellcolor{poscolor!47.2}0.35 & \cellcolor{poscolor!76.2}0.57 & \cellcolor{lightgrey!31.5}0.22 \\
\addlinespace[\rowspacing]
\rowcolor{altrow} provides a numbered list format & \cellcolor{poscolor!6.9}\transparent{0.3}0.05 & \cellcolor{poscolor!34.3}0.26 & \cellcolor{poscolor!16.0}0.12 & \cellcolor{poscolor!30.4}0.23 & \cellcolor{poscolor!13.3}\transparent{0.3}0.10 & \cellcolor{poscolor!26.3}0.20 & \cellcolor{lightgrey!29.8}0.21 \\
\addlinespace[\rowspacing]
has a friendlier tone & \cellcolor{poscolor!63.3}0.48 & \cellcolor{poscolor!46.8}0.35 & \cellcolor{negcolor!12.9}-0.10 & \cellcolor{poscolor!6.2}\transparent{0.3}0.05 & \cellcolor{poscolor!5.1}\transparent{0.3}0.04 & \cellcolor{poscolor!31.4}0.24 & \cellcolor{lightgrey!28.5}0.20 \\
\addlinespace[\rowspacing]
\rowcolor{altrow} more strictly follows the requested output format & \cellcolor{poscolor!8.8}\transparent{0.3}0.07 & \cellcolor{poscolor!15.7}\transparent{0.3}0.12 & \cellcolor{poscolor!6.2}\transparent{0.3}0.05 & \cellcolor{poscolor!31.2}0.23 & \cellcolor{poscolor!9.8}0.07 & \cellcolor{poscolor!18.7}0.14 & \cellcolor{lightgrey!27.1}0.19 \\
\addlinespace[\rowspacing]
actively engages the reader with rhetorical questions & \cellcolor{poscolor!20.8}0.16 & \cellcolor{poscolor!42.0}0.32 & \cellcolor{poscolor!10.0}0.08 & \cellcolor{poscolor!23.0}0.17 & \cellcolor{poscolor!11.7}0.09 & \cellcolor{poscolor!25.5}0.19 & \cellcolor{lightgrey!23.1}0.16 \\
\addlinespace[\rowspacing]
\rowcolor{altrow} uses more casual language & \cellcolor{poscolor!7.7}0.06 & \cellcolor{poscolor!0.0}\transparent{0.3}0.00 & \cellcolor{poscolor!3.8}\transparent{0.3}0.03 & \cellcolor{poscolor!19.8}0.15 & \cellcolor{poscolor!9.8}0.07 & \cellcolor{poscolor!30.4}0.23 & \cellcolor{lightgrey!22.4}0.15 \\
\addlinespace[\rowspacing]
is more factually correct & \cellcolor{poscolor!27.4}0.21 & \cellcolor{poscolor!23.7}0.18 & \cellcolor{poscolor!24.7}0.19 & \cellcolor{poscolor!45.0}0.34 & \cellcolor{poscolor!31.4}0.24 & \cellcolor{poscolor!46.3}0.35 & \cellcolor{lightgrey!22.1}0.15 \\
\addlinespace[\rowspacing]
\rowcolor{altrow} uses a more enthusiastic tone & \cellcolor{poscolor!47.9}0.36 & \cellcolor{poscolor!59.9}0.45 & \cellcolor{poscolor!1.6}\transparent{0.3}0.01 & \cellcolor{poscolor!17.1}0.13 & \cellcolor{poscolor!16.3}0.12 & \cellcolor{poscolor!36.0}0.27 & \cellcolor{lightgrey!21.5}0.15 \\
\addlinespace[\rowspacing]
is more empathetic to the user & \cellcolor{poscolor!39.9}0.30 & \cellcolor{poscolor!27.1}0.20 & \cellcolor{poscolor!3.8}\transparent{0.3}0.03 & \cellcolor{negcolor!13.7}-0.10 & \cellcolor{poscolor!19.2}0.14 & \cellcolor{poscolor!10.3}\transparent{0.3}0.08 & \cellcolor{lightgrey!19.1}0.13 \\
\addlinespace[\rowspacing]
\rowcolor{altrow} acknowledges own limitations or uncertainty more & \cellcolor{negcolor!8.5}-0.06 & \cellcolor{negcolor!24.3}-0.18 & \cellcolor{negcolor!6.7}\transparent{0.3}-0.05 & \cellcolor{negcolor!19.9}-0.15 & \cellcolor{negcolor!5.4}\transparent{0.3}-0.04 & \cellcolor{negcolor!22.9}-0.17 & \cellcolor{lightgrey!19.1}0.13 \\
\addlinespace[\rowspacing]
has more structured formatting & \cellcolor{poscolor!92.9}0.70 & \cellcolor{poscolor!90.5}0.68 & \cellcolor{poscolor!71.0}0.53 & \cellcolor{poscolor!83.5}0.63 & \cellcolor{poscolor!66.4}0.50 & \cellcolor{poscolor!82.9}0.62 & \cellcolor{lightgrey!18.0}0.12 \\
\addlinespace[\rowspacing]
\rowcolor{altrow} is more optimistic & \cellcolor{poscolor!8.2}0.06 & \cellcolor{poscolor!23.4}0.18 & \cellcolor{negcolor!0.8}\transparent{0.3}-0.01 & \cellcolor{negcolor!8.1}\transparent{0.3}-0.06 & \cellcolor{poscolor!3.8}\transparent{0.3}0.03 & \cellcolor{poscolor!4.3}\transparent{0.3}0.03 & \cellcolor{lightgrey!16.5}0.11 \\
\addlinespace[\rowspacing]
compliments the user's question or prompt & \cellcolor{poscolor!72.7}0.55 & \cellcolor{poscolor!65.5}0.49 & \cellcolor{negcolor!3.5}\transparent{0.3}-0.03 & \cellcolor{poscolor!7.3}\transparent{0.3}0.05 & \cellcolor{poscolor!3.8}\transparent{0.3}0.03 & \cellcolor{poscolor!18.4}0.14 & \cellcolor{lightgrey!15.9}0.11 \\
\addlinespace[\rowspacing]
\rowcolor{altrow} contains less harmful information & \cellcolor{poscolor!1.3}\transparent{0.3}0.01 & \cellcolor{negcolor!9.5}-0.07 & \cellcolor{negcolor!0.3}\transparent{0.3}-0.00 & \cellcolor{negcolor!12.9}-0.10 & \cellcolor{poscolor!0.3}\transparent{0.3}0.00 & \cellcolor{negcolor!8.4}\transparent{0.3}-0.06 & \cellcolor{lightgrey!13.8}0.10 \\
\addlinespace[\rowspacing]
includes more ethical considerations & \cellcolor{poscolor!14.1}0.11 & \cellcolor{poscolor!15.7}0.12 & \cellcolor{poscolor!8.1}0.06 & \cellcolor{negcolor!0.5}\transparent{0.3}-0.00 & \cellcolor{poscolor!11.9}0.09 & \cellcolor{poscolor!13.3}0.10 & \cellcolor{lightgrey!9.4}0.07 \\
\addlinespace[\rowspacing]
\rowcolor{altrow} reinforces user's beliefs more & \cellcolor{poscolor!6.1}0.05 & \cellcolor{poscolor!9.0}0.07 & \cellcolor{poscolor!0.5}\transparent{0.3}0.00 & \cellcolor{poscolor!7.0}0.05 & \cellcolor{poscolor!1.9}\transparent{0.3}0.01 & \cellcolor{poscolor!6.8}0.05 & \cellcolor{lightgrey!7.1}0.05 \\
\addlinespace[\rowspacing]
is more verbose & \cellcolor{poscolor!99.0}0.74 & \cellcolor{poscolor!99.3}0.75 & \cellcolor{poscolor!51.2}0.38 & \cellcolor{poscolor!57.7}0.43 & \cellcolor{poscolor!78.0}0.59 & \cellcolor{poscolor!72.4}0.54 & \cellcolor{lightgrey!7.1}0.05 \\
\addlinespace[\rowspacing]
\rowcolor{altrow} ends with a follow-up question & \cellcolor{negcolor!18.0}-0.14 & \cellcolor{negcolor!12.7}-0.10 & \cellcolor{negcolor!19.9}-0.15 & \cellcolor{negcolor!13.7}-0.10 & \cellcolor{negcolor!12.7}-0.10 & \cellcolor{negcolor!10.5}-0.08 & \cellcolor{lightgrey!6.8}0.05 \\
\addlinespace[\rowspacing]
agrees more with the user & \cellcolor{poscolor!10.9}0.08 & \cellcolor{poscolor!12.0}0.09 & \cellcolor{poscolor!2.7}\transparent{0.3}0.02 & \cellcolor{poscolor!8.1}0.06 & \cellcolor{poscolor!3.5}\transparent{0.3}0.03 & \cellcolor{poscolor!8.1}0.06 & \cellcolor{lightgrey!5.9}0.04 \\
\addlinespace[\rowspacing]
\rowcolor{altrow} uses more bold and italics text & \cellcolor{poscolor!94.7}0.71 & \cellcolor{poscolor!94.7}0.71 & \cellcolor{poscolor!86.7}0.65 & \cellcolor{poscolor!90.5}0.68 & \cellcolor{poscolor!90.8}0.68 & \cellcolor{poscolor!86.2}0.65 & \cellcolor{lightgrey!5.0}0.03 \\
\addlinespace[\rowspacing]
includes more references to other sources & \cellcolor{poscolor!9.6}0.07 & \cellcolor{poscolor!7.2}0.05 & \cellcolor{poscolor!11.7}0.09 & \cellcolor{poscolor!7.6}\transparent{0.3}0.06 & \cellcolor{poscolor!12.2}0.09 & \cellcolor{poscolor!11.7}0.09 & \cellcolor{lightgrey!4.4}0.03 \\
\addlinespace[\rowspacing]
\rowcolor{altrow} uses more mathematical symbols and notation & \cellcolor{negcolor!4.0}\transparent{0.3}-0.03 & \cellcolor{negcolor!4.5}\transparent{0.3}-0.03 & \cellcolor{poscolor!2.4}\transparent{0.3}0.02 & \cellcolor{poscolor!6.2}\transparent{0.3}0.05 & \cellcolor{negcolor!1.9}\transparent{0.3}-0.01 & \cellcolor{negcolor!1.1}\transparent{0.3}-0.01 & \cellcolor{lightgrey!4.1}0.03 \\
\addlinespace[\rowspacing]
uses more emojis & \cellcolor{negcolor!4.0}-0.03 & \cellcolor{negcolor!3.2}\transparent{0.3}-0.02 & \cellcolor{negcolor!3.2}\transparent{0.3}-0.02 & \cellcolor{negcolor!2.4}\transparent{0.3}-0.02 & \cellcolor{negcolor!0.5}\transparent{0.3}-0.00 & \cellcolor{poscolor!1.6}\transparent{0.3}0.01 & \cellcolor{lightgrey!2.4}0.02 \\
\addlinespace[\rowspacing]
\rowcolor{altrow} agrees with user even if factually incorrect & \cellcolor{poscolor!1.6}\transparent{0.3}0.01 & \cellcolor{poscolor!2.9}\transparent{0.3}0.02 & \cellcolor{poscolor!0.8}\transparent{0.3}0.01 & \cellcolor{poscolor!3.0}\transparent{0.3}0.02 & \cellcolor{poscolor!1.4}\transparent{0.3}0.01 & \cellcolor{poscolor!2.2}\transparent{0.3}0.02 & \cellcolor{lightgrey!2.4}0.02 \\
\addlinespace[\rowspacing]
suggests illegal activities & \cellcolor{poscolor!0.3}\transparent{0.3}0.00 & \cellcolor{poscolor!1.1}\transparent{0.3}0.01 & \cellcolor{poscolor!0.0}\transparent{0.3}0.00 & \cellcolor{poscolor!1.1}\transparent{0.3}0.01 & \cellcolor{poscolor!0.3}\transparent{0.3}0.00 & \cellcolor{poscolor!1.9}\transparent{0.3}0.01 & \cellcolor{lightgrey!1.8}0.01 \\
\addlinespace[\rowspacing]
\rowcolor{altrow} is more offensive & \cellcolor{poscolor!0.0}\transparent{0.3}0.00 & \cellcolor{poscolor!1.1}\transparent{0.3}0.01 & \cellcolor{poscolor!0.0}\transparent{0.3}0.00 & \cellcolor{poscolor!0.0}\transparent{0.3}0.00 & \cellcolor{poscolor!0.0}\transparent{0.3}0.00 & \cellcolor{poscolor!0.8}\transparent{0.3}0.01 & \cellcolor{lightgrey!1.2}0.01 \\
\addlinespace[\rowspacing]
refuses to answer the question & \cellcolor{negcolor!1.3}\transparent{0.3}-0.01 & \cellcolor{negcolor!2.4}\transparent{0.3}-0.02 & \cellcolor{negcolor!1.6}\transparent{0.3}-0.01 & \cellcolor{negcolor!1.6}\transparent{0.3}-0.01 & \cellcolor{negcolor!1.6}\transparent{0.3}-0.01 & \cellcolor{negcolor!2.2}\transparent{0.3}-0.02 & \cellcolor{lightgrey!1.2}0.01 \\
\addlinespace[\rowspacing]
\rowcolor{altrow} reinforces user's anger more & \cellcolor{poscolor!0.5}\transparent{0.3}0.00 & \cellcolor{poscolor!0.5}\transparent{0.3}0.00 & \cellcolor{poscolor!0.0}\transparent{0.3}0.00 & \cellcolor{poscolor!0.5}\transparent{0.3}0.00 & \cellcolor{poscolor!0.8}\transparent{0.3}0.01 & \cellcolor{poscolor!0.5}\transparent{0.3}0.00 & \cellcolor{lightgrey!0.6}0.00 \\
\addlinespace[\rowspacing]
includes inappropriate language & \cellcolor{poscolor!0.0}\transparent{0.3}0.00 & \cellcolor{poscolor!0.5}\transparent{0.3}0.00 & \cellcolor{poscolor!0.0}\transparent{0.3}0.00 & \cellcolor{poscolor!0.3}\transparent{0.3}0.00 & \cellcolor{poscolor!0.3}\transparent{0.3}0.00 & \cellcolor{poscolor!0.5}\transparent{0.3}0.00 & \cellcolor{lightgrey!0.6}0.00 \\
\end{tabular}
\end{minipage}
\caption{\textbf{Strength of all 40 traits for Google Gemini models under two annotators.} Subcolumns \emph{Gemini} are annotated by Gemini-2.5-Flash and subcolumns \emph{Claude} by Claude-Haiku-4.5. Positive values are shown in {\color{blue}blue} and negative values in {\color{red}red}. Insignificant results greyed out ($p < 0.05 / (40 \times 3)$). Directionally, the annotators agree on 68.3\% of cells (both insignificant, or both significant with the same sign). Most disagreements concern significance rather than sign: only 2.5\% of cells are significant in opposite directions. All measurements are relative to GPT-4o. Sorted by max difference between the annotators.}
\label{fig:app:results:family_bias_gemini}
\end{figure}

\clearpage

\subsection{Alternative agreement metrics}
\label{app:alternative_metrics}

We recomputed all figures of \Cref{sec:results:feedback} (general arena data, MultiPref, PRISM) and our main model comparisons from \Cref{sec:results:crossmodels} under our toolkit's built-in alternative metrics: standard Cohen's kappa with empirically estimated $p_e$ and agreement (\Cref{app:additional_metrics}).\footnote{Our toolkit's code names this metric \emph{accuracy} (\texttt{acc}). Its metric named \emph{agreement} (\texttt{agreement}) instead divides by all votes rather than by non-tie votes.} The main conclusions hold under every alternative, accounting for each alternative's own limitations, and each fails in a specific, measurable way that strength avoids.

\textbf{Metrics.} We use the following names in this section. \emph{Agreement} is $\texttt{agreement} = n_{\text{agreed}}/(n_{\text{agreed}} + n_{\text{disagreed}})$ over non-tie votes, and \emph{relevance} is the share of datapoints with such votes (\Cref{sec:method:metrics}). \emph{Unweighted kappa} is Cohen's kappa with $p_e$ fixed at 0.5, $\kappa_{0.5} = (\texttt{agreement} - 0.5)/(1 - 0.5)$. \emph{Strength}, our main metric, adds the relevance weight, $\texttt{strength} = \kappa_{0.5} \times \texttt{relevance}$. \emph{Standard kappa} is Cohen's kappa with an empirical estimate of $p_e$, $\kappa_{\text{std}} = (p_o - p_e)/(1 - p_e)$, where the observed agreement $p_o$ equals $\texttt{agreement}$ and $p_e$ is estimated from the observed distribution of labels (\Cref{sec:method:metrics}). \emph{Standard strength} is strength computed with standard kappa, $\kappa_{\text{std}} \times \texttt{relevance}$, so that it differs from strength only in $p_e$.

\textbf{Agreement loses signed scale and relevance weighting.} With $p_e$ fixed at 0.5, $\kappa_{0.5} = 2 \times \texttt{agreement} - 1$ over the same non-tie votes, so agreement orders traits identically to $\kappa_{0.5}$ (empirically verified cell-for-cell) and every conclusion remains consistent compared to that metric. Relative to strength it loses the meaningful zero separating encouraged from discouraged traits and the relevance weighting: ranking by agreement alone fails exactly as unweighted kappa does (see below).

\textbf{Standard kappa mistakes one-sided annotations for chance.} Standard kappa relies on an empirical $p_e$ that is computed from how often each annotation picks each stored position, which wrongly assumes that both annotators had access to the stored order (ours did not). On general arena data and PRISM, whose pairs are stored in randomized order, the estimate lands within 0.013 of 0.5 on every trait and nothing effectively changes (trait ordering of strength against standard strength Spearman 1.00, identical top-5/bottom-5). MultiPref shows the failure at moderate size: its stored pairs are not order-randomized (77\% of preference labels pick the second stored response), so empirical $p_e$ becomes non-random (0.32 to 0.72 across traits), incorrectly ignoring our trait annotator randomization, and standard strength values compress (mean/max shift 0.044/0.233, Spearman 0.93): \emph{is more concise} in \Cref{fig:app:results:multipref:top_bottom} falls from a strength of -0.29 to a standard strength of -0.06. Our model comparisons most directly illustrate why this metric cannot be used: the stored model order is fixed (model A first, model B second) and the target annotation always picks one model, thus the same position. Empirical $p_e$ reads this consistent selection as extreme bias, ignoring the randomization in our trait annotators: it equals observed agreement and standard kappa is 0 (or 0/0) on all 400 (model, trait) cells. Thus, this metric is not useful for our experiments.

\textbf{Unweighted kappa lets rare traits outrank ubiquitous ones.} Without the relevance weight, \emph{is more offensive}, applicable in 70 of 10,000 comparisons, becomes the most encouraged trait in general arena data (unweighted kappa +0.74 against strength +0.005), ahead of \emph{has more structured formatting} with 7,082 applicable comparisons (strength +0.17). The relevance factor is the reason \Cref{fig:results:arena:top_bottom} ranks widely observable traits first. Whilst analysis of more obscure traits is interesting, we think that, for many users, more widely relevant traits contain more useful information about a dataset. We make both analyses possible, but default to strength in the toolkit.

\clearpage

\subsection{Comparison to ICAI and VibeCheck}
\label{app:tool_comparison}

We ran \emph{Feedback Forensics}, \emph{ICAI} \citep{findeis2025inverse} and \emph{VibeCheck} \citep{dunlap2025VibeCheckDiscoverQuantifya} on three scenarios: the cross-annotator analysis of MultiPref (\Cref{fig:results:multipref:personality}) and the cross-model analysis (\Cref{fig:results:model_comparison}) from our paper, each on a 1k subset of the underlying datasets (three of ten models in the second), and a direct comparison of measurements on the same traits. To ensure comparability, all tools used the same judge model, \texttt{gpt-4o-mini-2024-07-18}, in their native configurations with some tuning of parameters: ICAI ran with 10, 40 and 400 clusters, and VibeCheck ran one discovery iteration, with its vibe de-duplication and parse retries staying on GPT-4o, as hard-coded in the original implementation. \Cref{tab:app:tool_comparison} summarizes the functional differences.

\begin{table}[h]
\centering
\caption{Functional differences between ICAI, VibeCheck and Feedback Forensics.}
\label{tab:app:tool_comparison}
\begin{minipage}[t]{1\textwidth}
\centering
\sffamily
\tablefontsize
\setlength{\arrayrulewidth}{0.5pt}
\newcommand{\shadedlabel}[1]{\multicolumn{1}{>{\columncolor{altrow}[\tabcolsep][\dimexpr 3\tabcolsep+0.22\textwidth+\arrayrulewidth\relax]\raggedright\arraybackslash}m{0.14\linewidth}|}{\textbf{#1}}}
\begin{tabular}{
    >{\raggedright\arraybackslash}m{0.14\linewidth}|
    >{\raggedright\arraybackslash}m{0.22\linewidth}
    >{\raggedright\arraybackslash}m{0.22\linewidth}
    >{\raggedright\arraybackslash}m{0.30\linewidth}
}
\toprule
 & \textbf{ICAI} & \textbf{VibeCheck} & \textbf{Feedback Forensics} \\
\midrule
\shadedlabel{Per-trait metrics} & Vote counts, no significance tests & \cellcolor{altrow}Per-vibe scores, coefficients with bootstrap standard deviations & \cellcolor{altrow}Strength with confidence interval and significance test, plus relevance, agreement and Cohen's $\kappa$ \\
\textbf{Same-trait comparability} & Vote shares comparable for a fixed trait, but not constitutional metrics & Coefficients standardized within each run, so not comparable across runs & Comparable across runs, one fixed $[-1, +1]$ axis shared across annotators, models and datasets \\
\shadedlabel{Multiple models} & No native model analysis & \cellcolor{altrow}Two models per run & \cellcolor{altrow}Natively supported, any number in one run \\
\textbf{User interface} & None & None (only external: Weights \& Biases, Plotly) & Interactive local app (\Cref{app:tutorial}) \\
\bottomrule
\end{tabular}
\end{minipage}
\end{table}

\begin{table}[h]
\centering
\caption{Traits discovered by ICAI and VibeCheck on a 1k MultiPref subset, one run per annotator (Scenario 1). ICAI lists the principles that survived its filter from the 400 tested. Neither tool returns a fully identical trait name for both annotation sets. Feedback Forensics ran without discovery, testing the curated traits in \Cref{tab:traits}.}
\label{tab:app:tool_comparison_discovery}
\begin{minipage}[t]{1\textwidth}
\centering
\sffamily
\tablefontsize
\begin{tabular}{
    >{\raggedright\arraybackslash}m{0.22\linewidth}
    >{\raggedright\arraybackslash}m{0.22\linewidth}
    >{\raggedright\arraybackslash}m{0.22\linewidth}
    >{\raggedright\arraybackslash}m{0.22\linewidth}
}
\toprule
\multicolumn{2}{c}{\textbf{ICAI}} & \multicolumn{2}{c}{\textbf{VibeCheck}} \\
\cmidrule(lr){1-2} \cmidrule(lr){3-4}
\textit{Human Expert} & \textit{Human Non-exp.} & \textit{Human Expert} & \textit{Human Non-exp.} \\
\midrule
\rowcolor{altrow} provides more detailed explanations & provides a more detailed analysis & Clarity and Structure & Clarity \\
provides detailed explanations and examples & uses a structured breakdown format & Completeness & Content Depth \\
\rowcolor{altrow} offers practical tips and strategies & provides more detailed information & Depth of Content & Creativity \\
 & provides more detailed categorization & Detail and Specificity & Engagement Level \\
\rowcolor{altrow}  & uses more varied vocabulary & Engagement and Responsiveness & Inclusivity of Perspectives \\
 &  & Safety Considerations & Practicality \\
\rowcolor{altrow}  &  & Technical Complexity & Safety Awareness \\
 &  & Tone and Empathy & Specificity \\
\rowcolor{altrow}  &  & Use of Examples & Structure \\
 &  &  & Tone \\
\bottomrule
\end{tabular}
\end{minipage}
\end{table}

\begin{figure}[t]
\centering
\begin{minipage}[t]{1\textwidth}
\centering
\sffamily
\tablefontsize
\setlength{\tabcolsep}{2pt}
\begin{tabular}{
    >{\raggedright\arraybackslash}m{0.27\linewidth}
    @{\hspace{6pt}}
    >{\centering\arraybackslash}m{0.14\linewidth}
    @{\hspace{6pt}}
    >{\centering\arraybackslash}m{0.14\linewidth}
    @{\hspace{6pt}}
    >{\centering\arraybackslash}m{0.14\linewidth}
    @{\hspace{6pt}}
    >{\centering\arraybackslash}m{0.14\linewidth}
}

\textbf{Generating a response that...} & \textbf{GPT-5.3-Chat} & \textbf{Claude-Opus-4.7} & \textbf{Gemini-3.1-Pro} & \textbf{Max diff} \\
\toprule
\rowcolor{altrow} is more concise & \cellcolor{poscolor!79.4}\makecell{0.34\\{\tiny(0.25, 0.42)}} & \cellcolor{negcolor!32.3}\makecell{\transparent{0.3}-0.13\\\transparent{0.3}{\tiny(-0.21, -0.03)}} & \cellcolor{negcolor!100.0}\makecell{-0.39\\{\tiny(-0.47, -0.30)}} & \cellcolor{lightgrey!100.0}0.73 \\
\addlinespace[\rowspacing]
is more verbose & \cellcolor{negcolor!65.4}\makecell{-0.26\\{\tiny(-0.35, -0.15)}} & \cellcolor{poscolor!57.4}\makecell{0.24\\{\tiny(0.14, 0.35)}} & \cellcolor{poscolor!100.0}\makecell{0.42\\{\tiny(0.32, 0.52)}} & \cellcolor{lightgrey!93.4}0.68 \\
\addlinespace[\rowspacing]
\rowcolor{altrow} has more structured formatting & \cellcolor{negcolor!83.8}\makecell{-0.33\\{\tiny(-0.39, -0.25)}} & \cellcolor{poscolor!17.0}\makecell{\transparent{0.3}0.07\\\transparent{0.3}{\tiny(-0.01, 0.16)}} & \cellcolor{poscolor!53.2}\makecell{0.23\\{\tiny(0.13, 0.31)}} & \cellcolor{lightgrey!76.0}0.55 \\
\addlinespace[\rowspacing]
provides more examples & \cellcolor{negcolor!40.8}\makecell{-0.16\\{\tiny(-0.23, -0.09)}} & \cellcolor{poscolor!36.9}\makecell{0.16\\{\tiny(0.08, 0.23)}} & \cellcolor{poscolor!86.5}\makecell{0.37\\{\tiny(0.29, 0.43)}} & \cellcolor{lightgrey!72.3}0.53 \\
\addlinespace[\rowspacing]
\rowcolor{altrow} provides a numbered list format & \cellcolor{negcolor!62.3}\makecell{-0.24\\{\tiny(-0.32, -0.16)}} & \cellcolor{poscolor!0.7}\makecell{\transparent{0.3}0.00\\\transparent{0.3}{\tiny(-0.08, 0.09)}} & \cellcolor{poscolor!41.8}\makecell{0.18\\{\tiny(0.09, 0.26)}} & \cellcolor{lightgrey!57.9}0.42 \\
\addlinespace[\rowspacing]
is more factually correct & \cellcolor{negcolor!44.6}\makecell{-0.17\\{\tiny(-0.26, -0.09)}} & \cellcolor{poscolor!46.8}\makecell{0.20\\{\tiny(0.11, 0.28)}} & \cellcolor{poscolor!43.3}\makecell{0.18\\{\tiny(0.10, 0.27)}} & \cellcolor{lightgrey!51.2}0.37 \\
\addlinespace[\rowspacing]
\rowcolor{altrow} makes more confident statements & \cellcolor{negcolor!18.5}\makecell{\transparent{0.3}-0.07\\\transparent{0.3}{\tiny(-0.16, 0.02)}} & \cellcolor{poscolor!30.5}\makecell{\transparent{0.3}0.13\\\transparent{0.3}{\tiny(0.04, 0.21)}} & \cellcolor{poscolor!69.5}\makecell{0.29\\{\tiny(0.20, 0.38)}} & \cellcolor{lightgrey!50.4}0.37 \\
\addlinespace[\rowspacing]
uses more formal language & \cellcolor{negcolor!74.6}\makecell{-0.29\\{\tiny(-0.38, -0.20)}} & \cellcolor{poscolor!14.2}\makecell{\transparent{0.3}0.06\\\transparent{0.3}{\tiny(-0.04, 0.15)}} & \cellcolor{poscolor!11.3}\makecell{\transparent{0.3}0.05\\\transparent{0.3}{\tiny(-0.05, 0.14)}} & \cellcolor{lightgrey!48.3}0.35 \\
\addlinespace[\rowspacing]
\rowcolor{altrow} uses more bold and italics text & \cellcolor{negcolor!16.9}\makecell{-0.07\\{\tiny(-0.10, -0.03)}} & \cellcolor{poscolor!25.5}\makecell{0.11\\{\tiny(0.06, 0.15)}} & \cellcolor{poscolor!65.2}\makecell{0.28\\{\tiny(0.22, 0.33)}} & \cellcolor{lightgrey!47.1}0.34 \\
\addlinespace[\rowspacing]
more strictly follows the requested output format & \cellcolor{negcolor!48.5}\makecell{-0.19\\{\tiny(-0.26, -0.11)}} & \cellcolor{poscolor!12.1}\makecell{\transparent{0.3}0.05\\\transparent{0.3}{\tiny(-0.03, 0.13)}} & \cellcolor{poscolor!3.5}\makecell{\transparent{0.3}0.02\\\transparent{0.3}{\tiny(-0.07, 0.10)}} & \cellcolor{lightgrey!33.1}0.24 \\
\end{tabular}
\end{minipage}
\caption{\textbf{Feedback Forensics on three models against GPT-4o, on 333 comparisons each (Scenario 2).} Sorted by max difference across rows. \strengthexpl{}}
\label{fig:app:results:tool_comparison_models}
\end{figure}

\begin{figure}[t]
\centering
\begin{minipage}[t]{1\textwidth}
\centering
\sffamily
\tablefontsize
\setlength{\tabcolsep}{2pt}
\begin{tabular}{
    >{\raggedright\arraybackslash}m{0.27\linewidth}
    @{\hspace{6pt}}
    >{\centering\arraybackslash}m{0.19\linewidth}
    @{\hspace{6pt}}
    >{\centering\arraybackslash}m{0.19\linewidth}
    @{\hspace{6pt}}
    >{\centering\arraybackslash}m{0.19\linewidth}
}

\textbf{Generating a response that...} & \textbf{Feedback Forensics} & \textbf{VibeCheck} & \textbf{Max diff} \\
\toprule
\rowcolor{altrow} is more concise & \cellcolor{negcolor!100.0}\makecell{-0.25\\{\tiny(-0.31, -0.18)}} & \cellcolor{poscolor!2.3}0.01 & \cellcolor{lightgrey!100.0}0.25 \\
\addlinespace[\rowspacing]
is more verbose & \cellcolor{poscolor!100.0}\makecell{0.23\\{\tiny(0.16, 0.30)}} & \cellcolor{negcolor!1.1}-0.00 & \cellcolor{lightgrey!93.5}0.24 \\
\addlinespace[\rowspacing]
\rowcolor{altrow} is more factually correct & \cellcolor{poscolor!85.4}\makecell{0.20\\{\tiny(0.14, 0.25)}} & \cellcolor{poscolor!1.2}0.00 & \cellcolor{lightgrey!77.8}0.20 \\
\addlinespace[\rowspacing]
makes more confident statements & \cellcolor{poscolor!88.9}\makecell{0.21\\{\tiny(0.15, 0.26)}} & \cellcolor{poscolor!50.9}0.12 & \cellcolor{lightgrey!35.1}0.09 \\
\addlinespace[\rowspacing]
\rowcolor{altrow} has more structured formatting & \cellcolor{poscolor!52.0}\makecell{0.12\\{\tiny(0.08, 0.17)}} & \cellcolor{poscolor!83.6}0.20 & \cellcolor{lightgrey!29.2}0.07 \\
\addlinespace[\rowspacing]
provides a numbered list format & \cellcolor{poscolor!52.6}\makecell{0.12\\{\tiny(0.07, 0.17)}} & \cellcolor{poscolor!27.5}0.06 & \cellcolor{lightgrey!23.2}0.06 \\
\addlinespace[\rowspacing]
\rowcolor{altrow} uses more formal language & \cellcolor{poscolor!71.9}\makecell{0.17\\{\tiny(0.10, 0.23)}} & \cellcolor{poscolor!51.5}0.12 & \cellcolor{lightgrey!18.9}0.05 \\
\addlinespace[\rowspacing]
provides more examples & \cellcolor{poscolor!78.9}\makecell{0.18\\{\tiny(0.13, 0.23)}} & \cellcolor{poscolor!95.9}0.22 & \cellcolor{lightgrey!15.7}0.04 \\
\addlinespace[\rowspacing]
\rowcolor{altrow} is more empathetic to the user & \cellcolor{poscolor!64.3}\makecell{0.15\\{\tiny(0.10, 0.21)}} & \cellcolor{poscolor!48.0}0.11 & \cellcolor{lightgrey!15.1}0.04 \\
\addlinespace[\rowspacing]
is more polite & \cellcolor{poscolor!52.6}\makecell{0.12\\{\tiny(0.07, 0.18)}} & \cellcolor{poscolor!52.6}0.12 & \cellcolor{lightgrey!0.0}0.00 \\
\end{tabular}
\end{minipage}
\caption{\textbf{Feedback Forensics and VibeCheck measuring the same 10 traits on a 1k MultiPref subset (Scenario 3, human expert annotations).} Feedback Forensics shows \emph{strength} with 95\% CI. VibeCheck, given these traits as fixed axes without discovery, shows its expert preference score, equivalent to our strength score but without a significance test. Max diff is the gap between them.}
\label{fig:app:results:tool_comparison_agreement}
\end{figure}

\textbf{Scenario 1: Expert against non-expert MultiPref annotations.} Feedback Forensics annotated the comparisons once and reused the trait annotations (independent of the preference labels) for both annotator columns, so its cost does not scale with the number of annotators. Both annotators encourage the same traits (verbosity, confident statements, examples), with the non-expert set encouraging most of the 10 strongest traits more strongly (7 of 10, including verbosity at 0.35 against 0.23, with the expert values in \Cref{fig:app:results:tool_comparison_agreement}), matching \Cref{fig:results:multipref:personality}. ICAI and VibeCheck each require a full run per annotator: the principles ICAI proposed were largely comparison-specific and its constitutions did not improve reconstruction over the unguided judge, and VibeCheck's runs shared no axis name and report on run-specific scales (\Cref{tab:app:tool_comparison_discovery}), so neither natively yields the cross-annotator table of \Cref{fig:results:multipref:personality}. ICAI's discovery rediscovered several curated traits (near-equivalents of conciseness, formality, empathy, structured formatting and examples) and surfaced one potential gap in our curated set, \emph{level of detail}. VibeCheck's discovery surfaced \emph{Completeness} and \emph{Technical Complexity}, two further potential gaps. VibeCheck's batch-based discovery may help it generalize more than the per-comparison proposal of Feedback Forensics (inherited from ICAI).

\textbf{Scenario 2: Three models against the GPT-4o reference.} Feedback Forensics measures all subjects with directly comparable results (\Cref{fig:app:results:tool_comparison_models}), as in \Cref{fig:results:model_comparison}: GPT-5.3-Chat is the concise outlier, with \emph{is more verbose} at -0.26 against 0.24 for Claude-Opus-4.7 and 0.42 for Gemini-3.1-Pro. This ordering matches \Cref{fig:app:results:claude_strength,fig:app:results:gpt_strength,fig:app:results:gemini_strength}, although individual values differ under this judge and subset, as would be expected. VibeCheck's three pairwise runs again shared no vibe verbatim (21 vibes under 19 distinct names), and ICAI does not natively support model analysis.

\textbf{Scenario 3: Agreement between the tools on the same traits.} On the expert annotations of Scenario 1, we forced VibeCheck to measure the 10 strongest traits of Feedback Forensics by supplying them as fixed axes, written by us with a high and a low pole, with discovery switched off. The two tools' signs agree on 8 of 10 (\Cref{fig:app:results:tool_comparison_agreement}). The two disagreements are \emph{is more verbose} and \emph{is more concise}, where VibeCheck's scores are close to zero.

\subsection{Extended pairwise feedback results}

\subsubsection{Across datasets}

First, in \Cref{fig:app:results:cross_strength,fig:app:results:cross_strength_b20,fig:app:results:cross_relevance,fig:app:results:cross_kappa,}, we provide a comprehensive comparison of the personality traits encouraged by the three preference datasets considered. 

\begin{figure}[ht]
    \centering
    \begin{minipage}[t]{1\textwidth}
\centering
\sffamily
\tablefontsize
\begin{tabular}{
    >{\raggedright\arraybackslash}m{0.3\linewidth}
    @{\hspace{13.5pt}} 
    >{\centering\arraybackslash}m{0.11363636363636363\linewidth}
    @{\hspace{13.5pt}} 
    >{\centering\arraybackslash}m{0.11363636363636363\linewidth}
    @{\hspace{13.5pt}} 
    >{\centering\arraybackslash}m{0.11363636363636363\linewidth}
    @{\hspace{13.5pt}} 
    >{\centering\arraybackslash}m{0.11363636363636363\linewidth}
}

\textbf{Generating a response that...} & \textbf{MultiPref} & \textbf{General arena data} & \textbf{PRISM} & \textbf{Max diff} \\
\toprule
\rowcolor{altrow} is more concise & \cellcolor{negcolor!100.0}\makecell{-0.29\\{\tiny(-0.30, -0.27)}} & \cellcolor{negcolor!32.1}\makecell{-0.09\\{\tiny(-0.11, -0.08)}} & \cellcolor{negcolor!78.5}\makecell{-0.23\\{\tiny(-0.25, -0.21)}} & \cellcolor{lightgrey!100.0}0.20 \\
\addlinespace[\rowspacing]
is more verbose & \cellcolor{poscolor!100.0}\makecell{0.34\\{\tiny(0.32, 0.35)}} & \cellcolor{poscolor!46.6}\makecell{0.16\\{\tiny(0.14, 0.18)}} & \cellcolor{poscolor!75.9}\makecell{0.26\\{\tiny(0.23, 0.28)}} & \cellcolor{lightgrey!91.4}0.18 \\
\addlinespace[\rowspacing]
\rowcolor{altrow} uses more bold and italics text & \cellcolor{poscolor!49.4}\makecell{0.17\\{\tiny(0.16, 0.18)}} & \cellcolor{poscolor!23.1}\makecell{0.08\\{\tiny(0.06, 0.09)}} & \cellcolor{poscolor!1.7}\makecell{0.01\\{\tiny(0.00, 0.01)}} & \cellcolor{lightgrey!81.8}0.16 \\
\addlinespace[\rowspacing]
is more polite & \cellcolor{poscolor!41.1}\makecell{0.14\\{\tiny(0.13, 0.15)}} & \cellcolor{poscolor!1.9}\makecell{\transparent{0.3}0.01\\\transparent{0.3}{\tiny(-0.01, 0.02)}} & \cellcolor{poscolor!44.7}\makecell{0.15\\{\tiny(0.13, 0.17)}} & \cellcolor{lightgrey!73.4}0.14 \\
\addlinespace[\rowspacing]
\rowcolor{altrow} uses more formal language & \cellcolor{poscolor!24.8}\makecell{0.08\\{\tiny(0.07, 0.10)}} & \cellcolor{poscolor!8.3}\makecell{0.03\\{\tiny(0.01, 0.04)}} & \cellcolor{poscolor!51.0}\makecell{0.17\\{\tiny(0.16, 0.19)}} & \cellcolor{lightgrey!73.1}0.14 \\
\addlinespace[\rowspacing]
has more structured formatting & \cellcolor{poscolor!69.5}\makecell{0.23\\{\tiny(0.22, 0.25)}} & \cellcolor{poscolor!51.1}\makecell{0.17\\{\tiny(0.16, 0.19)}} & \cellcolor{poscolor!26.9}\makecell{0.09\\{\tiny(0.08, 0.10)}} & \cellcolor{lightgrey!73.0}0.14 \\
\addlinespace[\rowspacing]
\rowcolor{altrow} uses more personal pronouns (I, we, you) & \cellcolor{poscolor!35.3}\makecell{0.12\\{\tiny(0.11, 0.13)}} & \cellcolor{negcolor!5.0}\makecell{\transparent{0.3}-0.01\\\transparent{0.3}{\tiny(-0.03, -0.00)}} & \cellcolor{poscolor!2.9}\makecell{\transparent{0.3}0.01\\\transparent{0.3}{\tiny(-0.01, 0.03)}} & \cellcolor{lightgrey!67.9}0.13 \\
\addlinespace[\rowspacing]
includes more ethical considerations & \cellcolor{poscolor!24.6}\makecell{0.08\\{\tiny(0.07, 0.09)}} & \cellcolor{negcolor!1.3}\makecell{\transparent{0.3}-0.00\\\transparent{0.3}{\tiny(-0.01, 0.00)}} & \cellcolor{poscolor!38.1}\makecell{0.13\\{\tiny(0.11, 0.14)}} & \cellcolor{lightgrey!67.2}0.13 \\
\addlinespace[\rowspacing]
\rowcolor{altrow} provides more examples & \cellcolor{poscolor!66.3}\makecell{0.22\\{\tiny(0.21, 0.24)}} & \cellcolor{poscolor!29.2}\makecell{0.10\\{\tiny(0.09, 0.11)}} & \cellcolor{poscolor!32.8}\makecell{0.11\\{\tiny(0.10, 0.12)}} & \cellcolor{lightgrey!63.6}0.12 \\
\addlinespace[\rowspacing]
has a friendlier tone & \cellcolor{poscolor!37.1}\makecell{0.12\\{\tiny(0.11, 0.14)}} & \cellcolor{poscolor!6.7}\makecell{0.02\\{\tiny(0.01, 0.03)}} & \cellcolor{poscolor!27.5}\makecell{0.09\\{\tiny(0.08, 0.11)}} & \cellcolor{lightgrey!52.2}0.10 \\
\addlinespace[\rowspacing]
\rowcolor{altrow} more actively engages with the user & \cellcolor{poscolor!28.3}\makecell{0.10\\{\tiny(0.09, 0.10)}} & \cellcolor{poscolor!2.3}\makecell{\transparent{0.3}0.01\\\transparent{0.3}{\tiny(-0.00, 0.02)}} & \cellcolor{poscolor!21.8}\makecell{0.07\\{\tiny(0.06, 0.09)}} & \cellcolor{lightgrey!44.5}0.09 \\
\addlinespace[\rowspacing]
is more empathetic to the user & \cellcolor{poscolor!30.9}\makecell{0.10\\{\tiny(0.09, 0.11)}} & \cellcolor{poscolor!5.6}\makecell{0.02\\{\tiny(0.01, 0.03)}} & \cellcolor{poscolor!29.1}\makecell{0.10\\{\tiny(0.08, 0.11)}} & \cellcolor{lightgrey!43.3}0.09 \\
\addlinespace[\rowspacing]
\rowcolor{altrow} uses more casual language & \cellcolor{poscolor!4.1}\makecell{0.01\\{\tiny(0.01, 0.02)}} & \cellcolor{poscolor!7.0}\makecell{0.02\\{\tiny(0.02, 0.03)}} & \cellcolor{negcolor!18.1}\makecell{-0.05\\{\tiny(-0.06, -0.04)}} & \cellcolor{lightgrey!38.6}0.08 \\
\addlinespace[\rowspacing]
ends with a follow-up question & \cellcolor{poscolor!6.7}\makecell{0.02\\{\tiny(0.02, 0.03)}} & \cellcolor{negcolor!9.4}\makecell{-0.03\\{\tiny(-0.04, -0.02)}} & \cellcolor{poscolor!13.3}\makecell{0.04\\{\tiny(0.03, 0.06)}} & \cellcolor{lightgrey!36.7}0.07 \\
\addlinespace[\rowspacing]
\rowcolor{altrow} has a more avoidant tone & \cellcolor{negcolor!0.6}\makecell{\transparent{0.3}-0.00\\\transparent{0.3}{\tiny(-0.01, 0.00)}} & \cellcolor{negcolor!24.5}\makecell{-0.07\\{\tiny(-0.08, -0.06)}} & \cellcolor{negcolor!19.2}\makecell{-0.06\\{\tiny(-0.07, -0.04)}} & \cellcolor{lightgrey!35.3}0.07 \\
\addlinespace[\rowspacing]
uses a more enthusiastic tone & \cellcolor{poscolor!25.9}\makecell{0.09\\{\tiny(0.08, 0.10)}} & \cellcolor{poscolor!9.3}\makecell{0.03\\{\tiny(0.02, 0.04)}} & \cellcolor{poscolor!5.7}\makecell{0.02\\{\tiny(0.01, 0.03)}} & \cellcolor{lightgrey!34.6}0.07 \\
\addlinespace[\rowspacing]
\rowcolor{altrow} contains less harmful information & \cellcolor{poscolor!5.8}\makecell{0.02\\{\tiny(0.01, 0.02)}} & \cellcolor{negcolor!6.3}\makecell{-0.02\\{\tiny(-0.02, -0.01)}} & \cellcolor{poscolor!13.6}\makecell{0.05\\{\tiny(0.04, 0.05)}} & \cellcolor{lightgrey!32.6}0.06 \\
\addlinespace[\rowspacing]
refuses to answer the question & \cellcolor{poscolor!2.6}\makecell{0.01\\{\tiny(0.00, 0.01)}} & \cellcolor{negcolor!16.8}\makecell{-0.05\\{\tiny(-0.05, -0.04)}} & \cellcolor{negcolor!18.0}\makecell{-0.05\\{\tiny(-0.06, -0.04)}} & \cellcolor{lightgrey!31.0}0.06 \\
\addlinespace[\rowspacing]
\rowcolor{altrow} acknowledges own limitations or uncertainty more & \cellcolor{poscolor!2.6}\makecell{\transparent{0.3}0.01\\\transparent{0.3}{\tiny(0.00, 0.02)}} & \cellcolor{negcolor!17.3}\makecell{-0.05\\{\tiny(-0.06, -0.04)}} & \cellcolor{negcolor!4.5}\makecell{\transparent{0.3}-0.01\\\transparent{0.3}{\tiny(-0.03, 0.00)}} & \cellcolor{lightgrey!29.8}0.06 \\
\addlinespace[\rowspacing]
is more factually correct & \cellcolor{poscolor!21.8}\makecell{0.07\\{\tiny(0.07, 0.08)}} & \cellcolor{poscolor!31.4}\makecell{0.11\\{\tiny(0.09, 0.12)}} & \cellcolor{poscolor!38.8}\makecell{0.13\\{\tiny(0.12, 0.14)}} & \cellcolor{lightgrey!29.2}0.06 \\
\end{tabular}
\end{minipage}

    \caption{\textbf{Comparison of investigated human feedback datasets in terms of \emph{strength} (top 20).} As usual, positive strength is shown in {\color{blue}blue} and negative strength in {\color{red}red}. MultiPref annotations considered here are a combination of all expert and non-expert human votes. Sorted by max difference. Whilst overall the personality traits each have similar strength across preference datasets, we observe some exceptions: annotations from general arena data do not appear to prefer \emph{polite} models as the other datasets do. Similarly, general arena data annotations do (approximately) not actively encourage \emph{less harmful} responses or responses with \emph{ethical considerations}.}
    \label{fig:app:results:cross_strength}
\end{figure}

\begin{figure}[ht]
    \centering
    \begin{minipage}[t]{1\textwidth}
\centering
\sffamily
\tablefontsize
\begin{tabular}{
    >{\raggedright\arraybackslash}m{0.3\linewidth}
    @{\hspace{13.5pt}} 
    >{\centering\arraybackslash}m{0.11363636363636363\linewidth}
    @{\hspace{13.5pt}} 
    >{\centering\arraybackslash}m{0.11363636363636363\linewidth}
    @{\hspace{13.5pt}} 
    >{\centering\arraybackslash}m{0.11363636363636363\linewidth}
    @{\hspace{13.5pt}} 
    >{\centering\arraybackslash}m{0.11363636363636363\linewidth}
}

\textbf{Generating a response that...} & \textbf{MultiPref} & \textbf{General arena data} & \textbf{PRISM} & \textbf{Max diff} \\
\toprule
\rowcolor{altrow} compliments the user's question or prompt & \cellcolor{poscolor!19.1}\makecell{0.06\\{\tiny(0.06, 0.07)}} & \cellcolor{poscolor!6.6}\makecell{0.02\\{\tiny(0.01, 0.03)}} & \cellcolor{poscolor!3.1}\makecell{0.01\\{\tiny(0.00, 0.02)}} & \cellcolor{lightgrey!27.5}0.05 \\
\addlinespace[\rowspacing]
provides a numbered list format & \cellcolor{poscolor!34.9}\makecell{0.12\\{\tiny(0.11, 0.13)}} & \cellcolor{poscolor!22.7}\makecell{0.08\\{\tiny(0.06, 0.09)}} & \cellcolor{poscolor!21.2}\makecell{0.07\\{\tiny(0.06, 0.08)}} & \cellcolor{lightgrey!23.4}0.05 \\
\addlinespace[\rowspacing]
\rowcolor{altrow} expresses more emotion & \cellcolor{poscolor!13.3}\makecell{0.04\\{\tiny(0.04, 0.05)}} & \cellcolor{poscolor!4.7}\makecell{0.02\\{\tiny(0.01, 0.02)}} & \cellcolor{poscolor!0.0}\makecell{\transparent{0.3}0.00\\\transparent{0.3}{\tiny(-0.01, 0.01)}} & \cellcolor{lightgrey!22.6}0.04 \\
\addlinespace[\rowspacing]
is more optimistic & \cellcolor{poscolor!14.8}\makecell{0.05\\{\tiny(0.04, 0.06)}} & \cellcolor{poscolor!5.5}\makecell{0.02\\{\tiny(0.01, 0.03)}} & \cellcolor{poscolor!17.9}\makecell{0.06\\{\tiny(0.05, 0.07)}} & \cellcolor{lightgrey!21.3}0.04 \\
\addlinespace[\rowspacing]
\rowcolor{altrow} is more creative and original & \cellcolor{poscolor!21.6}\makecell{0.07\\{\tiny(0.07, 0.08)}} & \cellcolor{poscolor!20.0}\makecell{0.07\\{\tiny(0.06, 0.08)}} & \cellcolor{poscolor!11.2}\makecell{0.04\\{\tiny(0.03, 0.04)}} & \cellcolor{lightgrey!17.8}0.04 \\
\addlinespace[\rowspacing]
agrees more with the user & \cellcolor{poscolor!0.4}\makecell{\transparent{0.3}0.00\\\transparent{0.3}{\tiny(-0.00, 0.01)}} & \cellcolor{poscolor!10.7}\makecell{0.04\\{\tiny(0.03, 0.04)}} & \cellcolor{poscolor!4.1}\makecell{0.01\\{\tiny(0.01, 0.02)}} & \cellcolor{lightgrey!17.5}0.03 \\
\addlinespace[\rowspacing]
\rowcolor{altrow} makes more confident statements & \cellcolor{poscolor!19.0}\makecell{0.06\\{\tiny(0.05, 0.07)}} & \cellcolor{poscolor!28.8}\makecell{0.10\\{\tiny(0.08, 0.11)}} & \cellcolor{poscolor!28.4}\makecell{0.10\\{\tiny(0.08, 0.11)}} & \cellcolor{lightgrey!16.9}0.03 \\
\addlinespace[\rowspacing]
actively engages the reader with rhetorical questions & \cellcolor{poscolor!6.7}\makecell{0.02\\{\tiny(0.02, 0.03)}} & \cellcolor{poscolor!3.0}\makecell{\transparent{0.3}0.01\\\transparent{0.3}{\tiny(0.00, 0.02)}} & \cellcolor{negcolor!1.9}\makecell{\transparent{0.3}-0.01\\\transparent{0.3}{\tiny(-0.01, 0.00)}} & \cellcolor{lightgrey!14.1}0.03 \\
\addlinespace[\rowspacing]
\rowcolor{altrow} agrees with user even if factually incorrect & \cellcolor{negcolor!2.4}\makecell{-0.01\\{\tiny(-0.01, -0.00)}} & \cellcolor{poscolor!4.7}\makecell{0.02\\{\tiny(0.01, 0.02)}} & \cellcolor{negcolor!1.3}\makecell{\transparent{0.3}-0.00\\\transparent{0.3}{\tiny(-0.01, 0.00)}} & \cellcolor{lightgrey!11.6}0.02 \\
\addlinespace[\rowspacing]
includes more references to other sources & \cellcolor{poscolor!6.8}\makecell{0.02\\{\tiny(0.02, 0.03)}} & \cellcolor{poscolor!3.1}\makecell{0.01\\{\tiny(0.00, 0.02)}} & \cellcolor{poscolor!0.5}\makecell{\transparent{0.3}0.00\\\transparent{0.3}{\tiny(-0.00, 0.01)}} & \cellcolor{lightgrey!10.7}0.02 \\
\addlinespace[\rowspacing]
\rowcolor{altrow} uses more humour & \cellcolor{poscolor!3.1}\makecell{0.01\\{\tiny(0.01, 0.01)}} & \cellcolor{poscolor!5.8}\makecell{0.02\\{\tiny(0.01, 0.02)}} & \cellcolor{negcolor!0.5}\makecell{\transparent{0.3}-0.00\\\transparent{0.3}{\tiny(-0.00, 0.00)}} & \cellcolor{lightgrey!10.7}0.02 \\
\addlinespace[\rowspacing]
reinforces user's beliefs more & \cellcolor{poscolor!0}\makecell{\transparent{0.3}0.00\\\transparent{0.3}{\tiny(-0.00, 0.00)}} & \cellcolor{poscolor!5.0}\makecell{0.02\\{\tiny(0.01, 0.02)}} & \cellcolor{poscolor!3.3}\makecell{0.01\\{\tiny(0.00, 0.02)}} & \cellcolor{lightgrey!8.5}0.02 \\
\addlinespace[\rowspacing]
\rowcolor{altrow} more strictly follows the requested output format & \cellcolor{poscolor!17.1}\makecell{0.06\\{\tiny(0.05, 0.07)}} & \cellcolor{poscolor!20.8}\makecell{0.07\\{\tiny(0.06, 0.08)}} & \cellcolor{poscolor!16.2}\makecell{0.05\\{\tiny(0.05, 0.06)}} & \cellcolor{lightgrey!8.0}0.02 \\
\addlinespace[\rowspacing]
provides conclusions without full reasoning & \cellcolor{negcolor!2.8}\makecell{-0.01\\{\tiny(-0.01, -0.01)}} & \cellcolor{negcolor!2.1}\makecell{-0.01\\{\tiny(-0.01, -0.00)}} & \cellcolor{negcolor!6.9}\makecell{-0.02\\{\tiny(-0.02, -0.02)}} & \cellcolor{lightgrey!7.0}0.01 \\
\addlinespace[\rowspacing]
\rowcolor{altrow} is more offensive & \cellcolor{negcolor!1.8}\makecell{-0.01\\{\tiny(-0.01, -0.00)}} & \cellcolor{poscolor!1.5}\makecell{0.01\\{\tiny(0.00, 0.01)}} & \cellcolor{negcolor!2.8}\makecell{-0.01\\{\tiny(-0.01, -0.01)}} & \cellcolor{lightgrey!6.7}0.01 \\
\addlinespace[\rowspacing]
uses more mathematical symbols and notation & \cellcolor{poscolor!0.8}\makecell{\transparent{0.3}0.00\\\transparent{0.3}{\tiny(-0.00, 0.01)}} & \cellcolor{poscolor!3.6}\makecell{0.01\\{\tiny(0.01, 0.02)}} & \cellcolor{negcolor!0.1}\makecell{\transparent{0.3}-0.00\\\transparent{0.3}{\tiny(-0.00, 0.00)}} & \cellcolor{lightgrey!6.4}0.01 \\
\addlinespace[\rowspacing]
\rowcolor{altrow} includes inappropriate language & \cellcolor{negcolor!1.0}\makecell{-0.00\\{\tiny(-0.00, -0.00)}} & \cellcolor{poscolor!1.1}\makecell{0.00\\{\tiny(0.00, 0.01)}} & \cellcolor{negcolor!1.2}\makecell{-0.00\\{\tiny(-0.01, -0.00)}} & \cellcolor{lightgrey!3.7}0.01 \\
\addlinespace[\rowspacing]
suggests illegal activities & \cellcolor{negcolor!1.0}\makecell{-0.00\\{\tiny(-0.00, -0.00)}} & \cellcolor{poscolor!0.9}\makecell{0.00\\{\tiny(0.00, 0.00)}} & \cellcolor{negcolor!0.5}\makecell{\transparent{0.3}-0.00\\\transparent{0.3}{\tiny(-0.00, 0.00)}} & \cellcolor{lightgrey!3.0}0.01 \\
\addlinespace[\rowspacing]
\rowcolor{altrow} uses more emojis & \cellcolor{poscolor!0.2}\makecell{\transparent{0.3}0.00\\\transparent{0.3}{\tiny(-0.00, 0.00)}} & \cellcolor{poscolor!0.6}\makecell{\transparent{0.3}0.00\\\transparent{0.3}{\tiny(-0.00, 0.00)}} & \cellcolor{negcolor!0.6}\makecell{\transparent{0.3}-0.00\\\transparent{0.3}{\tiny(-0.00, 0.00)}} & \cellcolor{lightgrey!1.8}0.00 \\
\addlinespace[\rowspacing]
reinforces user's anger more & \cellcolor{poscolor!0}\makecell{\transparent{0.3}0.00\\\transparent{0.3}{\tiny(-0.00, 0.00)}} & \cellcolor{poscolor!0.2}\makecell{\transparent{0.3}0.00\\\transparent{0.3}{\tiny(-0.00, 0.00)}} & \cellcolor{negcolor!0.6}\makecell{\transparent{0.3}-0.00\\\transparent{0.3}{\tiny(-0.00, 0.00)}} & \cellcolor{lightgrey!1.4}0.00 \\
\end{tabular}
\end{minipage}

    \caption{\textbf{Comparison of investigated human feedback datasets in terms of \emph{strength} (bottom 20).} As usual, positive strength is shown in {\color{blue}blue} and negative strength in {\color{red}red}. MultiPref annotations considered here are a combination of all expert and non-expert human votes. Sorted by max difference. Whilst overall the personality traits each have similar strength across preference datasets, we observe some exceptions: annotations from general arena data do not appear to prefer \emph{polite} models as the other datasets do. Similarly, the general arena data annotations do (approximately) not actively encourage \emph{less harmful} responses or responses with \emph{ethical considerations}.}
    \label{fig:app:results:cross_strength_b20}
\end{figure}

\begin{figure}[ht]
    \centering
    \begin{minipage}[t]{1\textwidth}
\centering
\sffamily
\tablefontsize
\begin{tabular}{
    >{\raggedright\arraybackslash}m{0.3\linewidth}
    @{\hspace{13.5pt}} 
    >{\centering\arraybackslash}m{0.11363636363636363\linewidth}
    @{\hspace{13.5pt}} 
    >{\centering\arraybackslash}m{0.11363636363636363\linewidth}
    @{\hspace{13.5pt}} 
    >{\centering\arraybackslash}m{0.11363636363636363\linewidth}
    @{\hspace{13.5pt}} 
    >{\centering\arraybackslash}m{0.11363636363636363\linewidth}
}

\textbf{Generating a response that...} & \textbf{MultiPref} & \textbf{General arena data} & \textbf{PRISM} & \textbf{Max diff} \\
\toprule
\rowcolor{altrow} uses more bold and italics text & \cellcolor{poscolor!32.7}0.31 & \cellcolor{poscolor!62.6}0.60 & \cellcolor{poscolor!1.7}0.02 & \cellcolor{lightgrey!100.0}0.59 \\
\addlinespace[\rowspacing]
has more structured formatting & \cellcolor{poscolor!49.2}0.47 & \cellcolor{poscolor!73.7}0.71 & \cellcolor{poscolor!20.3}0.20 & \cellcolor{lightgrey!87.6}0.51 \\
\addlinespace[\rowspacing]
\rowcolor{altrow} provides a numbered list format & \cellcolor{poscolor!31.5}0.30 & \cellcolor{poscolor!50.5}0.48 & \cellcolor{poscolor!16.3}0.16 & \cellcolor{lightgrey!56.0}0.33 \\
\addlinespace[\rowspacing]
is more concise & \cellcolor{poscolor!68.5}0.66 & \cellcolor{poscolor!48.7}0.47 & \cellcolor{poscolor!81.8}0.79 & \cellcolor{lightgrey!54.3}0.32 \\
\addlinespace[\rowspacing]
\rowcolor{altrow} includes more ethical considerations & \cellcolor{poscolor!27.7}0.27 & \cellcolor{poscolor!19.6}0.19 & \cellcolor{poscolor!50.0}0.48 & \cellcolor{lightgrey!49.9}0.29 \\
\addlinespace[\rowspacing]
is more polite & \cellcolor{poscolor!35.4}0.34 & \cellcolor{poscolor!43.9}0.42 & \cellcolor{poscolor!60.1}0.58 & \cellcolor{lightgrey!40.7}0.24 \\
\addlinespace[\rowspacing]
\rowcolor{altrow} acknowledges own limitations or uncertainty more & \cellcolor{poscolor!14.8}0.14 & \cellcolor{poscolor!28.2}0.27 & \cellcolor{poscolor!37.9}0.36 & \cellcolor{lightgrey!37.9}0.22 \\
\addlinespace[\rowspacing]
has a more avoidant tone & \cellcolor{poscolor!8.7}0.08 & \cellcolor{poscolor!11.2}0.11 & \cellcolor{poscolor!31.4}0.30 & \cellcolor{lightgrey!37.2}0.22 \\
\addlinespace[\rowspacing]
\rowcolor{altrow} uses more formal language & \cellcolor{poscolor!38.5}0.37 & \cellcolor{poscolor!48.0}0.46 & \cellcolor{poscolor!61.1}0.59 & \cellcolor{lightgrey!37.0}0.22 \\
\addlinespace[\rowspacing]
uses more personal pronouns (I, we, you) & \cellcolor{poscolor!46.6}0.45 & \cellcolor{poscolor!50.4}0.48 & \cellcolor{poscolor!68.9}0.66 & \cellcolor{lightgrey!36.5}0.21 \\
\addlinespace[\rowspacing]
\rowcolor{altrow} makes more confident statements & \cellcolor{poscolor!23.3}0.22 & \cellcolor{poscolor!44.4}0.43 & \cellcolor{poscolor!45.5}0.44 & \cellcolor{lightgrey!36.5}0.21 \\
\addlinespace[\rowspacing]
provides more examples & \cellcolor{poscolor!47.4}0.46 & \cellcolor{poscolor!43.3}0.42 & \cellcolor{poscolor!29.4}0.28 & \cellcolor{lightgrey!29.5}0.17 \\
\addlinespace[\rowspacing]
\rowcolor{altrow} is more factually correct & \cellcolor{poscolor!15.1}0.14 & \cellcolor{poscolor!31.6}0.30 & \cellcolor{poscolor!26.9}0.26 & \cellcolor{lightgrey!27.2}0.16 \\
\addlinespace[\rowspacing]
more strictly follows the requested output format & \cellcolor{poscolor!18.9}0.18 & \cellcolor{poscolor!26.9}0.26 & \cellcolor{poscolor!11.3}0.11 & \cellcolor{lightgrey!25.7}0.15 \\
\addlinespace[\rowspacing]
\rowcolor{altrow} ends with a follow-up question & \cellcolor{poscolor!13.3}0.13 & \cellcolor{poscolor!20.4}0.20 & \cellcolor{poscolor!28.9}0.28 & \cellcolor{lightgrey!25.7}0.15 \\
\addlinespace[\rowspacing]
more actively engages with the user & \cellcolor{poscolor!25.4}0.24 & \cellcolor{poscolor!35.9}0.34 & \cellcolor{poscolor!40.9}0.39 & \cellcolor{lightgrey!25.4}0.15 \\
\addlinespace[\rowspacing]
\rowcolor{altrow} is more empathetic to the user & \cellcolor{poscolor!23.9}0.23 & \cellcolor{poscolor!27.3}0.26 & \cellcolor{poscolor!37.8}0.36 & \cellcolor{lightgrey!22.9}0.13 \\
\addlinespace[\rowspacing]
is more creative and original & \cellcolor{poscolor!14.3}0.14 & \cellcolor{poscolor!23.3}0.22 & \cellcolor{poscolor!10.5}0.10 & \cellcolor{lightgrey!20.9}0.12 \\
\addlinespace[\rowspacing]
\rowcolor{altrow} has a friendlier tone & \cellcolor{poscolor!31.3}0.30 & \cellcolor{poscolor!37.4}0.36 & \cellcolor{poscolor!42.6}0.41 & \cellcolor{lightgrey!18.6}0.11 \\
\addlinespace[\rowspacing]
refuses to answer the question & \cellcolor{poscolor!5.0}0.05 & \cellcolor{poscolor!6.8}0.07 & \cellcolor{poscolor!16.2}0.16 & \cellcolor{lightgrey!18.3}0.11 \\
\addlinespace[\rowspacing]
\rowcolor{altrow} uses more casual language & \cellcolor{poscolor!7.2}0.07 & \cellcolor{poscolor!12.2}0.12 & \cellcolor{poscolor!18.0}0.17 & \cellcolor{lightgrey!17.8}0.10 \\
\addlinespace[\rowspacing]
compliments the user's question or prompt & \cellcolor{poscolor!14.3}0.14 & \cellcolor{poscolor!17.8}0.17 & \cellcolor{poscolor!7.1}0.07 & \cellcolor{lightgrey!17.6}0.10 \\
\addlinespace[\rowspacing]
\rowcolor{altrow} is more optimistic & \cellcolor{poscolor!13.0}0.13 & \cellcolor{poscolor!12.9}0.12 & \cellcolor{poscolor!23.4}0.22 & \cellcolor{lightgrey!17.2}0.10 \\
\addlinespace[\rowspacing]
agrees more with the user & \cellcolor{poscolor!4.3}0.04 & \cellcolor{poscolor!11.9}0.11 & \cellcolor{poscolor!11.7}0.11 & \cellcolor{lightgrey!12.5}0.07 \\
\addlinespace[\rowspacing]
\rowcolor{altrow} contains less harmful information & \cellcolor{poscolor!6.1}0.06 & \cellcolor{poscolor!5.8}0.06 & \cellcolor{poscolor!13.0}0.12 & \cellcolor{lightgrey!11.8}0.07 \\
\addlinespace[\rowspacing]
reinforces user's beliefs more & \cellcolor{poscolor!1.9}0.02 & \cellcolor{poscolor!4.8}0.05 & \cellcolor{poscolor!8.7}0.08 & \cellcolor{lightgrey!11.2}0.07 \\
\addlinespace[\rowspacing]
\rowcolor{altrow} uses more mathematical symbols and notation & \cellcolor{poscolor!2.7}0.03 & \cellcolor{poscolor!6.3}0.06 & \cellcolor{poscolor!0.1}0.00 & \cellcolor{lightgrey!10.1}0.06 \\
\addlinespace[\rowspacing]
expresses more emotion & \cellcolor{poscolor!10.7}0.10 & \cellcolor{poscolor!11.3}0.11 & \cellcolor{poscolor!15.9}0.15 & \cellcolor{lightgrey!8.7}0.05 \\
\addlinespace[\rowspacing]
\rowcolor{altrow} uses more humour & \cellcolor{poscolor!2.4}0.02 & \cellcolor{poscolor!5.4}0.05 & \cellcolor{poscolor!1.0}0.01 & \cellcolor{lightgrey!7.2}0.04 \\
\addlinespace[\rowspacing]
actively engages the reader with rhetorical questions & \cellcolor{poscolor!5.8}0.06 & \cellcolor{poscolor!9.5}0.09 & \cellcolor{poscolor!10.0}0.10 & \cellcolor{lightgrey!6.9}0.04 \\
\addlinespace[\rowspacing]
\rowcolor{altrow} uses a more enthusiastic tone & \cellcolor{poscolor!19.9}0.19 & \cellcolor{poscolor!19.0}0.18 & \cellcolor{poscolor!16.2}0.16 & \cellcolor{lightgrey!6.1}0.04 \\
\addlinespace[\rowspacing]
agrees with user even if factually incorrect & \cellcolor{poscolor!2.0}0.02 & \cellcolor{poscolor!5.7}0.05 & \cellcolor{poscolor!5.3}0.05 & \cellcolor{lightgrey!6.1}0.04 \\
\addlinespace[\rowspacing]
\rowcolor{altrow} provides conclusions without full reasoning & \cellcolor{poscolor!1.6}0.02 & \cellcolor{poscolor!2.4}0.02 & \cellcolor{poscolor!4.6}0.04 & \cellcolor{lightgrey!5.0}0.03 \\
\addlinespace[\rowspacing]
includes more references to other sources & \cellcolor{poscolor!5.4}0.05 & \cellcolor{poscolor!8.3}0.08 & \cellcolor{poscolor!5.6}0.05 & \cellcolor{lightgrey!4.6}0.03 \\
\addlinespace[\rowspacing]
\rowcolor{altrow} uses more emojis & \cellcolor{poscolor!0.6}0.01 & \cellcolor{poscolor!2.5}0.02 & \cellcolor{poscolor!0.8}0.01 & \cellcolor{lightgrey!3.1}0.02 \\
\addlinespace[\rowspacing]
is more verbose & \cellcolor{poscolor!99.7}0.96 & \cellcolor{poscolor!100.0}0.96 & \cellcolor{poscolor!98.2}0.94 & \cellcolor{lightgrey!2.9}0.02 \\
\addlinespace[\rowspacing]
\rowcolor{altrow} is more offensive & \cellcolor{poscolor!0.8}0.01 & \cellcolor{poscolor!0.7}0.01 & \cellcolor{poscolor!2.0}0.02 & \cellcolor{lightgrey!2.0}0.01 \\
\addlinespace[\rowspacing]
reinforces user's anger more & \cellcolor{poscolor!0.1}0.00 & \cellcolor{poscolor!0.4}0.00 & \cellcolor{poscolor!1.0}0.01 & \cellcolor{lightgrey!1.5}0.01 \\
\addlinespace[\rowspacing]
\rowcolor{altrow} includes inappropriate language & \cellcolor{poscolor!0.4}0.00 & \cellcolor{poscolor!0.6}0.01 & \cellcolor{poscolor!0.7}0.01 & \cellcolor{lightgrey!0.5}0.00 \\
\addlinespace[\rowspacing]
suggests illegal activities & \cellcolor{poscolor!0.7}0.01 & \cellcolor{poscolor!0.7}0.01 & \cellcolor{poscolor!0.6}0.01 & \cellcolor{lightgrey!0.2}0.00 \\
\end{tabular}
\end{minipage}

    \caption{\textbf{Comparison of investigated human feedback datasets in terms of \emph{relevance}.} Strong relevance is shown in {\color{blue}blue}. We observe notable differences between the datasets that are likely explained by the difference in domains. Whereas MultiPref and general arena data include a lot of text with \emph{structured formatting} (above 60\%), PRISM (focused on value-laden topics) does not (below 30\%). On the other hand we observe that \emph{friendlier} and \emph{more polite} tone appear to be more relevant in the PRISM context.}
    \label{fig:app:results:cross_relevance}
\end{figure}

\begin{figure}[ht]
    \centering
    \begin{minipage}[t]{1\textwidth}
\centering
\sffamily
\tablefontsize
\begin{tabular}{
    >{\raggedright\arraybackslash}m{0.3\linewidth}
    @{\hspace{13.5pt}} 
    >{\centering\arraybackslash}m{0.11363636363636363\linewidth}
    @{\hspace{13.5pt}} 
    >{\centering\arraybackslash}m{0.11363636363636363\linewidth}
    @{\hspace{13.5pt}} 
    >{\centering\arraybackslash}m{0.11363636363636363\linewidth}
    @{\hspace{13.5pt}} 
    >{\centering\arraybackslash}m{0.11363636363636363\linewidth}
}

\textbf{Generating a response that...} & \textbf{MultiPref} & \textbf{General arena data} & \textbf{PRISM} & \textbf{Max diff} \\
\toprule
\rowcolor{altrow} includes inappropriate language & \cellcolor{negcolor!100.0}-0.75 & \cellcolor{poscolor!94.7}0.70 & \cellcolor{negcolor!63.2}-0.47 & \cellcolor{lightgrey!100.0}1.45 \\
\addlinespace[\rowspacing]
is more offensive & \cellcolor{negcolor!93.2}-0.70 & \cellcolor{poscolor!100.0}0.74 & \cellcolor{negcolor!56.9}-0.43 & \cellcolor{lightgrey!99.2}1.44 \\
\addlinespace[\rowspacing]
\rowcolor{altrow} refuses to answer the question & \cellcolor{poscolor!24.5}0.18 & \cellcolor{negcolor!99.4}-0.75 & \cellcolor{negcolor!44.6}-0.33 & \cellcolor{lightgrey!63.8}0.93 \\
\addlinespace[\rowspacing]
suggests illegal activities & \cellcolor{negcolor!54.5}-0.41 & \cellcolor{poscolor!55.0}0.41 & \cellcolor{negcolor!29.9}-0.22 & \cellcolor{lightgrey!56.2}0.82 \\
\addlinespace[\rowspacing]
\rowcolor{altrow} contains less harmful information & \cellcolor{poscolor!45.0}0.33 & \cellcolor{negcolor!43.5}-0.33 & \cellcolor{poscolor!49.4}0.37 & \cellcolor{lightgrey!47.7}0.69 \\
\addlinespace[\rowspacing]
agrees with user even if factually incorrect & \cellcolor{negcolor!48.2}-0.36 & \cellcolor{poscolor!39.4}0.29 & \cellcolor{negcolor!10.2}-0.08 & \cellcolor{lightgrey!45.0}0.65 \\
\addlinespace[\rowspacing]
\rowcolor{altrow} has a more avoidant tone & \cellcolor{negcolor!2.7}-0.02 & \cellcolor{negcolor!88.1}-0.66 & \cellcolor{negcolor!24.5}-0.18 & \cellcolor{lightgrey!44.1}0.64 \\
\addlinespace[\rowspacing]
uses more humour & \cellcolor{poscolor!61.7}0.46 & \cellcolor{poscolor!51.0}0.38 & \cellcolor{negcolor!21.1}-0.16 & \cellcolor{lightgrey!42.4}0.62 \\
\addlinespace[\rowspacing]
\rowcolor{altrow} uses more casual language & \cellcolor{poscolor!26.8}0.20 & \cellcolor{poscolor!26.8}0.20 & \cellcolor{negcolor!40.3}-0.30 & \cellcolor{lightgrey!34.5}0.50 \\
\addlinespace[\rowspacing]
uses more mathematical symbols and notation & \cellcolor{poscolor!13.7}0.10 & \cellcolor{poscolor!26.8}0.20 & \cellcolor{negcolor!36.4}-0.27 & \cellcolor{lightgrey!32.5}0.47 \\
\addlinespace[\rowspacing]
\rowcolor{altrow} actively engages the reader with rhetorical questions & \cellcolor{poscolor!54.0}0.40 & \cellcolor{poscolor!14.7}0.11 & \cellcolor{negcolor!7.5}-0.06 & \cellcolor{lightgrey!31.5}0.46 \\
\addlinespace[\rowspacing]
expresses more emotion & \cellcolor{poscolor!58.6}0.44 & \cellcolor{poscolor!19.6}0.15 & \cellcolor{poscolor!0.1}0.00 & \cellcolor{lightgrey!29.9}0.43 \\
\addlinespace[\rowspacing]
\rowcolor{altrow} includes more references to other sources & \cellcolor{poscolor!58.8}0.44 & \cellcolor{poscolor!17.5}0.13 & \cellcolor{poscolor!4.4}0.03 & \cellcolor{lightgrey!27.8}0.40 \\
\addlinespace[\rowspacing]
uses more bold and italics text & \cellcolor{poscolor!71.3}0.53 & \cellcolor{poscolor!17.4}0.13 & \cellcolor{poscolor!47.7}0.35 & \cellcolor{lightgrey!27.5}0.40 \\
\addlinespace[\rowspacing]
\rowcolor{altrow} is more polite & \cellcolor{poscolor!54.9}0.41 & \cellcolor{poscolor!2.0}0.01 & \cellcolor{poscolor!35.0}0.26 & \cellcolor{lightgrey!27.0}0.39 \\
\addlinespace[\rowspacing]
reinforces user's anger more & \cellcolor{poscolor!0}0.00 & \cellcolor{poscolor!26.9}0.20 & \cellcolor{negcolor!25.3}-0.19 & \cellcolor{lightgrey!26.8}0.39 \\
\addlinespace[\rowspacing]
\rowcolor{altrow} is more empathetic to the user & \cellcolor{poscolor!60.9}0.45 & \cellcolor{poscolor!9.7}0.07 & \cellcolor{poscolor!36.3}0.27 & \cellcolor{lightgrey!26.2}0.38 \\
\addlinespace[\rowspacing]
more actively engages with the user & \cellcolor{poscolor!52.4}0.39 & \cellcolor{poscolor!3.0}0.02 & \cellcolor{poscolor!25.2}0.19 & \cellcolor{lightgrey!25.3}0.37 \\
\addlinespace[\rowspacing]
\rowcolor{altrow} reinforces user's beliefs more & \cellcolor{poscolor!0}0.00 & \cellcolor{poscolor!48.6}0.36 & \cellcolor{poscolor!18.1}0.13 & \cellcolor{lightgrey!24.8}0.36 \\
\addlinespace[\rowspacing]
has a friendlier tone & \cellcolor{poscolor!55.9}0.42 & \cellcolor{poscolor!8.4}0.06 & \cellcolor{poscolor!30.4}0.23 & \cellcolor{lightgrey!24.3}0.35 \\
\addlinespace[\rowspacing]
\rowcolor{altrow} compliments the user's question or prompt & \cellcolor{poscolor!63.2}0.47 & \cellcolor{poscolor!17.5}0.13 & \cellcolor{poscolor!20.5}0.15 & \cellcolor{lightgrey!23.4}0.34 \\
\addlinespace[\rowspacing]
uses a more enthusiastic tone & \cellcolor{poscolor!61.3}0.46 & \cellcolor{poscolor!23.0}0.17 & \cellcolor{poscolor!16.7}0.12 & \cellcolor{lightgrey!22.8}0.33 \\
\addlinespace[\rowspacing]
\rowcolor{altrow} includes more ethical considerations & \cellcolor{poscolor!41.9}0.31 & \cellcolor{negcolor!2.7}-0.02 & \cellcolor{poscolor!35.9}0.27 & \cellcolor{lightgrey!22.8}0.33 \\
\addlinespace[\rowspacing]
uses more emojis & \cellcolor{poscolor!13.9}0.10 & \cellcolor{poscolor!10.7}0.08 & \cellcolor{negcolor!29.4}-0.22 & \cellcolor{lightgrey!22.3}0.32 \\
\addlinespace[\rowspacing]
\rowcolor{altrow} ends with a follow-up question & \cellcolor{poscolor!24.0}0.18 & \cellcolor{negcolor!18.6}-0.14 & \cellcolor{poscolor!21.7}0.16 & \cellcolor{lightgrey!21.9}0.32 \\
\addlinespace[\rowspacing]
provides a numbered list format & \cellcolor{poscolor!52.2}0.39 & \cellcolor{poscolor!21.2}0.16 & \cellcolor{poscolor!61.3}0.46 & \cellcolor{lightgrey!20.5}0.30 \\
\addlinespace[\rowspacing]
\rowcolor{altrow} uses more personal pronouns (I, we, you) & \cellcolor{poscolor!35.8}0.27 & \cellcolor{negcolor!4.0}-0.03 & \cellcolor{poscolor!2.0}0.01 & \cellcolor{lightgrey!20.3}0.30 \\
\addlinespace[\rowspacing]
agrees more with the user & \cellcolor{poscolor!4.9}0.04 & \cellcolor{poscolor!42.2}0.31 & \cellcolor{poscolor!16.4}0.12 & \cellcolor{lightgrey!19.1}0.28 \\
\addlinespace[\rowspacing]
\rowcolor{altrow} provides conclusions without full reasoning & \cellcolor{negcolor!69.7}-0.52 & \cellcolor{negcolor!35.6}-0.27 & \cellcolor{negcolor!59.9}-0.45 & \cellcolor{lightgrey!17.6}0.26 \\
\addlinespace[\rowspacing]
provides more examples & \cellcolor{poscolor!66.0}0.49 & \cellcolor{poscolor!31.8}0.24 & \cellcolor{poscolor!52.5}0.39 & \cellcolor{lightgrey!17.5}0.25 \\
\addlinespace[\rowspacing]
\rowcolor{altrow} has more structured formatting & \cellcolor{poscolor!66.7}0.50 & \cellcolor{poscolor!32.7}0.24 & \cellcolor{poscolor!62.4}0.46 & \cellcolor{lightgrey!17.4}0.25 \\
\addlinespace[\rowspacing]
is more optimistic & \cellcolor{poscolor!53.7}0.40 & \cellcolor{poscolor!20.3}0.15 & \cellcolor{poscolor!36.2}0.27 & \cellcolor{lightgrey!17.1}0.25 \\
\addlinespace[\rowspacing]
\rowcolor{altrow} acknowledges own limitations or uncertainty more & \cellcolor{poscolor!8.1}0.06 & \cellcolor{negcolor!24.6}-0.18 & \cellcolor{negcolor!4.8}-0.04 & \cellcolor{lightgrey!16.9}0.24 \\
\addlinespace[\rowspacing]
is more concise & \cellcolor{negcolor!58.6}-0.44 & \cellcolor{negcolor!26.5}-0.20 & \cellcolor{negcolor!38.5}-0.29 & \cellcolor{lightgrey!16.6}0.24 \\
\addlinespace[\rowspacing]
\rowcolor{altrow} more strictly follows the requested output format & \cellcolor{poscolor!42.6}0.32 & \cellcolor{poscolor!36.5}0.27 & \cellcolor{poscolor!67.9}0.50 & \cellcolor{lightgrey!16.0}0.23 \\
\addlinespace[\rowspacing]
uses more formal language & \cellcolor{poscolor!30.3}0.23 & \cellcolor{poscolor!8.2}0.06 & \cellcolor{poscolor!39.4}0.29 & \cellcolor{lightgrey!15.9}0.23 \\
\addlinespace[\rowspacing]
\rowcolor{altrow} is more creative and original & \cellcolor{poscolor!71.2}0.53 & \cellcolor{poscolor!40.5}0.30 & \cellcolor{poscolor!50.1}0.37 & \cellcolor{lightgrey!15.7}0.23 \\
\addlinespace[\rowspacing]
is more verbose & \cellcolor{poscolor!47.3}0.35 & \cellcolor{poscolor!22.0}0.16 & \cellcolor{poscolor!36.4}0.27 & \cellcolor{lightgrey!12.9}0.19 \\
\addlinespace[\rowspacing]
\rowcolor{altrow} is more factually correct & \cellcolor{poscolor!68.1}0.51 & \cellcolor{poscolor!46.8}0.35 & \cellcolor{poscolor!68.1}0.51 & \cellcolor{lightgrey!10.9}0.16 \\
\addlinespace[\rowspacing]
makes more confident statements & \cellcolor{poscolor!38.4}0.29 & \cellcolor{poscolor!30.6}0.23 & \cellcolor{poscolor!29.5}0.22 & \cellcolor{lightgrey!4.6}0.07 \\
\end{tabular}
\end{minipage}

    \caption{\textbf{Comparison of investigated human feedback datasets in terms of \emph{Cohen's kappa} ($\kappa$).} As with \emph{strength}, positive $\kappa$ is shown in {\color{blue}blue} and negative $\kappa$ in {\color{red}red}. We observe why the strength metric is helpful: whilst some personality traits have high $\kappa$ here, their relevance to the overall dataset is minimal (as seen in \Cref{fig:app:results:cross_relevance}), for example \emph{inappropriate language}.}
    \label{fig:app:results:cross_kappa}
\end{figure}

\clearpage

\subsubsection{Chatbot Arena (general arena data)} 

\begin{figure}[ht]
    \centering
    \begin{minipage}[t]{0.48\textwidth}
\centering
\sffamily
\tablefontsize
\textbf{Five most encouraged personality traits}\\[0.5em]
\begin{tabular}{
    >{\raggedright\arraybackslash}m{0.7\linewidth}
    @{\hspace{10pt}} 
    >{\centering\arraybackslash}m{0.19\linewidth}
}

\textbf{Generating a response that...} & \textbf{Strength} \\
\toprule
\rowcolor{altrow} has more structured formatting & \cellcolor{poscolor!100.0}\makecell{0.17\\{\tiny(0.16, 0.19)}} \\
\addlinespace[\rowspacing]
is more verbose & \cellcolor{poscolor!91.3}\makecell{0.16\\{\tiny(0.14, 0.18)}} \\
\addlinespace[\rowspacing]
\rowcolor{altrow} is more factually correct & \cellcolor{poscolor!61.5}\makecell{0.11\\{\tiny(0.10, 0.12)}} \\
\addlinespace[\rowspacing]
provides more examples & \cellcolor{poscolor!57.1}\makecell{0.10\\{\tiny(0.09, 0.11)}} \\
\addlinespace[\rowspacing]
\rowcolor{altrow} makes more confident statements & \cellcolor{poscolor!56.4}\makecell{0.10\\{\tiny(0.08, 0.11)}} \\
\addlinespace[\rowspacing]
uses more bold and italics text & \cellcolor{poscolor!45.3}\makecell{0.08\\{\tiny(0.06, 0.09)}} \\
\addlinespace[\rowspacing]
\rowcolor{altrow} provides a numbered list format & \cellcolor{poscolor!44.4}\makecell{0.08\\{\tiny(0.06, 0.09)}} \\
\addlinespace[\rowspacing]
more strictly follows the requested output format & \cellcolor{poscolor!40.8}\makecell{0.07\\{\tiny(0.06, 0.08)}} \\
\addlinespace[\rowspacing]
\rowcolor{altrow} is more creative and original & \cellcolor{poscolor!39.2}\makecell{0.07\\{\tiny(0.06, 0.08)}} \\
\addlinespace[\rowspacing]
agrees more with the user & \cellcolor{poscolor!20.9}\makecell{0.04\\{\tiny(0.03, 0.04)}} \\
\end{tabular}
\end{minipage}
\hfill
\begin{minipage}[t]{0.48\textwidth}
\centering
\sffamily
\tablefontsize
\textbf{Five least encouraged personality traits}\\[0.5em]
\begin{tabular}{
    >{\raggedright\arraybackslash}m{0.7\linewidth}
    @{\hspace{10pt}} 
    >{\centering\arraybackslash}m{0.19\linewidth}
}

\textbf{Generating a response that...} & \textbf{Strength} \\
\toprule
\rowcolor{altrow} is more concise & \cellcolor{negcolor!100.0}\makecell{-0.09\\{\tiny(-0.11, -0.08)}} \\
\addlinespace[\rowspacing]
has a more avoidant tone & \cellcolor{negcolor!76.4}\makecell{-0.07\\{\tiny(-0.08, -0.06)}} \\
\addlinespace[\rowspacing]
\rowcolor{altrow} acknowledges own limitations or uncertainty more & \cellcolor{negcolor!53.8}\makecell{-0.05\\{\tiny(-0.06, -0.04)}} \\
\addlinespace[\rowspacing]
refuses to answer the question & \cellcolor{negcolor!52.4}\makecell{-0.05\\{\tiny(-0.05, -0.04)}} \\
\addlinespace[\rowspacing]
\rowcolor{altrow} ends with a follow-up question & \cellcolor{negcolor!29.4}\makecell{-0.03\\{\tiny(-0.04, -0.02)}} \\
\addlinespace[\rowspacing]
contains less harmful information & \cellcolor{negcolor!19.6}\makecell{-0.02\\{\tiny(-0.02, -0.01)}} \\
\addlinespace[\rowspacing]
\rowcolor{altrow} uses more personal pronouns (I, we, you) & \cellcolor{negcolor!15.5}\makecell{\transparent{0.3}-0.01\\\transparent{0.3}{\tiny(-0.03, -0.00)}} \\
\addlinespace[\rowspacing]
provides conclusions without full reasoning & \cellcolor{negcolor!6.7}\makecell{-0.01\\{\tiny(-0.01, -0.00)}} \\
\addlinespace[\rowspacing]
\rowcolor{altrow} includes more ethical considerations & \cellcolor{negcolor!4.1}\makecell{\transparent{0.3}-0.00\\\transparent{0.3}{\tiny(-0.01, 0.00)}} \\
\addlinespace[\rowspacing]
reinforces user's anger more & \cellcolor{poscolor!0.5}\makecell{\transparent{0.3}0.00\\\transparent{0.3}{\tiny(-0.00, 0.00)}} \\
\end{tabular}
\end{minipage}

    \caption{\textbf{Extended list of most} {\color{blue}(blue)} \textbf{and least} {\color{red}(red)} \textbf{encouraged personality traits in Chatbot Arena (general arena data).}}
    \label{fig:app:results:arena:top_bottom}
\end{figure}

\subsubsection{MultiPref}

\begin{figure}[ht]
    \centering
    \begin{minipage}[t]{0.48\textwidth}
\centering
\sffamily
\tablefontsize
\textbf{Five most encouraged personality traits}\\[0.5em]
\begin{tabular}{
    >{\raggedright\arraybackslash}m{0.7\linewidth}
    @{\hspace{10pt}} 
    >{\centering\arraybackslash}m{0.19\linewidth}
}

\textbf{Generating a response that...} & \textbf{Strength} \\
\toprule
\rowcolor{altrow} is more verbose & \cellcolor{poscolor!100.0}\makecell{0.34\\{\tiny(0.32, 0.35)}} \\
\addlinespace[\rowspacing]
has more structured formatting & \cellcolor{poscolor!69.5}\makecell{0.23\\{\tiny(0.22, 0.25)}} \\
\addlinespace[\rowspacing]
\rowcolor{altrow} provides more examples & \cellcolor{poscolor!66.3}\makecell{0.22\\{\tiny(0.21, 0.24)}} \\
\addlinespace[\rowspacing]
uses more bold and italics text & \cellcolor{poscolor!49.4}\makecell{0.17\\{\tiny(0.16, 0.18)}} \\
\addlinespace[\rowspacing]
\rowcolor{altrow} is more polite & \cellcolor{poscolor!41.1}\makecell{0.14\\{\tiny(0.13, 0.15)}} \\
\addlinespace[\rowspacing]
has a friendlier tone & \cellcolor{poscolor!37.1}\makecell{0.12\\{\tiny(0.11, 0.14)}} \\
\addlinespace[\rowspacing]
\rowcolor{altrow} uses more personal pronouns (I, we, you) & \cellcolor{poscolor!35.3}\makecell{0.12\\{\tiny(0.11, 0.13)}} \\
\addlinespace[\rowspacing]
provides a numbered list format & \cellcolor{poscolor!34.9}\makecell{0.12\\{\tiny(0.11, 0.13)}} \\
\addlinespace[\rowspacing]
\rowcolor{altrow} is more empathetic to the user & \cellcolor{poscolor!30.9}\makecell{0.10\\{\tiny(0.09, 0.11)}} \\
\addlinespace[\rowspacing]
more actively engages with the user & \cellcolor{poscolor!28.3}\makecell{0.10\\{\tiny(0.09, 0.10)}} \\
\end{tabular}
\end{minipage}
\hfill
\begin{minipage}[t]{0.48\textwidth}
\centering
\sffamily
\tablefontsize
\textbf{Five least encouraged personality traits}\\[0.5em]
\begin{tabular}{
    >{\raggedright\arraybackslash}m{0.7\linewidth}
    @{\hspace{10pt}} 
    >{\centering\arraybackslash}m{0.19\linewidth}
}

\textbf{Generating a response that...} & \textbf{Strength} \\
\toprule
\rowcolor{altrow} is more concise & \cellcolor{negcolor!100.0}\makecell{-0.29\\{\tiny(-0.30, -0.27)}} \\
\addlinespace[\rowspacing]
provides conclusions without full reasoning & \cellcolor{negcolor!2.8}\makecell{-0.01\\{\tiny(-0.01, -0.01)}} \\
\addlinespace[\rowspacing]
\rowcolor{altrow} agrees with user even if factually incorrect & \cellcolor{negcolor!2.4}\makecell{-0.01\\{\tiny(-0.01, -0.00)}} \\
\addlinespace[\rowspacing]
is more offensive & \cellcolor{negcolor!1.8}\makecell{-0.01\\{\tiny(-0.01, -0.00)}} \\
\addlinespace[\rowspacing]
\rowcolor{altrow} includes inappropriate language & \cellcolor{negcolor!1.0}\makecell{-0.00\\{\tiny(-0.00, -0.00)}} \\
\addlinespace[\rowspacing]
suggests illegal activities & \cellcolor{negcolor!1.0}\makecell{-0.00\\{\tiny(-0.00, -0.00)}} \\
\addlinespace[\rowspacing]
\rowcolor{altrow} has a more avoidant tone & \cellcolor{negcolor!0.6}\makecell{\transparent{0.3}-0.00\\\transparent{0.3}{\tiny(-0.01, 0.00)}} \\
\addlinespace[\rowspacing]
reinforces user's anger more & \cellcolor{poscolor!0}\makecell{\transparent{0.3}0.00\\\transparent{0.3}{\tiny(-0.00, 0.00)}} \\
\addlinespace[\rowspacing]
\rowcolor{altrow} reinforces user's beliefs more & \cellcolor{poscolor!0}\makecell{\transparent{0.3}0.00\\\transparent{0.3}{\tiny(-0.00, 0.00)}} \\
\addlinespace[\rowspacing]
uses more emojis & \cellcolor{poscolor!0.2}\makecell{\transparent{0.3}0.00\\\transparent{0.3}{\tiny(-0.00, 0.00)}} \\
\end{tabular}
\end{minipage}

    \caption{\textbf{Extended list of most} {\color{blue}(blue)} \textbf{and least} {\color{red}(red)} \textbf{encouraged personality traits in MultiPref.}}
    \label{fig:app:results:multipref:top_bottom}
\end{figure}

\clearpage\subsubsection{PRISM}
\begin{figure}[ht]
    \centering
    \begin{minipage}[t]{0.48\textwidth}
\centering
\sffamily
\tablefontsize
\textbf{Five most encouraged personality traits}\\[0.5em]
\begin{tabular}{
    >{\raggedright\arraybackslash}m{0.7\linewidth}
    @{\hspace{10pt}} 
    >{\centering\arraybackslash}m{0.19\linewidth}
}

\textbf{Generating a response that...} & \textbf{Strength} \\
\toprule
\rowcolor{altrow} is more verbose & \cellcolor{poscolor!100.0}\makecell{0.26\\{\tiny(0.23, 0.28)}} \\
\addlinespace[\rowspacing]
uses more formal language & \cellcolor{poscolor!67.2}\makecell{0.17\\{\tiny(0.16, 0.19)}} \\
\addlinespace[\rowspacing]
\rowcolor{altrow} is more polite & \cellcolor{poscolor!58.9}\makecell{0.15\\{\tiny(0.13, 0.17)}} \\
\addlinespace[\rowspacing]
is more factually correct & \cellcolor{poscolor!51.2}\makecell{0.13\\{\tiny(0.12, 0.14)}} \\
\addlinespace[\rowspacing]
\rowcolor{altrow} includes more ethical considerations & \cellcolor{poscolor!50.2}\makecell{0.13\\{\tiny(0.11, 0.14)}} \\
\addlinespace[\rowspacing]
provides more examples & \cellcolor{poscolor!43.2}\makecell{0.11\\{\tiny(0.10, 0.12)}} \\
\addlinespace[\rowspacing]
\rowcolor{altrow} is more empathetic to the user & \cellcolor{poscolor!38.4}\makecell{0.10\\{\tiny(0.08, 0.11)}} \\
\addlinespace[\rowspacing]
makes more confident statements & \cellcolor{poscolor!37.5}\makecell{0.10\\{\tiny(0.08, 0.11)}} \\
\addlinespace[\rowspacing]
\rowcolor{altrow} has a friendlier tone & \cellcolor{poscolor!36.2}\makecell{0.09\\{\tiny(0.08, 0.11)}} \\
\addlinespace[\rowspacing]
has more structured formatting & \cellcolor{poscolor!35.5}\makecell{0.09\\{\tiny(0.08, 0.10)}} \\
\end{tabular}
\end{minipage}
\hfill
\begin{minipage}[t]{0.48\textwidth}
\centering
\sffamily
\tablefontsize
\textbf{Five least encouraged personality traits}\\[0.5em]
\begin{tabular}{
    >{\raggedright\arraybackslash}m{0.7\linewidth}
    @{\hspace{10pt}} 
    >{\centering\arraybackslash}m{0.19\linewidth}
}

\textbf{Generating a response that...} & \textbf{Strength} \\
\toprule
\rowcolor{altrow} is more concise & \cellcolor{negcolor!100.0}\makecell{-0.23\\{\tiny(-0.25, -0.21)}} \\
\addlinespace[\rowspacing]
has a more avoidant tone & \cellcolor{negcolor!24.4}\makecell{-0.06\\{\tiny(-0.07, -0.04)}} \\
\addlinespace[\rowspacing]
\rowcolor{altrow} uses more casual language & \cellcolor{negcolor!23.1}\makecell{-0.05\\{\tiny(-0.06, -0.04)}} \\
\addlinespace[\rowspacing]
refuses to answer the question & \cellcolor{negcolor!22.9}\makecell{-0.05\\{\tiny(-0.06, -0.04)}} \\
\addlinespace[\rowspacing]
\rowcolor{altrow} provides conclusions without full reasoning & \cellcolor{negcolor!8.8}\makecell{-0.02\\{\tiny(-0.02, -0.02)}} \\
\addlinespace[\rowspacing]
acknowledges own limitations or uncertainty more & \cellcolor{negcolor!5.8}\makecell{\transparent{0.3}-0.01\\\transparent{0.3}{\tiny(-0.03, 0.00)}} \\
\addlinespace[\rowspacing]
\rowcolor{altrow} is more offensive & \cellcolor{negcolor!3.5}\makecell{-0.01\\{\tiny(-0.01, -0.01)}} \\
\addlinespace[\rowspacing]
actively engages the reader with rhetorical questions & \cellcolor{negcolor!2.4}\makecell{\transparent{0.3}-0.01\\\transparent{0.3}{\tiny(-0.01, 0.00)}} \\
\addlinespace[\rowspacing]
\rowcolor{altrow} agrees with user even if factually incorrect & \cellcolor{negcolor!1.7}\makecell{\transparent{0.3}-0.00\\\transparent{0.3}{\tiny(-0.01, 0.00)}} \\
\addlinespace[\rowspacing]
includes inappropriate language & \cellcolor{negcolor!1.5}\makecell{-0.00\\{\tiny(-0.01, -0.00)}} \\
\end{tabular}
\end{minipage}

    \caption{\textbf{List of most} {\color{blue}(blue)} \textbf{and least} {\color{red}(red)} \textbf{encouraged personality traits in PRISM.}}
    \label{fig:app:results:prism:top_bottom}
\end{figure}

\clearpage

\subsection{Extended model results}

\subsubsection{General model comparison over time}
\label{app:model_comparison_time}

\Cref{fig:app:results:claude_strength,fig:app:results:gpt_strength,fig:app:results:gemini_strength} provide extended \emph{strength} metrics for each model family. These figures provide a more comprehensive view of the results shared in \Cref{sec:results:crossmodels}. Considering the full set of tested personality traits, we can observe differences in the \emph{stability of personality} across model families. For each family and trait, we compute the \emph{max diff}: the maximum difference in terms of strength value for any two models in that family on a specific trait (included in \Cref{fig:app:results:claude_strength,fig:app:results:gpt_strength,fig:app:results:gemini_strength}). Then, averaging the \emph{max diff} over the full set of 40 test personality traits, we observe that the ChatGPT family has a \emph{mean max diff} of $0.26$, whereas both Gemini and Anthropic have a value of $0.13$. This quantitative observation appears to match anecdotal evidence \citep{openai2025ExpandingWhatWe,openaicommunity2025ExcessiveBulletPoint} that highlight notable changes in ChatGPT's personality over time. Note that for this analysis we only considered the last three ChatGPT variants included in the data, since the fourth (GPT-4o) is older than any other model considered. Nevertheless, with the remaining three models, we obtain this result even though the release dates of those ChatGPT models are closer together than for either of the other families. Finally, note that we consider \emph{default} personalities of the models here. Some services provide user configuration parameters that change this personality, though we consider it likely that the default personality remains widely-used by many users.

\begin{figure}[ht]
    \centering
    \begin{minipage}[t]{1\textwidth}
\centering
\sffamily
\tablefontsize
\begin{tabular}{
    >{\raggedright\arraybackslash}m{0.23\linewidth}
    @{\hspace{13.5pt}} 
    >{\centering\arraybackslash}m{0.10363636363636365\linewidth}
    @{\hspace{13.5pt}} 
    >{\centering\arraybackslash}m{0.10363636363636365\linewidth}
    @{\hspace{13.5pt}} 
    >{\centering\arraybackslash}m{0.10363636363636365\linewidth}
    @{\hspace{13.5pt}} 
    >{\centering\arraybackslash}m{0.10363636363636365\linewidth}
    @{\hspace{13.5pt}} 
    >{\centering\arraybackslash}m{0.10363636363636365\linewidth}
}

\textbf{Generating a response that...} & \textbf{Anthropic \textit{Claude-3.7-Sonnet}} & \textbf{Anthropic \textit{Claude-Sonnet-4}} & \textbf{Anthropic \textit{Claude-Sonnet-4.5}} & \textbf{Anthropic \textit{Claude-Opus-4.7}} & \textbf{Max diff} \\
\toprule
\rowcolor{altrow} uses more bold and italics text & \cellcolor{negcolor!100.0}\makecell{-0.42\\{\tiny(-0.48, -0.35)}} & \cellcolor{poscolor!31.8}\makecell{\transparent{0.3}0.11\\\transparent{0.3}{\tiny(0.03, 0.18)}} & \cellcolor{poscolor!93.5}\makecell{0.32\\{\tiny(0.24, 0.39)}} & \cellcolor{poscolor!100.0}\makecell{0.34\\{\tiny(0.27, 0.41)}} & \cellcolor{lightgrey!100.0}0.76 \\
\addlinespace[\rowspacing]
has more structured formatting & \cellcolor{negcolor!50.0}\makecell{-0.21\\{\tiny(-0.29, -0.14)}} & \cellcolor{poscolor!14.1}\makecell{\transparent{0.3}0.05\\\transparent{0.3}{\tiny(-0.03, 0.13)}} & \cellcolor{poscolor!56.8}\makecell{0.19\\{\tiny(0.11, 0.27)}} & \cellcolor{poscolor!89.3}\makecell{0.30\\{\tiny(0.23, 0.38)}} & \cellcolor{lightgrey!67.5}0.51 \\
\addlinespace[\rowspacing]
\rowcolor{altrow} more actively engages with the user & \cellcolor{negcolor!19.2}\makecell{\transparent{0.3}-0.08\\\transparent{0.3}{\tiny(-0.13, -0.03)}} & \cellcolor{poscolor!35.4}\makecell{0.12\\{\tiny(0.07, 0.17)}} & \cellcolor{poscolor!24.9}\makecell{\transparent{0.3}0.08\\\transparent{0.3}{\tiny(0.03, 0.13)}} & \cellcolor{poscolor!80.5}\makecell{0.27\\{\tiny(0.22, 0.33)}} & \cellcolor{lightgrey!46.5}0.35 \\
\addlinespace[\rowspacing]
is more verbose & \cellcolor{negcolor!62.0}\makecell{-0.26\\{\tiny(-0.34, -0.18)}} & \cellcolor{poscolor!25.9}\makecell{\transparent{0.3}0.09\\\transparent{0.3}{\tiny(0.00, 0.17)}} & \cellcolor{negcolor!9.5}\makecell{\transparent{0.3}-0.04\\\transparent{0.3}{\tiny(-0.13, 0.05)}} & \cellcolor{poscolor!26.6}\makecell{\transparent{0.3}0.09\\\transparent{0.3}{\tiny(0.00, 0.17)}} & \cellcolor{lightgrey!46.2}0.35 \\
\addlinespace[\rowspacing]
\rowcolor{altrow} provides more examples & \cellcolor{negcolor!30.3}\makecell{-0.13\\{\tiny(-0.18, -0.07)}} & \cellcolor{poscolor!40.7}\makecell{0.14\\{\tiny(0.08, 0.20)}} & \cellcolor{poscolor!40.8}\makecell{0.14\\{\tiny(0.08, 0.20)}} & \cellcolor{poscolor!62.7}\makecell{0.21\\{\tiny(0.15, 0.27)}} & \cellcolor{lightgrey!44.7}0.34 \\
\addlinespace[\rowspacing]
ends with a follow-up question & \cellcolor{negcolor!7.7}\makecell{\transparent{0.3}-0.03\\\transparent{0.3}{\tiny(-0.08, 0.01)}} & \cellcolor{poscolor!17.7}\makecell{\transparent{0.3}0.06\\\transparent{0.3}{\tiny(0.01, 0.10)}} & \cellcolor{poscolor!17.8}\makecell{\transparent{0.3}0.06\\\transparent{0.3}{\tiny(0.01, 0.11)}} & \cellcolor{poscolor!69.2}\makecell{0.23\\{\tiny(0.18, 0.29)}} & \cellcolor{lightgrey!35.1}0.27 \\
\addlinespace[\rowspacing]
\rowcolor{altrow} is more concise & \cellcolor{poscolor!50.2}\makecell{0.17\\{\tiny(0.10, 0.24)}} & \cellcolor{negcolor!16.1}\makecell{\transparent{0.3}-0.07\\\transparent{0.3}{\tiny(-0.13, -0.00)}} & \cellcolor{poscolor!4.7}\makecell{\transparent{0.3}0.02\\\transparent{0.3}{\tiny(-0.05, 0.08)}} & \cellcolor{negcolor!8.6}\makecell{\transparent{0.3}-0.04\\\transparent{0.3}{\tiny(-0.09, 0.02)}} & \cellcolor{lightgrey!31.3}0.24 \\
\addlinespace[\rowspacing]
makes more confident statements & \cellcolor{poscolor!0.6}\makecell{\transparent{0.3}0.00\\\transparent{0.3}{\tiny(-0.04, 0.05)}} & \cellcolor{poscolor!20.6}\makecell{\transparent{0.3}0.07\\\transparent{0.3}{\tiny(0.03, 0.11)}} & \cellcolor{poscolor!40.2}\makecell{0.14\\{\tiny(0.09, 0.19)}} & \cellcolor{poscolor!59.2}\makecell{0.20\\{\tiny(0.15, 0.25)}} & \cellcolor{lightgrey!26.1}0.20 \\
\addlinespace[\rowspacing]
\rowcolor{altrow} is more factually correct & \cellcolor{poscolor!0}\makecell{\transparent{0.3}0.00\\\transparent{0.3}{\tiny(-0.03, 0.03)}} & \cellcolor{poscolor!11.2}\makecell{\transparent{0.3}0.04\\\transparent{0.3}{\tiny(0.01, 0.07)}} & \cellcolor{poscolor!16.6}\makecell{\transparent{0.3}0.06\\\transparent{0.3}{\tiny(0.02, 0.09)}} & \cellcolor{poscolor!54.4}\makecell{0.18\\{\tiny(0.15, 0.22)}} & \cellcolor{lightgrey!24.3}0.18 \\
\addlinespace[\rowspacing]
uses more personal pronouns (I, we, you) & \cellcolor{negcolor!3.4}\makecell{\transparent{0.3}-0.01\\\transparent{0.3}{\tiny(-0.07, 0.04)}} & \cellcolor{poscolor!47.7}\makecell{0.16\\{\tiny(0.11, 0.22)}} & \cellcolor{poscolor!16.6}\makecell{\transparent{0.3}0.06\\\transparent{0.3}{\tiny(-0.00, 0.11)}} & \cellcolor{poscolor!32.5}\makecell{0.11\\{\tiny(0.06, 0.16)}} & \cellcolor{lightgrey!23.1}0.18 \\
\addlinespace[\rowspacing]
\rowcolor{altrow} has a friendlier tone & \cellcolor{negcolor!40.4}\makecell{-0.17\\{\tiny(-0.21, -0.13)}} & \cellcolor{negcolor!3.8}\makecell{\transparent{0.3}-0.02\\\transparent{0.3}{\tiny(-0.06, 0.03)}} & \cellcolor{negcolor!20.9}\makecell{-0.09\\{\tiny(-0.13, -0.04)}} & \cellcolor{negcolor!2.9}\makecell{\transparent{0.3}-0.01\\\transparent{0.3}{\tiny(-0.06, 0.04)}} & \cellcolor{lightgrey!20.8}0.16 \\
\addlinespace[\rowspacing]
uses more formal language & \cellcolor{negcolor!22.6}\makecell{-0.10\\{\tiny(-0.15, -0.04)}} & \cellcolor{negcolor!34.1}\makecell{-0.14\\{\tiny(-0.19, -0.09)}} & \cellcolor{negcolor!60.0}\makecell{-0.25\\{\tiny(-0.30, -0.20)}} & \cellcolor{negcolor!37.6}\makecell{-0.16\\{\tiny(-0.21, -0.10)}} & \cellcolor{lightgrey!20.7}0.16 \\
\addlinespace[\rowspacing]
\rowcolor{altrow} uses more casual language & \cellcolor{negcolor!6.7}\makecell{\transparent{0.3}-0.03\\\transparent{0.3}{\tiny(-0.05, -0.01)}} & \cellcolor{poscolor!13.0}\makecell{\transparent{0.3}0.04\\\transparent{0.3}{\tiny(0.02, 0.07)}} & \cellcolor{poscolor!20.1}\makecell{0.07\\{\tiny(0.04, 0.09)}} & \cellcolor{poscolor!35.5}\makecell{0.12\\{\tiny(0.09, 0.15)}} & \cellcolor{lightgrey!19.6}0.15 \\
\addlinespace[\rowspacing]
is more creative and original & \cellcolor{poscolor!14.9}\makecell{\transparent{0.3}0.05\\\transparent{0.3}{\tiny(0.02, 0.08)}} & \cellcolor{poscolor!34.2}\makecell{0.12\\{\tiny(0.08, 0.15)}} & \cellcolor{poscolor!41.4}\makecell{0.14\\{\tiny(0.10, 0.18)}} & \cellcolor{poscolor!58.6}\makecell{0.20\\{\tiny(0.16, 0.24)}} & \cellcolor{lightgrey!19.5}0.15 \\
\addlinespace[\rowspacing]
\rowcolor{altrow} uses a more enthusiastic tone & \cellcolor{negcolor!18.8}\makecell{-0.08\\{\tiny(-0.11, -0.05)}} & \cellcolor{poscolor!7.7}\makecell{\transparent{0.3}0.03\\\transparent{0.3}{\tiny(-0.01, 0.06)}} & \cellcolor{poscolor!6.5}\makecell{\transparent{0.3}0.02\\\transparent{0.3}{\tiny(-0.01, 0.06)}} & \cellcolor{poscolor!16.6}\makecell{\transparent{0.3}0.06\\\transparent{0.3}{\tiny(0.02, 0.09)}} & \cellcolor{lightgrey!17.8}0.14 \\
\addlinespace[\rowspacing]
is more polite & \cellcolor{negcolor!51.4}\makecell{-0.22\\{\tiny(-0.26, -0.17)}} & \cellcolor{negcolor!24.2}\makecell{-0.10\\{\tiny(-0.15, -0.06)}} & \cellcolor{negcolor!54.7}\makecell{-0.23\\{\tiny(-0.27, -0.18)}} & \cellcolor{negcolor!36.2}\makecell{-0.15\\{\tiny(-0.20, -0.10)}} & \cellcolor{lightgrey!16.9}0.13 \\
\addlinespace[\rowspacing]
\rowcolor{altrow} is more empathetic to the user & \cellcolor{negcolor!20.2}\makecell{-0.09\\{\tiny(-0.12, -0.05)}} & \cellcolor{poscolor!11.8}\makecell{\transparent{0.3}0.04\\\transparent{0.3}{\tiny(0.01, 0.07)}} & \cellcolor{negcolor!7.1}\makecell{\transparent{0.3}-0.03\\\transparent{0.3}{\tiny(-0.07, 0.00)}} & \cellcolor{poscolor!5.9}\makecell{\transparent{0.3}0.02\\\transparent{0.3}{\tiny(-0.02, 0.06)}} & \cellcolor{lightgrey!16.4}0.13 \\
\addlinespace[\rowspacing]
actively engages the reader with rhetorical questions & \cellcolor{negcolor!1.0}\makecell{\transparent{0.3}-0.00\\\transparent{0.3}{\tiny(-0.02, 0.02)}} & \cellcolor{poscolor!21.8}\makecell{0.07\\{\tiny(0.05, 0.10)}} & \cellcolor{poscolor!32.0}\makecell{0.11\\{\tiny(0.08, 0.14)}} & \cellcolor{poscolor!29.6}\makecell{0.10\\{\tiny(0.07, 0.13)}} & \cellcolor{lightgrey!14.8}0.11 \\
\addlinespace[\rowspacing]
\rowcolor{altrow} includes more references to other sources & \cellcolor{negcolor!3.4}\makecell{\transparent{0.3}-0.01\\\transparent{0.3}{\tiny(-0.04, 0.01)}} & \cellcolor{poscolor!0.6}\makecell{\transparent{0.3}0.00\\\transparent{0.3}{\tiny(-0.02, 0.02)}} & \cellcolor{negcolor!0.5}\makecell{\transparent{0.3}-0.00\\\transparent{0.3}{\tiny(-0.02, 0.02)}} & \cellcolor{poscolor!26.0}\makecell{0.09\\{\tiny(0.06, 0.12)}} & \cellcolor{lightgrey!13.5}0.10 \\
\addlinespace[\rowspacing]
more strictly follows the requested output format & \cellcolor{negcolor!9.1}\makecell{\transparent{0.3}-0.04\\\transparent{0.3}{\tiny(-0.08, 0.00)}} & \cellcolor{negcolor!12.8}\makecell{\transparent{0.3}-0.05\\\transparent{0.3}{\tiny(-0.10, -0.01)}} & \cellcolor{poscolor!1.2}\makecell{\transparent{0.3}0.00\\\transparent{0.3}{\tiny(-0.04, 0.05)}} & \cellcolor{poscolor!10.1}\makecell{\transparent{0.3}0.03\\\transparent{0.3}{\tiny(-0.01, 0.08)}} & \cellcolor{lightgrey!11.6}0.09 \\
\addlinespace[\rowspacing]
\rowcolor{altrow} uses more emojis & \cellcolor{negcolor!6.7}\makecell{\transparent{0.3}-0.03\\\transparent{0.3}{\tiny(-0.05, -0.01)}} & \cellcolor{poscolor!3.5}\makecell{\transparent{0.3}0.01\\\transparent{0.3}{\tiny(-0.01, 0.04)}} & \cellcolor{poscolor!13.6}\makecell{\transparent{0.3}0.05\\\transparent{0.3}{\tiny(0.02, 0.07)}} & \cellcolor{poscolor!16.6}\makecell{0.06\\{\tiny(0.03, 0.08)}} & \cellcolor{lightgrey!11.1}0.08 \\
\addlinespace[\rowspacing]
provides a numbered list format & \cellcolor{negcolor!69.7}\makecell{-0.29\\{\tiny(-0.35, -0.23)}} & \cellcolor{negcolor!52.1}\makecell{-0.22\\{\tiny(-0.28, -0.15)}} & \cellcolor{negcolor!52.4}\makecell{-0.22\\{\tiny(-0.28, -0.16)}} & \cellcolor{negcolor!63.3}\makecell{-0.27\\{\tiny(-0.32, -0.21)}} & \cellcolor{lightgrey!9.7}0.07 \\
\addlinespace[\rowspacing]
\rowcolor{altrow} expresses more emotion & \cellcolor{negcolor!8.2}\makecell{-0.03\\{\tiny(-0.05, -0.02)}} & \cellcolor{poscolor!4.1}\makecell{\transparent{0.3}0.01\\\transparent{0.3}{\tiny(-0.01, 0.04)}} & \cellcolor{poscolor!7.1}\makecell{\transparent{0.3}0.02\\\transparent{0.3}{\tiny(0.00, 0.05)}} & \cellcolor{poscolor!10.7}\makecell{\transparent{0.3}0.04\\\transparent{0.3}{\tiny(0.01, 0.06)}} & \cellcolor{lightgrey!9.3}0.07 \\
\addlinespace[\rowspacing]
uses more mathematical symbols and notation & \cellcolor{negcolor!18.3}\makecell{-0.08\\{\tiny(-0.10, -0.05)}} & \cellcolor{negcolor!18.5}\makecell{-0.08\\{\tiny(-0.11, -0.05)}} & \cellcolor{negcolor!15.2}\makecell{-0.06\\{\tiny(-0.09, -0.04)}} & \cellcolor{negcolor!4.3}\makecell{\transparent{0.3}-0.02\\\transparent{0.3}{\tiny(-0.05, 0.01)}} & \cellcolor{lightgrey!7.9}0.06 \\
\addlinespace[\rowspacing]
\rowcolor{altrow} compliments the user's question or prompt & \cellcolor{negcolor!17.3}\makecell{-0.07\\{\tiny(-0.10, -0.05)}} & \cellcolor{negcolor!4.3}\makecell{\transparent{0.3}-0.02\\\transparent{0.3}{\tiny(-0.05, 0.01)}} & \cellcolor{negcolor!11.4}\makecell{\transparent{0.3}-0.05\\\transparent{0.3}{\tiny(-0.08, -0.01)}} & \cellcolor{negcolor!3.3}\makecell{\transparent{0.3}-0.01\\\transparent{0.3}{\tiny(-0.05, 0.02)}} & \cellcolor{lightgrey!7.7}0.06 \\
\addlinespace[\rowspacing]
includes more ethical considerations & \cellcolor{negcolor!8.7}\makecell{\transparent{0.3}-0.04\\\transparent{0.3}{\tiny(-0.06, -0.01)}} & \cellcolor{poscolor!0.6}\makecell{\transparent{0.3}0.00\\\transparent{0.3}{\tiny(-0.03, 0.03)}} & \cellcolor{poscolor!0.6}\makecell{\transparent{0.3}0.00\\\transparent{0.3}{\tiny(-0.03, 0.03)}} & \cellcolor{poscolor!3.6}\makecell{\transparent{0.3}0.01\\\transparent{0.3}{\tiny(-0.02, 0.04)}} & \cellcolor{lightgrey!6.4}0.05 \\
\addlinespace[\rowspacing]
\rowcolor{altrow} uses more humour & \cellcolor{poscolor!3.6}\makecell{\transparent{0.3}0.01\\\transparent{0.3}{\tiny(-0.00, 0.03)}} & \cellcolor{poscolor!10.0}\makecell{0.03\\{\tiny(0.02, 0.05)}} & \cellcolor{poscolor!10.7}\makecell{0.04\\{\tiny(0.02, 0.06)}} & \cellcolor{poscolor!17.2}\makecell{0.06\\{\tiny(0.04, 0.08)}} & \cellcolor{lightgrey!6.1}0.05 \\
\addlinespace[\rowspacing]
acknowledges own limitations or uncertainty more & \cellcolor{negcolor!4.8}\makecell{\transparent{0.3}-0.02\\\transparent{0.3}{\tiny(-0.05, 0.01)}} & \cellcolor{poscolor!7.1}\makecell{\transparent{0.3}0.02\\\transparent{0.3}{\tiny(-0.01, 0.06)}} & \cellcolor{poscolor!4.1}\makecell{\transparent{0.3}0.01\\\transparent{0.3}{\tiny(-0.02, 0.05)}} & \cellcolor{poscolor!5.9}\makecell{\transparent{0.3}0.02\\\transparent{0.3}{\tiny(-0.02, 0.06)}} & \cellcolor{lightgrey!5.8}0.04 \\
\addlinespace[\rowspacing]
\rowcolor{altrow} agrees more with the user & \cellcolor{negcolor!0.5}\makecell{\transparent{0.3}-0.00\\\transparent{0.3}{\tiny(-0.02, 0.02)}} & \cellcolor{poscolor!1.2}\makecell{\transparent{0.3}0.00\\\transparent{0.3}{\tiny(-0.01, 0.02)}} & \cellcolor{negcolor!4.8}\makecell{\transparent{0.3}-0.02\\\transparent{0.3}{\tiny(-0.04, -0.00)}} & \cellcolor{negcolor!2.9}\makecell{\transparent{0.3}-0.01\\\transparent{0.3}{\tiny(-0.03, 0.01)}} & \cellcolor{lightgrey!3.2}0.02 \\
\addlinespace[\rowspacing]
is more optimistic & \cellcolor{negcolor!3.4}\makecell{\transparent{0.3}-0.01\\\transparent{0.3}{\tiny(-0.03, 0.01)}} & \cellcolor{poscolor!1.2}\makecell{\transparent{0.3}0.00\\\transparent{0.3}{\tiny(-0.01, 0.02)}} & \cellcolor{negcolor!1.9}\makecell{\transparent{0.3}-0.01\\\transparent{0.3}{\tiny(-0.03, 0.01)}} & \cellcolor{negcolor!1.9}\makecell{\transparent{0.3}-0.01\\\transparent{0.3}{\tiny(-0.03, 0.01)}} & \cellcolor{lightgrey!2.4}0.02 \\
\end{tabular}
\end{minipage}

    \caption{\textbf{Top 30 traits for Anthropic Claude models in terms of change in \emph{strength} over time.} Sorted by maximum difference. \strengthexpl{} All measurements are relative to GPT-4o. Mean maximum difference is 0.13.}
    \label{fig:app:results:claude_strength}
\end{figure}

\begin{figure}[ht]
    \centering
    \begin{minipage}[t]{1\textwidth}
\centering
\sffamily
\tablefontsize
\begin{tabular}{
    >{\raggedright\arraybackslash}m{0.23\linewidth}
    @{\hspace{13.5pt}} 
    >{\centering\arraybackslash}m{0.12954545454545455\linewidth}
    @{\hspace{13.5pt}} 
    >{\centering\arraybackslash}m{0.12954545454545455\linewidth}
    @{\hspace{13.5pt}} 
    >{\centering\arraybackslash}m{0.12954545454545455\linewidth}
    @{\hspace{13.5pt}} 
    >{\centering\arraybackslash}m{0.12954545454545455\linewidth}
}

\textbf{Generating a response that...} & \textbf{OpenAI \textit{GPT-5-Chat}} & \textbf{OpenAI \textit{GPT-5.1-Chat}} & \textbf{OpenAI \textit{GPT-5.3-Chat}} & \textbf{Max diff} \\
\toprule
\rowcolor{altrow} uses more bold and italics text & \cellcolor{poscolor!87.4}\makecell{0.69\\{\tiny(0.63, 0.74)}} & \cellcolor{negcolor!58.7}\makecell{-0.50\\{\tiny(-0.56, -0.43)}} & \cellcolor{negcolor!50.7}\makecell{-0.43\\{\tiny(-0.49, -0.36)}} & \cellcolor{lightgrey!100.0}1.19 \\
\addlinespace[\rowspacing]
is more verbose & \cellcolor{poscolor!29.3}\makecell{0.23\\{\tiny(0.14, 0.31)}} & \cellcolor{negcolor!100.0}\makecell{-0.85\\{\tiny(-0.89, -0.80)}} & \cellcolor{negcolor!68.6}\makecell{-0.58\\{\tiny(-0.65, -0.51)}} & \cellcolor{lightgrey!90.8}1.08 \\
\addlinespace[\rowspacing]
\rowcolor{altrow} has more structured formatting & \cellcolor{poscolor!53.3}\makecell{0.42\\{\tiny(0.35, 0.49)}} & \cellcolor{negcolor!74.1}\makecell{-0.63\\{\tiny(-0.68, -0.57)}} & \cellcolor{negcolor!55.4}\makecell{-0.47\\{\tiny(-0.53, -0.40)}} & \cellcolor{lightgrey!88.2}1.05 \\
\addlinespace[\rowspacing]
is more concise & \cellcolor{negcolor!19.8}\makecell{-0.17\\{\tiny(-0.23, -0.11)}} & \cellcolor{poscolor!100.0}\makecell{0.79\\{\tiny(0.74, 0.83)}} & \cellcolor{poscolor!61.4}\makecell{0.49\\{\tiny(0.43, 0.54)}} & \cellcolor{lightgrey!80.7}0.96 \\
\addlinespace[\rowspacing]
\rowcolor{altrow} more actively engages with the user & \cellcolor{poscolor!67.2}\makecell{0.53\\{\tiny(0.48, 0.58)}} & \cellcolor{negcolor!1.2}\makecell{\transparent{0.3}-0.01\\\transparent{0.3}{\tiny(-0.07, 0.05)}} & \cellcolor{poscolor!30.3}\makecell{0.24\\{\tiny(0.18, 0.29)}} & \cellcolor{lightgrey!45.5}0.54 \\
\addlinespace[\rowspacing]
is more polite & \cellcolor{poscolor!23.2}\makecell{0.18\\{\tiny(0.13, 0.23)}} & \cellcolor{negcolor!33.5}\makecell{-0.28\\{\tiny(-0.33, -0.24)}} & \cellcolor{negcolor!26.9}\makecell{-0.23\\{\tiny(-0.27, -0.18)}} & \cellcolor{lightgrey!39.3}0.47 \\
\addlinespace[\rowspacing]
\rowcolor{altrow} provides more examples & \cellcolor{poscolor!17.4}\makecell{0.14\\{\tiny(0.08, 0.19)}} & \cellcolor{negcolor!36.8}\makecell{-0.31\\{\tiny(-0.36, -0.26)}} & \cellcolor{negcolor!16.3}\makecell{-0.14\\{\tiny(-0.19, -0.08)}} & \cellcolor{lightgrey!37.8}0.45 \\
\addlinespace[\rowspacing]
has a friendlier tone & \cellcolor{poscolor!33.1}\makecell{0.26\\{\tiny(0.21, 0.31)}} & \cellcolor{negcolor!21.5}\makecell{-0.18\\{\tiny(-0.23, -0.14)}} & \cellcolor{poscolor!17.7}\makecell{0.14\\{\tiny(0.08, 0.19)}} & \cellcolor{lightgrey!37.3}0.44 \\
\addlinespace[\rowspacing]
\rowcolor{altrow} ends with a follow-up question & \cellcolor{poscolor!58.6}\makecell{0.46\\{\tiny(0.41, 0.51)}} & \cellcolor{poscolor!8.3}\makecell{\transparent{0.3}0.07\\\transparent{0.3}{\tiny(0.02, 0.11)}} & \cellcolor{poscolor!20.7}\makecell{0.16\\{\tiny(0.11, 0.22)}} & \cellcolor{lightgrey!33.4}0.40 \\
\addlinespace[\rowspacing]
uses more formal language & \cellcolor{negcolor!6.8}\makecell{\transparent{0.3}-0.06\\\transparent{0.3}{\tiny(-0.11, -0.00)}} & \cellcolor{negcolor!46.2}\makecell{-0.39\\{\tiny(-0.44, -0.34)}} & \cellcolor{negcolor!50.2}\makecell{-0.43\\{\tiny(-0.47, -0.37)}} & \cellcolor{lightgrey!30.9}0.37 \\
\addlinespace[\rowspacing]
\rowcolor{altrow} uses more personal pronouns (I, we, you) & \cellcolor{poscolor!25.3}\makecell{0.20\\{\tiny(0.14, 0.26)}} & \cellcolor{negcolor!18.9}\makecell{-0.16\\{\tiny(-0.21, -0.11)}} & \cellcolor{poscolor!18.4}\makecell{0.15\\{\tiny(0.09, 0.20)}} & \cellcolor{lightgrey!30.3}0.36 \\
\addlinespace[\rowspacing]
provides a numbered list format & \cellcolor{negcolor!10.8}\makecell{\transparent{0.3}-0.09\\\transparent{0.3}{\tiny(-0.15, -0.04)}} & \cellcolor{negcolor!48.6}\makecell{-0.41\\{\tiny(-0.47, -0.36)}} & \cellcolor{negcolor!47.2}\makecell{-0.40\\{\tiny(-0.45, -0.35)}} & \cellcolor{lightgrey!26.9}0.32 \\
\addlinespace[\rowspacing]
\rowcolor{altrow} compliments the user's question or prompt & \cellcolor{poscolor!25.5}\makecell{0.20\\{\tiny(0.16, 0.24)}} & \cellcolor{negcolor!11.8}\makecell{-0.10\\{\tiny(-0.13, -0.07)}} & \cellcolor{negcolor!0.2}\makecell{\transparent{0.3}-0.00\\\transparent{0.3}{\tiny(-0.04, 0.03)}} & \cellcolor{lightgrey!25.4}0.30 \\
\addlinespace[\rowspacing]
uses more casual language & \cellcolor{poscolor!15.2}\makecell{0.12\\{\tiny(0.09, 0.15)}} & \cellcolor{poscolor!3.8}\makecell{\transparent{0.3}0.03\\\transparent{0.3}{\tiny(0.00, 0.06)}} & \cellcolor{poscolor!41.7}\makecell{0.33\\{\tiny(0.28, 0.37)}} & \cellcolor{lightgrey!25.2}0.30 \\
\addlinespace[\rowspacing]
\rowcolor{altrow} makes more confident statements & \cellcolor{poscolor!16.2}\makecell{0.13\\{\tiny(0.08, 0.18)}} & \cellcolor{negcolor!14.6}\makecell{-0.12\\{\tiny(-0.17, -0.08)}} & \cellcolor{poscolor!1.8}\makecell{\transparent{0.3}0.01\\\transparent{0.3}{\tiny(-0.04, 0.07)}} & \cellcolor{lightgrey!21.2}0.25 \\
\addlinespace[\rowspacing]
uses a more enthusiastic tone & \cellcolor{poscolor!14.4}\makecell{0.11\\{\tiny(0.08, 0.15)}} & \cellcolor{negcolor!16.3}\makecell{-0.14\\{\tiny(-0.17, -0.11)}} & \cellcolor{negcolor!5.4}\makecell{\transparent{0.3}-0.05\\\transparent{0.3}{\tiny(-0.08, -0.01)}} & \cellcolor{lightgrey!21.2}0.25 \\
\addlinespace[\rowspacing]
\rowcolor{altrow} is more empathetic to the user & \cellcolor{poscolor!18.9}\makecell{0.15\\{\tiny(0.11, 0.19)}} & \cellcolor{negcolor!11.1}\makecell{-0.09\\{\tiny(-0.13, -0.06)}} & \cellcolor{poscolor!11.6}\makecell{0.09\\{\tiny(0.05, 0.13)}} & \cellcolor{lightgrey!20.5}0.24 \\
\addlinespace[\rowspacing]
is more creative and original & \cellcolor{poscolor!16.4}\makecell{0.13\\{\tiny(0.09, 0.17)}} & \cellcolor{negcolor!6.1}\makecell{\transparent{0.3}-0.05\\\transparent{0.3}{\tiny(-0.08, -0.02)}} & \cellcolor{poscolor!9.1}\makecell{0.07\\{\tiny(0.04, 0.11)}} & \cellcolor{lightgrey!15.3}0.18 \\
\addlinespace[\rowspacing]
\rowcolor{altrow} is more factually correct & \cellcolor{poscolor!12.9}\makecell{0.10\\{\tiny(0.07, 0.14)}} & \cellcolor{negcolor!6.1}\makecell{\transparent{0.3}-0.05\\\transparent{0.3}{\tiny(-0.09, -0.02)}} & \cellcolor{poscolor!5.3}\makecell{\transparent{0.3}0.04\\\transparent{0.3}{\tiny(0.01, 0.07)}} & \cellcolor{lightgrey!12.9}0.15 \\
\addlinespace[\rowspacing]
uses more mathematical symbols and notation & \cellcolor{poscolor!5.6}\makecell{0.04\\{\tiny(0.02, 0.07)}} & \cellcolor{negcolor!10.4}\makecell{-0.09\\{\tiny(-0.11, -0.06)}} & \cellcolor{negcolor!9.0}\makecell{-0.08\\{\tiny(-0.10, -0.05)}} & \cellcolor{lightgrey!11.1}0.13 \\
\addlinespace[\rowspacing]
\rowcolor{altrow} actively engages the reader with rhetorical questions & \cellcolor{poscolor!13.1}\makecell{0.10\\{\tiny(0.07, 0.14)}} & \cellcolor{negcolor!2.4}\makecell{\transparent{0.3}-0.02\\\transparent{0.3}{\tiny(-0.04, -0.00)}} & \cellcolor{poscolor!3.8}\makecell{\transparent{0.3}0.03\\\transparent{0.3}{\tiny(0.01, 0.06)}} & \cellcolor{lightgrey!10.4}0.12 \\
\addlinespace[\rowspacing]
uses more emojis & \cellcolor{poscolor!10.4}\makecell{0.08\\{\tiny(0.05, 0.11)}} & \cellcolor{negcolor!4.0}\makecell{-0.03\\{\tiny(-0.05, -0.02)}} & \cellcolor{poscolor!6.3}\makecell{0.05\\{\tiny(0.02, 0.08)}} & \cellcolor{lightgrey!9.7}0.12 \\
\addlinespace[\rowspacing]
\rowcolor{altrow} more strictly follows the requested output format & \cellcolor{poscolor!1.5}\makecell{\transparent{0.3}0.01\\\transparent{0.3}{\tiny(-0.02, 0.05)}} & \cellcolor{negcolor!4.7}\makecell{\transparent{0.3}-0.04\\\transparent{0.3}{\tiny(-0.09, 0.01)}} & \cellcolor{negcolor!11.8}\makecell{-0.10\\{\tiny(-0.14, -0.05)}} & \cellcolor{lightgrey!9.4}0.11 \\
\addlinespace[\rowspacing]
includes more references to other sources & \cellcolor{poscolor!6.8}\makecell{0.05\\{\tiny(0.03, 0.08)}} & \cellcolor{negcolor!4.2}\makecell{\transparent{0.3}-0.04\\\transparent{0.3}{\tiny(-0.06, -0.01)}} & \cellcolor{negcolor!2.6}\makecell{\transparent{0.3}-0.02\\\transparent{0.3}{\tiny(-0.04, -0.00)}} & \cellcolor{lightgrey!7.6}0.09 \\
\addlinespace[\rowspacing]
\rowcolor{altrow} expresses more emotion & \cellcolor{poscolor!2.8}\makecell{\transparent{0.3}0.02\\\transparent{0.3}{\tiny(0.00, 0.04)}} & \cellcolor{negcolor!5.2}\makecell{-0.04\\{\tiny(-0.06, -0.02)}} & \cellcolor{poscolor!1.3}\makecell{\transparent{0.3}0.01\\\transparent{0.3}{\tiny(-0.01, 0.03)}} & \cellcolor{lightgrey!5.5}0.07 \\
\addlinespace[\rowspacing]
provides conclusions without full reasoning & \cellcolor{negcolor!1.2}\makecell{\transparent{0.3}-0.01\\\transparent{0.3}{\tiny(-0.02, 0.00)}} & \cellcolor{poscolor!5.8}\makecell{0.05\\{\tiny(0.03, 0.07)}} & \cellcolor{poscolor!2.0}\makecell{\transparent{0.3}0.02\\\transparent{0.3}{\tiny(0.00, 0.03)}} & \cellcolor{lightgrey!4.7}0.06 \\
\addlinespace[\rowspacing]
\rowcolor{altrow} includes more ethical considerations & \cellcolor{negcolor!2.1}\makecell{\transparent{0.3}-0.02\\\transparent{0.3}{\tiny(-0.05, 0.01)}} & \cellcolor{negcolor!7.5}\makecell{-0.06\\{\tiny(-0.09, -0.04)}} & \cellcolor{negcolor!4.5}\makecell{\transparent{0.3}-0.04\\\transparent{0.3}{\tiny(-0.07, -0.01)}} & \cellcolor{lightgrey!3.9}0.05 \\
\addlinespace[\rowspacing]
acknowledges own limitations or uncertainty more & \cellcolor{negcolor!0.9}\makecell{\transparent{0.3}-0.01\\\transparent{0.3}{\tiny(-0.04, 0.02)}} & \cellcolor{negcolor!2.1}\makecell{\transparent{0.3}-0.02\\\transparent{0.3}{\tiny(-0.05, 0.01)}} & \cellcolor{poscolor!2.3}\makecell{\transparent{0.3}0.02\\\transparent{0.3}{\tiny(-0.01, 0.05)}} & \cellcolor{lightgrey!3.0}0.04 \\
\addlinespace[\rowspacing]
\rowcolor{altrow} is more optimistic & \cellcolor{negcolor!0.2}\makecell{\transparent{0.3}-0.00\\\transparent{0.3}{\tiny(-0.02, 0.02)}} & \cellcolor{negcolor!4.5}\makecell{-0.04\\{\tiny(-0.06, -0.02)}} & \cellcolor{negcolor!2.6}\makecell{\transparent{0.3}-0.02\\\transparent{0.3}{\tiny(-0.04, 0.00)}} & \cellcolor{lightgrey!3.0}0.04 \\
\addlinespace[\rowspacing]
uses more humour & \cellcolor{poscolor!1.3}\makecell{\transparent{0.3}0.01\\\transparent{0.3}{\tiny(-0.01, 0.03)}} & \cellcolor{negcolor!0.7}\makecell{\transparent{0.3}-0.01\\\transparent{0.3}{\tiny(-0.02, 0.01)}} & \cellcolor{poscolor!3.5}\makecell{\transparent{0.3}0.03\\\transparent{0.3}{\tiny(0.01, 0.05)}} & \cellcolor{lightgrey!2.9}0.03 \\
\end{tabular}
\end{minipage}

    \caption{\textbf{Top 30 traits for OpenAI GPT models (ChatGPT variants) in terms of change in \emph{strength} over time.} The reference model (GPT-4o) is omitted here, as by definition its strength values are 0. Sorted by maximum difference. \strengthexpl{} All measurements are relative to GPT-4o. Mean maximum difference is 0.26.}
    \label{fig:app:results:gpt_strength}
\end{figure}

\begin{figure}[ht]
    \centering
    \begin{minipage}[t]{1\textwidth}
\centering
\sffamily
\tablefontsize
\begin{tabular}{
    >{\raggedright\arraybackslash}m{0.23\linewidth}
    @{\hspace{13.5pt}} 
    >{\centering\arraybackslash}m{0.12954545454545455\linewidth}
    @{\hspace{13.5pt}} 
    >{\centering\arraybackslash}m{0.12954545454545455\linewidth}
    @{\hspace{13.5pt}} 
    >{\centering\arraybackslash}m{0.12954545454545455\linewidth}
    @{\hspace{13.5pt}} 
    >{\centering\arraybackslash}m{0.12954545454545455\linewidth}
}

\textbf{Generating a response that...} & \textbf{Google \textit{Gemini-2.5-Pro}} & \textbf{Google \textit{Gemini-3-Pro}} & \textbf{Google \textit{Gemini-3.1-Pro}} & \textbf{Max diff} \\
\toprule
\rowcolor{altrow} is more polite & \cellcolor{poscolor!63.2}\makecell{0.47\\{\tiny(0.41, 0.53)}} & \cellcolor{negcolor!38.6}\makecell{-0.18\\{\tiny(-0.22, -0.13)}} & \cellcolor{negcolor!15.7}\makecell{\transparent{0.3}-0.07\\\transparent{0.3}{\tiny(-0.12, -0.02)}} & \cellcolor{lightgrey!100.0}0.65 \\
\addlinespace[\rowspacing]
has a friendlier tone & \cellcolor{poscolor!64.0}\makecell{0.48\\{\tiny(0.41, 0.53)}} & \cellcolor{negcolor!21.5}\makecell{-0.10\\{\tiny(-0.14, -0.05)}} & \cellcolor{poscolor!5.2}\makecell{\transparent{0.3}0.04\\\transparent{0.3}{\tiny(-0.01, 0.09)}} & \cellcolor{lightgrey!88.9}0.57 \\
\addlinespace[\rowspacing]
\rowcolor{altrow} compliments the user's question or prompt & \cellcolor{poscolor!73.4}\makecell{0.55\\{\tiny(0.50, 0.59)}} & \cellcolor{negcolor!5.8}\makecell{\transparent{0.3}-0.03\\\transparent{0.3}{\tiny(-0.06, 0.00)}} & \cellcolor{poscolor!3.8}\makecell{\transparent{0.3}0.03\\\transparent{0.3}{\tiny(-0.01, 0.06)}} & \cellcolor{lightgrey!88.7}0.57 \\
\addlinespace[\rowspacing]
is more verbose & \cellcolor{poscolor!100.0}\makecell{0.74\\{\tiny(0.68, 0.79)}} & \cellcolor{poscolor!51.7}\makecell{0.38\\{\tiny(0.30, 0.46)}} & \cellcolor{poscolor!78.8}\makecell{0.59\\{\tiny(0.51, 0.65)}} & \cellcolor{lightgrey!55.7}0.36 \\
\addlinespace[\rowspacing]
\rowcolor{altrow} uses a more enthusiastic tone & \cellcolor{poscolor!48.4}\makecell{0.36\\{\tiny(0.31, 0.41)}} & \cellcolor{poscolor!1.6}\makecell{\transparent{0.3}0.01\\\transparent{0.3}{\tiny(-0.02, 0.05)}} & \cellcolor{poscolor!16.4}\makecell{0.12\\{\tiny(0.08, 0.16)}} & \cellcolor{lightgrey!53.9}0.35 \\
\addlinespace[\rowspacing]
more actively engages with the user & \cellcolor{poscolor!34.1}\makecell{0.25\\{\tiny(0.19, 0.31)}} & \cellcolor{negcolor!17.5}\makecell{-0.08\\{\tiny(-0.12, -0.03)}} & \cellcolor{poscolor!4.4}\makecell{\transparent{0.3}0.03\\\transparent{0.3}{\tiny(-0.02, 0.08)}} & \cellcolor{lightgrey!51.7}0.33 \\
\addlinespace[\rowspacing]
\rowcolor{altrow} uses more personal pronouns (I, we, you) & \cellcolor{poscolor!51.9}\makecell{0.39\\{\tiny(0.33, 0.44)}} & \cellcolor{poscolor!9.6}\makecell{\transparent{0.3}0.07\\\transparent{0.3}{\tiny(0.02, 0.12)}} & \cellcolor{poscolor!20.3}\makecell{0.15\\{\tiny(0.10, 0.20)}} & \cellcolor{lightgrey!48.8}0.31 \\
\addlinespace[\rowspacing]
is more empathetic to the user & \cellcolor{poscolor!40.3}\makecell{0.30\\{\tiny(0.25, 0.34)}} & \cellcolor{poscolor!3.8}\makecell{\transparent{0.3}0.03\\\transparent{0.3}{\tiny(-0.01, 0.06)}} & \cellcolor{poscolor!19.4}\makecell{0.14\\{\tiny(0.11, 0.18)}} & \cellcolor{lightgrey!42.1}0.27 \\
\addlinespace[\rowspacing]
\rowcolor{altrow} is more concise & \cellcolor{negcolor!100.0}\makecell{-0.45\\{\tiny(-0.50, -0.40)}} & \cellcolor{negcolor!40.8}\makecell{-0.19\\{\tiny(-0.24, -0.13)}} & \cellcolor{negcolor!59.7}\makecell{-0.27\\{\tiny(-0.32, -0.21)}} & \cellcolor{lightgrey!41.6}0.27 \\
\addlinespace[\rowspacing]
provides more examples & \cellcolor{poscolor!72.6}\makecell{0.54\\{\tiny(0.49, 0.59)}} & \cellcolor{poscolor!40.8}\makecell{0.30\\{\tiny(0.24, 0.36)}} & \cellcolor{poscolor!47.6}\makecell{0.35\\{\tiny(0.30, 0.41)}} & \cellcolor{lightgrey!36.7}0.24 \\
\addlinespace[\rowspacing]
\rowcolor{altrow} has more structured formatting & \cellcolor{poscolor!93.8}\makecell{0.70\\{\tiny(0.64, 0.75)}} & \cellcolor{poscolor!71.7}\makecell{0.53\\{\tiny(0.47, 0.59)}} & \cellcolor{poscolor!67.1}\makecell{0.50\\{\tiny(0.43, 0.56)}} & \cellcolor{lightgrey!30.8}0.20 \\
\addlinespace[\rowspacing]
uses more formal language & \cellcolor{poscolor!22.6}\makecell{0.17\\{\tiny(0.11, 0.22)}} & \cellcolor{poscolor!3.6}\makecell{\transparent{0.3}0.03\\\transparent{0.3}{\tiny(-0.03, 0.08)}} & \cellcolor{poscolor!0.8}\makecell{\transparent{0.3}0.01\\\transparent{0.3}{\tiny(-0.05, 0.06)}} & \cellcolor{lightgrey!25.1}0.16 \\
\addlinespace[\rowspacing]
\rowcolor{altrow} makes more confident statements & \cellcolor{poscolor!76.1}\makecell{0.57\\{\tiny(0.51, 0.61)}} & \cellcolor{poscolor!54.7}\makecell{0.41\\{\tiny(0.36, 0.45)}} & \cellcolor{poscolor!64.1}\makecell{0.48\\{\tiny(0.42, 0.52)}} & \cellcolor{lightgrey!24.6}0.16 \\
\addlinespace[\rowspacing]
actively engages the reader with rhetorical questions & \cellcolor{poscolor!21.0}\makecell{0.16\\{\tiny(0.12, 0.19)}} & \cellcolor{poscolor!10.1}\makecell{0.08\\{\tiny(0.05, 0.10)}} & \cellcolor{poscolor!11.8}\makecell{0.09\\{\tiny(0.06, 0.12)}} & \cellcolor{lightgrey!12.5}0.08 \\
\addlinespace[\rowspacing]
\rowcolor{altrow} is more creative and original & \cellcolor{poscolor!45.2}\makecell{0.34\\{\tiny(0.29, 0.38)}} & \cellcolor{poscolor!35.6}\makecell{0.26\\{\tiny(0.22, 0.30)}} & \cellcolor{poscolor!41.1}\makecell{0.31\\{\tiny(0.26, 0.35)}} & \cellcolor{lightgrey!11.0}0.07 \\
\addlinespace[\rowspacing]
provides a numbered list format & \cellcolor{poscolor!7.0}\makecell{\transparent{0.3}0.05\\\transparent{0.3}{\tiny(-0.01, 0.12)}} & \cellcolor{poscolor!16.2}\makecell{0.12\\{\tiny(0.06, 0.18)}} & \cellcolor{poscolor!13.4}\makecell{\transparent{0.3}0.10\\\transparent{0.3}{\tiny(0.04, 0.16)}} & \cellcolor{lightgrey!10.6}0.07 \\
\addlinespace[\rowspacing]
\rowcolor{altrow} is more optimistic & \cellcolor{poscolor!8.3}\makecell{0.06\\{\tiny(0.04, 0.09)}} & \cellcolor{negcolor!1.3}\makecell{\transparent{0.3}-0.01\\\transparent{0.3}{\tiny(-0.03, 0.02)}} & \cellcolor{poscolor!3.8}\makecell{\transparent{0.3}0.03\\\transparent{0.3}{\tiny(0.00, 0.05)}} & \cellcolor{lightgrey!10.6}0.07 \\
\addlinespace[\rowspacing]
agrees more with the user & \cellcolor{poscolor!11.0}\makecell{0.08\\{\tiny(0.06, 0.11)}} & \cellcolor{poscolor!2.7}\makecell{\transparent{0.3}0.02\\\transparent{0.3}{\tiny(0.00, 0.04)}} & \cellcolor{poscolor!3.6}\makecell{\transparent{0.3}0.03\\\transparent{0.3}{\tiny(0.00, 0.05)}} & \cellcolor{lightgrey!9.6}0.06 \\
\addlinespace[\rowspacing]
\rowcolor{altrow} uses more bold and italics text & \cellcolor{poscolor!95.7}\makecell{0.71\\{\tiny(0.65, 0.76)}} & \cellcolor{poscolor!87.6}\makecell{0.65\\{\tiny(0.59, 0.71)}} & \cellcolor{poscolor!91.7}\makecell{0.68\\{\tiny(0.63, 0.73)}} & \cellcolor{lightgrey!9.3}0.06 \\
\addlinespace[\rowspacing]
ends with a follow-up question & \cellcolor{negcolor!30.0}\makecell{-0.14\\{\tiny(-0.17, -0.10)}} & \cellcolor{negcolor!33.2}\makecell{-0.15\\{\tiny(-0.18, -0.11)}} & \cellcolor{negcolor!21.1}\makecell{-0.10\\{\tiny(-0.13, -0.06)}} & \cellcolor{lightgrey!8.5}0.05 \\
\addlinespace[\rowspacing]
\rowcolor{altrow} is more factually correct & \cellcolor{poscolor!27.7}\makecell{0.21\\{\tiny(0.17, 0.24)}} & \cellcolor{poscolor!24.9}\makecell{0.19\\{\tiny(0.14, 0.22)}} & \cellcolor{poscolor!31.8}\makecell{0.24\\{\tiny(0.19, 0.28)}} & \cellcolor{lightgrey!7.9}0.05 \\
\addlinespace[\rowspacing]
uses more mathematical symbols and notation & \cellcolor{negcolor!6.6}\makecell{\transparent{0.3}-0.03\\\transparent{0.3}{\tiny(-0.06, 0.00)}} & \cellcolor{poscolor!2.5}\makecell{\transparent{0.3}0.02\\\transparent{0.3}{\tiny(-0.02, 0.05)}} & \cellcolor{negcolor!3.1}\makecell{\transparent{0.3}-0.01\\\transparent{0.3}{\tiny(-0.04, 0.02)}} & \cellcolor{lightgrey!7.5}0.05 \\
\addlinespace[\rowspacing]
\rowcolor{altrow} includes more ethical considerations & \cellcolor{poscolor!14.2}\makecell{0.11\\{\tiny(0.07, 0.14)}} & \cellcolor{poscolor!8.2}\makecell{0.06\\{\tiny(0.03, 0.09)}} & \cellcolor{poscolor!12.0}\makecell{0.09\\{\tiny(0.06, 0.12)}} & \cellcolor{lightgrey!7.0}0.04 \\
\addlinespace[\rowspacing]
uses more casual language & \cellcolor{poscolor!7.8}\makecell{0.06\\{\tiny(0.03, 0.09)}} & \cellcolor{poscolor!3.8}\makecell{\transparent{0.3}0.03\\\transparent{0.3}{\tiny(0.00, 0.05)}} & \cellcolor{poscolor!9.9}\makecell{0.07\\{\tiny(0.05, 0.10)}} & \cellcolor{lightgrey!6.9}0.04 \\
\addlinespace[\rowspacing]
\rowcolor{altrow} expresses more emotion & \cellcolor{poscolor!6.2}\makecell{0.05\\{\tiny(0.02, 0.07)}} & \cellcolor{poscolor!0.8}\makecell{\transparent{0.3}0.01\\\transparent{0.3}{\tiny(-0.02, 0.03)}} & \cellcolor{poscolor!6.8}\makecell{0.05\\{\tiny(0.03, 0.07)}} & \cellcolor{lightgrey!6.9}0.04 \\
\addlinespace[\rowspacing]
reinforces user's beliefs more & \cellcolor{poscolor!6.2}\makecell{0.05\\{\tiny(0.03, 0.07)}} & \cellcolor{poscolor!0.5}\makecell{\transparent{0.3}0.00\\\transparent{0.3}{\tiny(-0.01, 0.01)}} & \cellcolor{poscolor!1.9}\makecell{\transparent{0.3}0.01\\\transparent{0.3}{\tiny(-0.00, 0.03)}} & \cellcolor{lightgrey!6.5}0.04 \\
\addlinespace[\rowspacing]
\rowcolor{altrow} more strictly follows the requested output format & \cellcolor{poscolor!8.9}\makecell{\transparent{0.3}0.07\\\transparent{0.3}{\tiny(0.03, 0.11)}} & \cellcolor{poscolor!6.3}\makecell{\transparent{0.3}0.05\\\transparent{0.3}{\tiny(0.01, 0.09)}} & \cellcolor{poscolor!9.9}\makecell{0.07\\{\tiny(0.03, 0.11)}} & \cellcolor{lightgrey!4.1}0.03 \\
\addlinespace[\rowspacing]
uses more emojis & \cellcolor{negcolor!6.6}\makecell{-0.03\\{\tiny(-0.05, -0.01)}} & \cellcolor{negcolor!5.4}\makecell{\transparent{0.3}-0.02\\\transparent{0.3}{\tiny(-0.04, -0.01)}} & \cellcolor{negcolor!0.9}\makecell{\transparent{0.3}-0.00\\\transparent{0.3}{\tiny(-0.02, 0.01)}} & \cellcolor{lightgrey!4.0}0.03 \\
\addlinespace[\rowspacing]
\rowcolor{altrow} acknowledges own limitations or uncertainty more & \cellcolor{negcolor!14.1}\makecell{-0.06\\{\tiny(-0.10, -0.03)}} & \cellcolor{negcolor!11.2}\makecell{\transparent{0.3}-0.05\\\transparent{0.3}{\tiny(-0.09, -0.02)}} & \cellcolor{negcolor!9.0}\makecell{\transparent{0.3}-0.04\\\transparent{0.3}{\tiny(-0.08, -0.01)}} & \cellcolor{lightgrey!3.6}0.02 \\
\addlinespace[\rowspacing]
includes more references to other sources & \cellcolor{poscolor!9.7}\makecell{0.07\\{\tiny(0.04, 0.10)}} & \cellcolor{poscolor!11.8}\makecell{0.09\\{\tiny(0.06, 0.12)}} & \cellcolor{poscolor!12.3}\makecell{0.09\\{\tiny(0.06, 0.12)}} & \cellcolor{lightgrey!3.0}0.02 \\
\end{tabular}
\end{minipage}

    \caption{\textbf{Top 30 traits for Google Gemini models in terms of change in \emph{strength} over time.} Sorted by maximum difference. \strengthexpl{} All measurements are relative to GPT-4o. Mean maximum difference is 0.13.}
    \label{fig:app:results:gemini_strength}
\end{figure}

\clearpage

\subsubsection{Llama-4-Maverick analysis}

\begin{figure}[ht]
    \centering
    \begin{minipage}[t]{0.48\textwidth}
\centering
\sffamily
\tablefontsize
\textbf{Traits stronger in arena relative to public model}\\[0.5em]
\begin{tabular}{
    >{\raggedright\arraybackslash}m{0.7\linewidth}
    @{\hspace{10pt}} 
    >{\centering\arraybackslash}m{0.19\linewidth}
}

\textbf{Generating a response that...} & \textbf{Strength} \\
\toprule
\rowcolor{altrow} is more verbose & \cellcolor{poscolor!100.0}\makecell{0.97\\{\tiny(0.96, 0.98)}} \\
\addlinespace[\rowspacing]
uses more bold and italics text & \cellcolor{poscolor!98.5}\makecell{0.96\\{\tiny(0.95, 0.97)}} \\
\addlinespace[\rowspacing]
\rowcolor{altrow} uses a more enthusiastic tone & \cellcolor{poscolor!97.8}\makecell{0.95\\{\tiny(0.94, 0.96)}} \\
\addlinespace[\rowspacing]
more actively engages with the user & \cellcolor{poscolor!97.8}\makecell{0.95\\{\tiny(0.94, 0.96)}} \\
\addlinespace[\rowspacing]
\rowcolor{altrow} uses more personal pronouns (I, we, you) & \cellcolor{poscolor!96.5}\makecell{0.94\\{\tiny(0.93, 0.95)}} \\
\addlinespace[\rowspacing]
compliments the user's question or prompt & \cellcolor{poscolor!94.4}\makecell{0.92\\{\tiny(0.91, 0.93)}} \\
\addlinespace[\rowspacing]
\rowcolor{altrow} has a friendlier tone & \cellcolor{poscolor!94.3}\makecell{0.92\\{\tiny(0.90, 0.93)}} \\
\addlinespace[\rowspacing]
expresses more emotion & \cellcolor{poscolor!89.6}\makecell{0.87\\{\tiny(0.86, 0.89)}} \\
\addlinespace[\rowspacing]
\rowcolor{altrow} is more empathetic to the user & \cellcolor{poscolor!85.9}\makecell{0.84\\{\tiny(0.82, 0.85)}} \\
\addlinespace[\rowspacing]
uses more casual language & \cellcolor{poscolor!84.8}\makecell{0.83\\{\tiny(0.81, 0.84)}} \\
\end{tabular}
\end{minipage}
\hfill
\begin{minipage}[t]{0.48\textwidth}
\centering
\sffamily
\tablefontsize
\textbf{Traits weaker in arena relative to public model}\\[0.5em]
\begin{tabular}{
    >{\raggedright\arraybackslash}m{0.7\linewidth}
    @{\hspace{10pt}} 
    >{\centering\arraybackslash}m{0.19\linewidth}
}

\textbf{Generating a response that...} & \textbf{Strength} \\
\toprule
\rowcolor{altrow} is more concise & \cellcolor{negcolor!100.0}\makecell{-0.75\\{\tiny(-0.76, -0.73)}} \\
\addlinespace[\rowspacing]
uses more formal language & \cellcolor{negcolor!49.2}\makecell{-0.37\\{\tiny(-0.40, -0.34)}} \\
\addlinespace[\rowspacing]
\rowcolor{altrow} more strictly follows the requested output format & \cellcolor{negcolor!18.8}\makecell{-0.14\\{\tiny(-0.16, -0.11)}} \\
\addlinespace[\rowspacing]
has a more avoidant tone & \cellcolor{negcolor!9.5}\makecell{-0.07\\{\tiny(-0.08, -0.06)}} \\
\addlinespace[\rowspacing]
\rowcolor{altrow} acknowledges own limitations or uncertainty more & \cellcolor{negcolor!4.7}\makecell{-0.03\\{\tiny(-0.06, -0.01)}} \\
\addlinespace[\rowspacing]
provides conclusions without full reasoning & \cellcolor{negcolor!3.4}\makecell{-0.03\\{\tiny(-0.03, -0.02)}} \\
\addlinespace[\rowspacing]
\rowcolor{altrow} contains less harmful information & \cellcolor{negcolor!2.4}\makecell{-0.02\\{\tiny(-0.03, -0.01)}} \\
\addlinespace[\rowspacing]
refuses to answer the question & \cellcolor{negcolor!2.3}\makecell{-0.02\\{\tiny(-0.02, -0.01)}} \\
\addlinespace[\rowspacing]
\rowcolor{altrow} suggests illegal activities & \cellcolor{poscolor!0.3}\makecell{\transparent{0.3}0.00\\\transparent{0.3}{\tiny(0.00, 0.01)}} \\
\addlinespace[\rowspacing]
is more offensive & \cellcolor{poscolor!0.8}\makecell{0.01\\{\tiny(0.00, 0.01)}} \\
\end{tabular}
\end{minipage}

    \caption{\textbf{Extended comparison of personality traits of the Chatbot Arena \emph{(arena)} and publicly released \emph{(public)} versions of Llama-4-Maverick.}}
    \label{fig:app:results:llama-4}
\end{figure}

\section{Models}
\label{app:models}

Throughout our experiments we use a diverse set of models from multiple providers. Below is a list of all models used, including their \emph{full name} (including provider) and the \emph{short name} used in the paper (in brackets). All models used via \url{https://openrouter.ai/}.

\begin{enumerate}
    \item \textbf{Anthropic}
    \begin{enumerate}
        \item \texttt{anthropic/claude-3.7-sonnet} (\texttt{Claude-3.7-Sonnet})
        \item \texttt{anthropic/claude-sonnet-4} (\texttt{Claude-Sonnet-4})
        \item \texttt{anthropic/claude-sonnet-4.5} (\texttt{Claude-Sonnet-4.5})
        \item \texttt{anthropic/claude-haiku-4.5} (\texttt{Claude-Haiku-4.5})
        \item \texttt{anthropic/claude-opus-4.7} (\texttt{Claude-Opus-4.7})
    \end{enumerate}
    \item \textbf{Google}
    \begin{enumerate}
        \item \texttt{google/gemini-2.5-pro} (\texttt{Gemini-2.5-Pro})
        \item \texttt{google/gemini-2.5-flash} (\texttt{Gemini-2.5-Flash})
        \item \texttt{google/gemini-3-pro-preview} (\texttt{Gemini-3-Pro})
        \item \texttt{google/gemini-3.1-pro-preview} (\texttt{Gemini-3.1-Pro})
    \end{enumerate}

    \item \textbf{Meta}
    \begin{enumerate}
        \item \texttt{meta-llama/llama-4-maverick} (\texttt{Llama-4-Maverick})\footnote{Note that, in addition, responses from a different non-public version of Maverick were used in \Cref{sec:results:llama4analysis}}
    \end{enumerate}
    \item \textbf{Mistral}
    \begin{enumerate}
        \item \texttt{mistralai/mistral-medium-3.1} (\texttt{Mistral-Medium-3.1})
        \item \texttt{mistralai/mistral-medium-3-5} (\texttt{Mistral-Medium-3.5})
    \end{enumerate}
    \item \textbf{OpenAI}
    \begin{enumerate}
        \item \texttt{openai/gpt-4.1-2025-04-14} (\texttt{GPT-4.1})
        \item \texttt{openai/gpt-4.1-mini} (\texttt{GPT-4.1-mini})
        \item \texttt{openai/gpt-4o-2024-11-20} (\texttt{GPT-4o})
        \item \texttt{openai/gpt-4o-mini-2024-07-18} (\texttt{GPT-4o-mini})
        \item \texttt{openai/gpt-5-2025-08-07} (\texttt{GPT-5})
        \item \texttt{openai/gpt-5-mini-2025-08-07} (\texttt{GPT-5-mini})
        \item \texttt{openai/gpt-5-chat} (\texttt{GPT-5-Chat})
        \item \texttt{openai/gpt-5.1-chat} (\texttt{GPT-5.1-Chat})
        \item \texttt{openai/gpt-5.3-chat} (\texttt{GPT-5.3-Chat})
        \item \texttt{openai/gpt-5.5} (\texttt{GPT-5.5})
        \item \texttt{openai/gpt-oss-20b} (\texttt{GPT-oss-20b})
    \end{enumerate}

    \item \textbf{xAI}
    \begin{enumerate}
        \item \texttt{x-ai/grok-4} (\texttt{Grok-4})
    \end{enumerate}
\end{enumerate}

\section{Compute resources}
\label{app:computeusage}

The overall compute costs for all new annotations created as part of the experiments included in this paper version is approximated to be slightly less than 150 USD.

\clearpage

\section{Prompts}
\label{app:prompts}

\subsection{Annotator prompt}
\label{app:prompt:annotator}

To instruct our annotators, we use the prompt shown in Listing \ref{lst:annotatorprompt} from the \emph{Inverse Constitutional AI} \citep{findeis2025inverse} package. To enable compute-efficient annotation, the annotator is asked to annotate multiple personality traits at the same time. We thank all contributors to the  package for their help improving this and the other prompts in the ICAI package.

\begin{lstlisting}[caption={Personality-selecting annotator prompt}, label={lst:annotatorprompt}]
<|im_start|>system
Your job is to check which sample is should be selected according to the given rules. You're an expert at this.
<|im_end|>
<|im_start|>user
Sample A:
{sample_a}

Sample B:
{sample_b}

Given the samples data above, check for each rule below which sample should be selected:
{summaries}

Answer in json format, e.g. {{0: "A", 1: "B", 2: "None", 3: "Both",...}}.
Put "A" if A is selected according to that rule. 
Put "B" if B is selected according to that rule.
Put "Both" if both A and B should be selected, and the rule is categorical so it is impossible to select only one.
Put "None" if a rule is not applicable to the two samples.
Otherwise, no ties are allowed, only one of "A", "B", "Both" or "None".
Vote for all rules, even if you are unsure.
DO NOT respond with any text apart from the json format above!
DO NOT add markdown formatting around JSON.
ONLY REPLY IN JSON FORMAT
<|im_end|>
\end{lstlisting}

\section{Implementation details}

\subsection{Model annotations}

\begin{table}[!h]
  \centering
  \caption{Target annotations for GPT-5.3 on an example dataset (a response is selected only if GPT-5.3 generated exactly one of the two, otherwise the annotation is null).}
  \label{app:tab:model_annotations}
  \resizebox{\linewidth}{!}{%
  \begin{tabular}{lllll}
    \toprule
    \textbf{\begin{tabular}[b]{@{}l@{}}prompt,\\ response\_a, response\_b\end{tabular}}
      & \textbf{model\_a} & \textbf{model\_b} & \textbf{Case}
      & \textbf{\begin{tabular}[b]{@{}l@{}}Target annotation\\ for GPT-5.3\end{tabular}} \\
    \midrule
    (\ldots) & GPT-5.3    & Opus-4.7    & target generated A only  & \texttt{text\_a} \\
    (\ldots) & Gemini-3.1 & GPT-5.3     & target generated B only  & \texttt{text\_b} \\
    (\ldots) & GPT-5.3    & GPT-5.3     & target generated both    & null \\
    (\ldots) & Sonnet-4   & gpt-oss-20b & target generated neither & null \\
    \bottomrule
  \end{tabular}%
  }
\end{table}

\clearpage

\end{document}